\documentclass[runningheads]{llncs}
\usepackage{graphicx}
\usepackage{comment}
\usepackage{amsmath,amssymb} 
\usepackage{color}

\usepackage{diagbox}
\usepackage{multirow}
\usepackage{subcaption}
\usepackage{xspace}
\usepackage{xparse}
\usepackage{xstring}
\usepackage{makecell}
\usepackage{paralist}
\usepackage[usenames,dvipsnames]{xcolor}
\usepackage{hyperref}
\usepackage{wrapfig}

\newcommand{\eyenetname}{{EyeNet}\xspace}
\newcommand{\eyenetstaticname}{\eyenetname{}\textsubscript{\tiny static}\xspace}
\newcommand{\eyenetrnnname}{\eyenetname{}\textsubscript{\tiny RNN}\xspace}
\newcommand{\eyenetlstmname}{\eyenetname{}\textsubscript{\tiny LSTM}\xspace}
\newcommand{\eyenetgruname}{\eyenetname{}\textsubscript{\tiny GRU}\xspace}
\newcommand{\refinenetname}{{GazeRefineNet}\xspace}
\newcommand{\refinenetstaticname}{\refinenetname{}\textsubscript{\tiny static}\xspace}
\newcommand{\refinenetrnnname}{\refinenetname{}\textsubscript{\tiny RNN}\xspace}
\newcommand{\refinenetlstmname}{\refinenetname{}\textsubscript{\tiny LSTM}\xspace}
\newcommand{\refinenetgruname}{\refinenetname{}\textsubscript{\tiny GRU}\xspace}
\newcommand{\datasetname}{\textsc{EVE}\xspace}

\newcommand{\dataseturl}{\url{https://ait.ethz.ch/projects/2020/EVE}}

\ExplSyntaxOn
\NewDocumentCommand{\fpcompare}{mmm}
 {
  \fp_compare:nTF { #1 } { #2 } { #3 }
 }
\ExplSyntaxOff
\makeatletter
\def\mathcolor#1#{\@mathcolor{#1}}
\def\@mathcolor#1#2#3{
  \protect\leavevmode
  \begingroup
    \color#1{#2}#3
  \endgroup
}
\makeatother
\newcommand{\result}[2]{
  \fpcompare{#2 < 0.0}
  {\ensuremath{#1\,^\mathrm{\color{Mahogany}{\uparrow\,\StrGobbleLeft{#2}{1}\%}}}}
  {\ensuremath{#1\,^\mathrm{\color{ForestGreen}{\downarrow\,#2\%}}}}
}
\newcommand{\resultb}[2]{\result{\mathbf{#1}}{#2}}

\makeatletter
\newcommand\semitiny{\@setfontsize\notsotiny{6.31415}{7.1828}}
\makeatother

\long\def\ignorethis#1{}

\definecolor{lightgray}{rgb}{0.92,0.92,0.92}
\definecolor{gray}{rgb}{0.35,0.35,0.35}
\definecolor{darkgreen}{rgb}{0.5,0.5,0}
\definecolor{MyBlue}{rgb}{0,0.2,0.8}
\definecolor{MyRed}{rgb}{0.8,0.2,0}
\definecolor{MyGreen}{rgb}{0.0,0.5,0.1}
\definecolor{MyGray}{rgb}{0.4,0.4,0.4}
\definecolor{airforceblue}{rgb}{0.36, 0.54, 0.66}

\newlength\paramargin
\newlength\figmarginstart
\newlength\figmargin
\newlength\subfigmargin
\newlength\secmargin
\newlength\subsecmargin
\newlength\tabmargin
\newlength\tabmarginstart
\newlength\eqmargin
\newlength\algmargin

\usepackage{array}
\newcolumntype{L}[1]{>{\raggedright\let\newline\\\arraybackslash\hspace{0pt}}m{#1}}
\newcolumntype{C}[1]{>{\centering\let\newline\\\arraybackslash\hspace{0pt}}m{#1}}
\newcolumntype{R}[1]{>{\raggedleft\let\newline\\\arraybackslash\hspace{0pt}}m{#1}}

\def\etal{et~al.\xspace}

\newcommand{\bv}[1]{\mathbf{#1}}
\newcommand{\bvh}[1]{\hat{\mathbf{#1}}}

\begin{document}
\pagestyle{headings}
\mainmatter
\def\ECCVSubNumber{1728}  

\title{Towards End-to-end Video-based Eye-Tracking} 

\begin{comment}
\titlerunning{ECCV-20 submission ID \ECCVSubNumber} 
\authorrunning{ECCV-20 submission ID \ECCVSubNumber} 
\author{Anonymous ECCV submission}
\institute{Paper ID \ECCVSubNumber}
\end{comment}

\titlerunning{Towards End-to-end Video-based Eye-tracking}
\author{Seonwook Park \and Emre Aksan \and Xucong Zhang \and Otmar Hilliges}
\authorrunning{S. Park et al.}
\institute{
Department of Computer Science, ETH Zurich \\
\email{\{firstname.lastname\}@inf.ethz.ch}
}
\maketitle

\begin{abstract}
Estimating eye-gaze from images alone is a challenging task, in large parts due to un-observable person-specific factors. Achieving high accuracy typically requires labeled data from test users which may not be attainable in real applications. 
We observe that there exists a strong relationship between what users are looking at and the appearance of the user's eyes. In response to this understanding, we propose a novel dataset and accompanying method which aims to explicitly learn these semantic and temporal relationships.
Our video dataset consists of time-synchronized screen recordings, user-facing camera views, and eye gaze data, which allows for new benchmarks in temporal gaze tracking as well as label-free refinement of gaze.
Importantly, we demonstrate that the fusion of information from visual stimuli as well as eye images can lead towards achieving performance similar to literature-reported figures acquired through supervised personalization.
Our final method yields significant performance improvements on our proposed \datasetname dataset, with up to $28\%$ improvement in Point-of-Gaze estimates (resulting in $2.49^\circ$ in angular error), paving the path towards high-accuracy screen-based eye tracking purely from webcam sensors.
The dataset and reference source code are available at \dataseturl

\keywords{Eye Tracking, Gaze Estimation, Computer Vision Dataset}
\end{abstract}
 \section{Introduction}

The task of gaze estimation from a single low-cost RGB sensor is an important topic in Computer Vision and Machine Learning.
It is an essential component in intelligent user interfaces \cite{Feit2017CHI,Biedert2010CHIEA}, user state awareness \cite{Huang2016CHI,Fridman2018CHI}, and serves as input modality to Computer Vision problems such as zero-shot learning \cite{Karessli2017CVPR}, object referral \cite{Vasudevan2018CVPR}, and human attention estimation \cite{Chong2018ECCV}.
Un-observable person-specific differences inherent in the problem are challenging to tackle and as such high accuracy general purpose gaze estimators are hard to attain. In response, person-specific adaptation techniques \cite{Park2019ICCV,Liu2018BMVC,Linden2019ICCVW} have seen much attention, albeit at the cost of requiring test-user-specific labels.
We propose a dataset and accompanying method which holistically combine multiple sources of information explicitly. This novel approach yields large performance improvements without needing ground-truth labels from the final target user.
Our large-scale dataset (\datasetname) and network architecture (\refinenetname) effectively showcase the newly proposed task and demonstrate up to $28\%$ in performance improvements.

The human gaze can be seen as a closed-loop feedback system, whereby the appearance of target objects or regions (or visual stimuli) incur particular movements in the eyes. Many works consider this interplay in related but largely separate strands of research, for instance in estimating gaze from images of the user (bottom-up, e.g.~\cite{Wang2019CVPR_Bayesian}) or post-hoc comparison of the eye movements with the visual distribution of the presented stimuli (top-down, e.g.~\cite{Sugano2015UIST}).
Furthermore, gaze estimation is often posed as a frame-by-frame estimation problem despite its rich temporal dynamics.
In this paper, we suggest that by taking advantage of the interaction between user's eye movements and what they are looking at, significant improvements in gaze estimation accuracy can be attained even in the \emph{absence of labeled samples} from the final target. This can be done without explicit gaze estimator personalization.
We are not aware of existing datasets that would allow for the study of these semantic relations and temporal dynamics.
Therefore, we introduce a novel dataset designed to facilitate research on the joint contributions of dynamic eye gaze and visual stimuli. We dub this dataset the \datasetname dataset (\textbf{E}nd-to-end \textbf{V}ideo-based \textbf{E}ye-tracking).
\datasetname is collected from 54 participants and consists of 4 camera views, over 12 million frames and 1327 unique visual stimuli (images, video, text), adding up to approximately 105 hours of video data in total.

Accompanying the proposed \datasetname dataset, we introduce a novel bottom-up-and-top-down approach to estimating the user's point of gaze.
The Point-of-Gaze (PoG) refers to the actual target of a person's gaze as measured on the screen plane in metric units or pixels.
In our method, we exploit the fact that more visually salient regions on a screen often coincide with the gaze.
Unlike previous methods which adopt and thus depend on pre-trained models of visual saliency \cite{Sugano2015UIST,Sugano2010CVPR,Chen2011CVPR}, we define our task as that of online and conditional PoG refinement.  In this setting a model takes raw screen content and an initial gaze estimate as explicit conditions, to predict the final and refined PoG.
Our final architecture yields significant improvements in predicted PoG accuracy on the proposed dataset.
We achieve a mean test error of $2.49$ degrees in gaze direction or $2.75$cm ($95.59$ pixels) in screen-space Euclidean distance. This is an improvement of up to $28\%$ compared to estimates of gaze from an architecture that does not consider screen content.
We thus demonstrate a meaningful step towards the proliferation of screen-based eye tracking technology.

In summary, we propose the following contributions:
\begin{itemize}[\textbullet]
\item A new task of online point-of-gaze (PoG) refinement, which combines bottom-up (eye appearance) and top-down (screen content) information to allow for a truly end-to-end learning of human gaze,

\item \datasetname, a large-scale video dataset of over 12 million frames from 54 participants consisting of 4 camera views, natural eye movements (as opposed to following specific intructions or smoothly moving targets), pupil size annotations, and screen content video to enable the new task, 

\item a novel method for eye gaze refinement which exploits the complementary sources of information jointly for improved PoG estimates, in the absence of ground-truth annotations from the user.
\end{itemize}
In combination these contributions allow us to demonstrate a gaze estimator performance of $2.49^\circ$ in angular error, comparing favorably with supervised person-specific model adaptation methods \cite{Park2019ICCV,Linden2019ICCVW,Chen2020WACV}.

 \begin{table}[t]
    \caption{Comparison of \datasetname with existing screen-based datasets. \datasetname is the first to provide natural eye movements (free-viewing, without specific instructions) synchronized with full-frame user-facing video and screen content}
    \label{tab:comparisons}
    \centering
    \renewcommand\arraystretch{1.2}
    \renewcommand\theadfont{\scriptsize}
    \scriptsize
    \begin{tabular}{|lr|c|c|c|c|c|c|c|c|}
        \hline
        Name && 
        Region & 
        \thead{\#\\Subjects} & 
        \thead{\#\\Samples} & 
        \thead{Temporal\\Data} & 
        \thead{Natural\\Eye\\Movements} & 
        \thead{Screen\\Content\\Video} & 
        \thead{Publicly\\Available} \\
        \hline
        Columbia Gaze & \cite{Smith2013UIST}           & Frame &    56 &      5,800 &          - & N     &    N & Y \\
        EYEDIAP       & \cite{FunesMora2014ETRA}       & Frame &    16 &     62,500 &       30Hz & \hspace{1.5mm}N$^*$ &    N & Y \\
        UT Multiview  & \cite{Sugano2014CVPR}          &  Eyes &    50 &     64,000 &          - & N     &    N & Y \\
        MPIIGaze      & \cite{Zhang2015CVPR}           &  Eyes &    15 &    213,659 &          - & N     &    N & Y \\
        TabletGaze    & \cite{Huang2017MVA}            & Frame &    51 &      1,785 &          - & N     &    N & Y \\
        GazeCapture   & \cite{Krafka2016CVPR}          & Frame & 1,474 &  2,129,980 &          - & N     &    N & Y \\
        Deng and Zhu  & \cite{Deng2017ICCV}            &  Eyes &   200 &    240,000 &          - & N     &    N & N \\
        MPIIFaceGaze  & \cite{Zhang2017CVPRW}          &  Face &    15 &     37,639 &          - & N     &    N & Y \\
        DynamicGaze   & \cite{Wang2019CVPR_Temporal}   &  Eyes &    20 &    645,000 & $\sim$30Hz & Y     &    N & N \\
        \hline    
        \datasetname (Ours) &                          & Frame &    54 & 12,308,334 & 30Hz, 60Hz & Y     & 30Hz & Y \\
        \hline
    \end{tabular}
    \\[1mm]
    \begin{minipage}{\columnwidth}
    \raggedleft $^*$ Only smooth pursuits eye movements are available. \hspace*{3mm}
    \end{minipage}
\end{table}

\section{Related Work}

In our work we consider the task of remote gaze estimation from RGB, where a monocular camera is located away from and facing a user. 
We outline here recent approaches, proposed datasets, and relevant methods for refining gaze estimates.

\subsection{Remote Gaze Estimation}

Remote gaze estimation from unmodified monocular sensors is challenging due to the lack of reference features such as reflections from near infra-red light sources. 
Recent methods have increasingly used machine learning methods to tackle this problem \cite{Baluja1993NeurIPS,Lu2011ICCV,Papoutsaki2016IJCAI} with extensions to allow for greater variations in head pose \cite{Lu2011BMVC,Ranjan2018CVPRW,Deng2017ICCV}. 
The task of cross-person gaze estimation is defined as one where a model is evaluated on a previously unseen set of participants. Several extensions have been proposed for this challenging task in terms of self-assessed uncertainty \cite{Cheng2018ECCV}, novel representations \cite{Park2018ECCV,Park2018ETRA,Yu2020CVPR}, and Bayesian learning \cite{Wang2018CVPR,Wang2019CVPR_Bayesian}.

Novel datasets have contributed to the progress of gaze estimation methods and the reporting of their performance, notably in challenging illumination settings \cite{Zhang2015CVPR,Zhang2017CVPRW,Krafka2016CVPR}, or at greater distances from the user \cite{Kellnhofer2019ICCV,Fischer2018ECCV,FunesMora2014ETRA} where image details are lost. Screen-based gaze estimation datasets have had a particular focus \cite{Zhang2015CVPR,Zhang2017CVPRW,Krafka2016CVPR,Huang2017MVA,Deng2017ICCV,FunesMora2014ETRA,Martinikorena2018PETMEI} due to the practical implications in modern times, with digital devices being used more frequently.
Very few existing datasets include videos, and even then often consist of participants gazing at points \cite{Huang2017MVA} or following smoothly moving targets only (via smooth pursuits) \cite{FunesMora2014ETRA}. While the RT-GENE dataset includes natural eye movement patterns such as fixations and saccades, it is not designed for the task of screen-based gaze estimation \cite{Fischer2018ECCV}. The recently proposed DynamicGaze dataset \cite{Wang2019CVPR_Temporal} includes natural eye movements from 20 participants gazing upon video stimuli.
However, it is yet to be publicly released and it is unclear if it will contain screen-content synchronization.
We are the first to provide a video dataset with full camera frames and associated eye gaze and pupil size data, in conjunction with screen content. Furthermore, \datasetname includes a large number of participants (=54) and frames (12.3M) over a large set of visual stimuli (1004 images, 161 videos, and 162 wikipedia pages).

\subsection{Temporal Models for Gaze Estimation}

Temporal modelling of eye gaze is an emerging research topic. An initial work demonstrates the use of a recurrent neural network (RNN) in conjunction with a convolutional neural network (CNN) for feature extraction \cite{Palmero2018BMVC}. While no improvements are shown for gaze estimates in the screen-space, results on smooth pursuits sequence of the EYEDIAP dataset \cite{FunesMora2014ETRA} are encouraging. 
In \cite{Wang2019CVPR_Temporal}, a top-down approach for gaze signal filtering is presented, where a probabilistic estimate of state (fixation, saccade, or smooth pursuits) is initially made, and consequently a state-specific linear dynamical system is applied to refine the initially predicted gaze. Improvements in gaze estimation performance are demonstrated on a custom dataset.
As one of our evaluations, we re-confirm previous findings that a temporal gaze estimation network can improve on a static gaze estimation network. We demonstrate this on our novel video dataset, which due to its diversity of visual stimuli and large number of participants should allow for future works to benchmark their improvements well.

\subsection{Refining Gaze Estimates}

While eye gaze direction (and subsequent Point-of-Gaze) can be predicted just from images of the eyes or face of a given user, an initial estimate can be improved with additional data. Accordingly, various methods have been proposed to this end.
A primary example is that of using few samples of labeled data - often dubbed ``person-specific gaze estimation'' - where a pre-trained neural network is fine-tuned or otherwise adapted on very few samples of a target test person's data, to yield performance improvements on the final test data from the same person.
Building on top of initial works \cite{Krafka2016CVPR,Park2018ETRA}, more recent works have demonstrated significant performance improvements with as few as $9$ calibration samples or less \cite{Liu2018BMVC,Park2019ICCV,Linden2019ICCVW,Yu2019CVPR,Chen2020WACV}.
Although the performance improvements are impressive, all such methods still require labeled samples from the final user.

Alternative approaches to refining gaze estimates in the screen-based setting, consider the predicted visual saliency of the screen content. Given a sufficient time horizon, it is possible to align estimates for PoG so-far, with an estimate for visual saliency \cite{Sugano2010CVPR,Sugano2008ECCV,Chen2011CVPR,Alnajar2013ICCV,Wang2016ETRA}. However, visual saliency estimation methods can over-fit to presented training data. Hence, methods have been suggested to merge estimates of multiple saliency models \cite{Sugano2015UIST} or use face positions as likely gaze targets \cite{Sugano2010CVPR}.
We propose an alternate and direct approach, which formulates the problem of gaze refinement as one that is conditioned explicitly on screen content and an initial gaze estimate.

 \section{The \datasetname Dataset}\label{sec:dataset}

\begin{figure}[t]
    \centering
    \begin{subfigure}[c]{0.515\columnwidth}
        \includegraphics[width=\columnwidth]{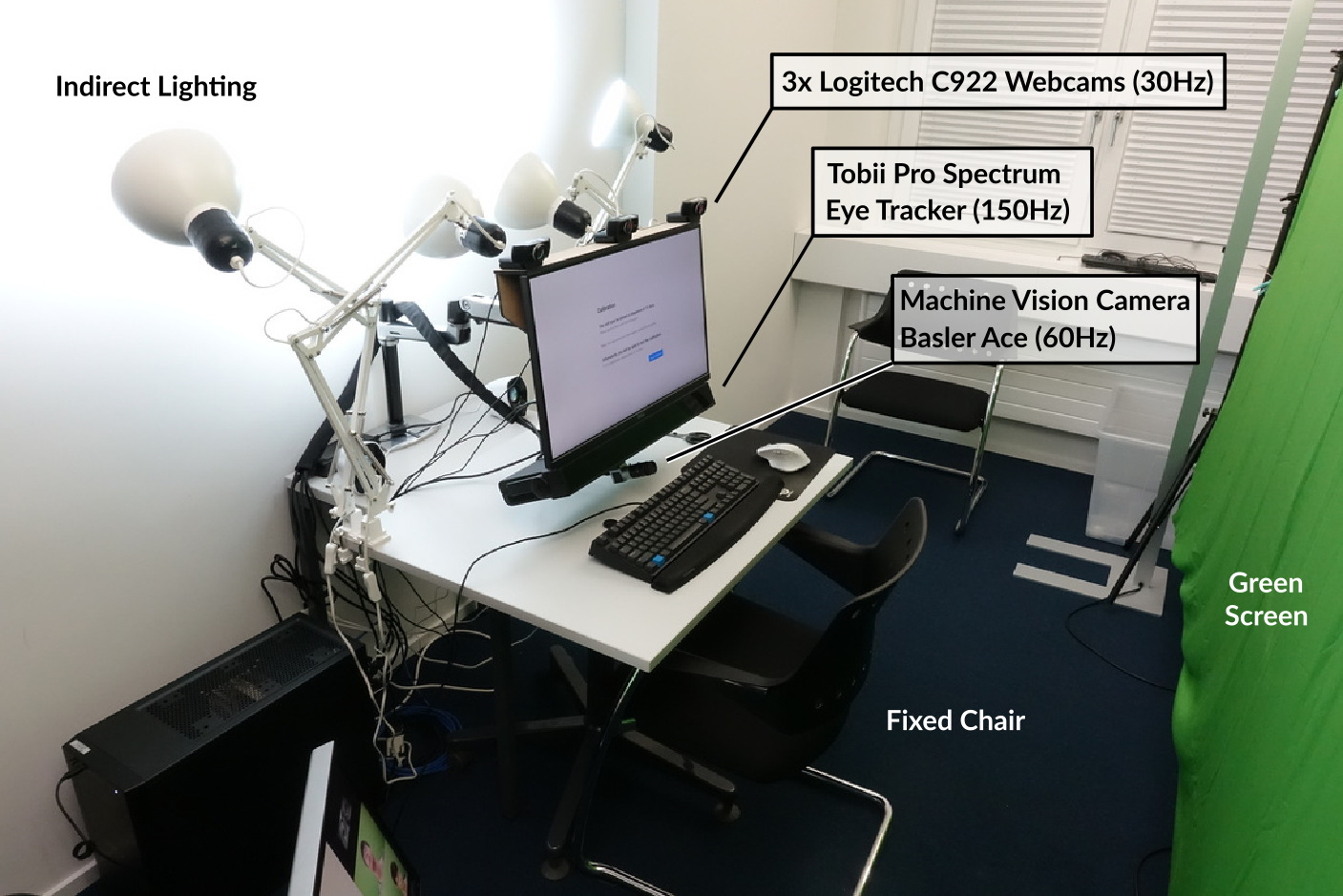}
        \caption{Collection Setup} \label{fig:collection_setup}
    \end{subfigure}
    \hfill
    \begin{subfigure}[c]{0.47\columnwidth}
        \begin{minipage}[t]{0.48\columnwidth}\centering
            \includegraphics[width=\columnwidth]{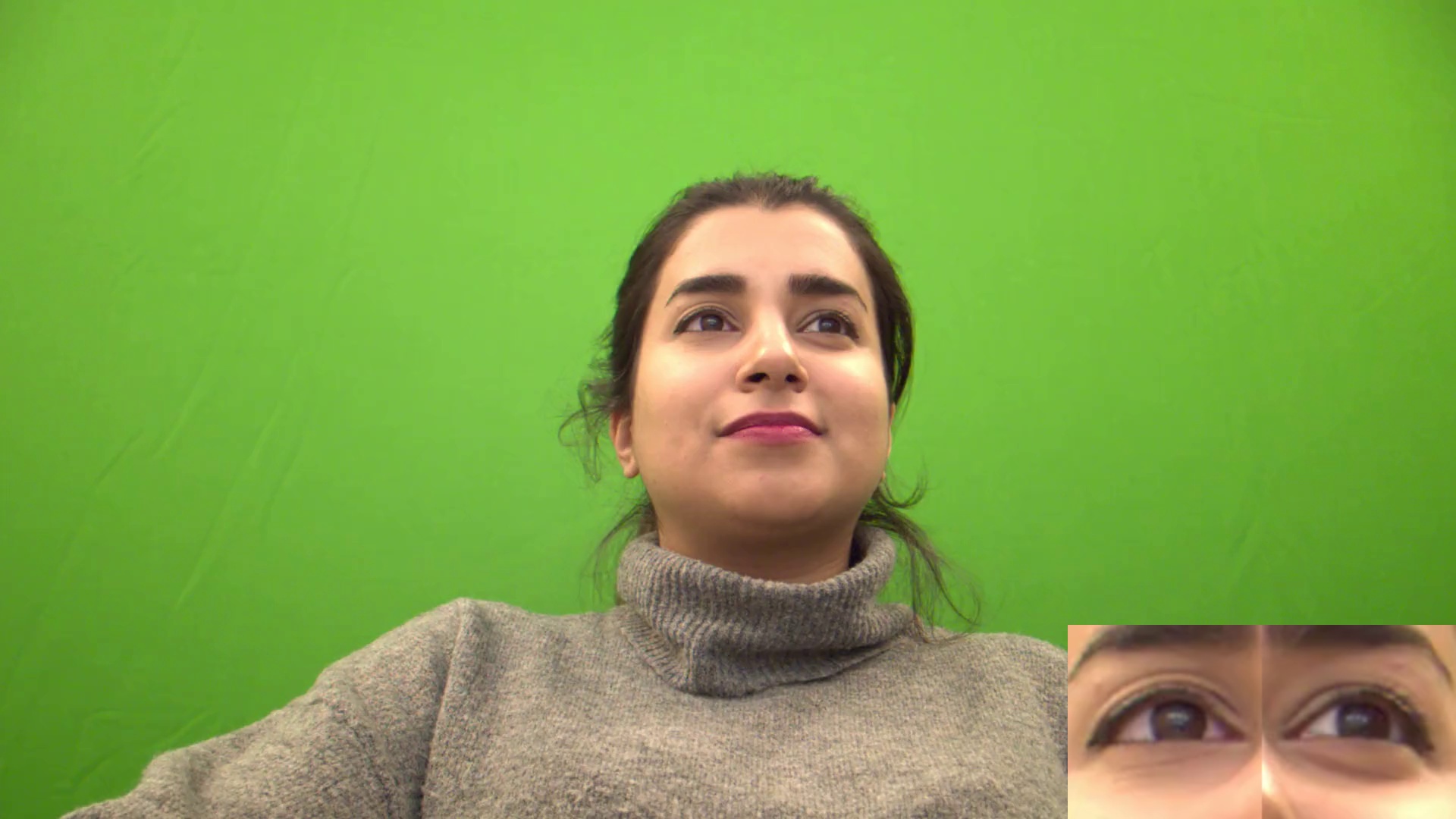}
            {\tiny Machine Vision Camera}
        \end{minipage}
        \hfill
        \begin{minipage}[t]{0.48\columnwidth}\centering
            \includegraphics[width=\columnwidth]{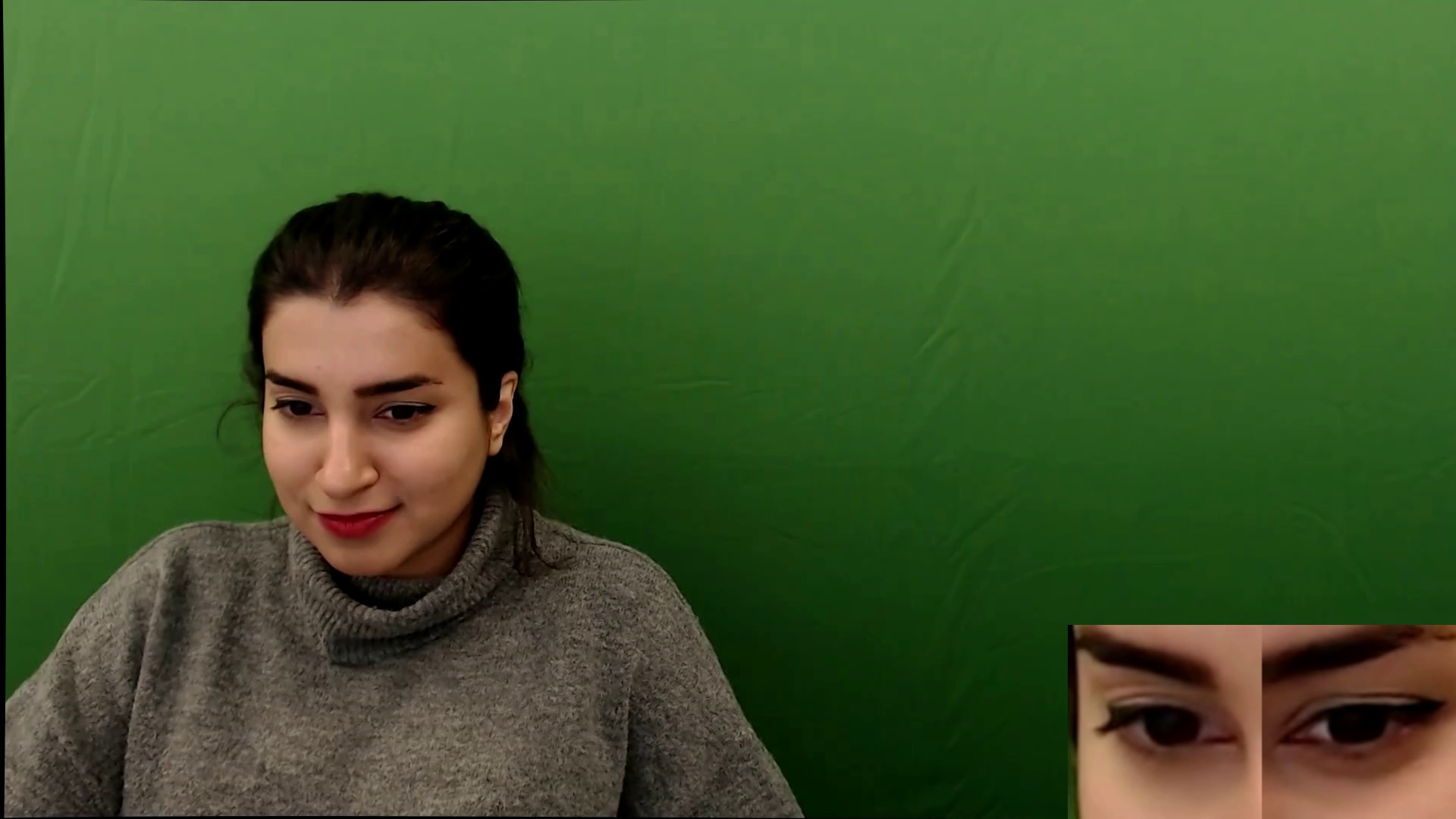}
            {\tiny Webcam (Left)}
        \end{minipage}
        \\[2mm]
        \begin{minipage}[t]{0.48\columnwidth}\centering
            \includegraphics[width=\columnwidth]{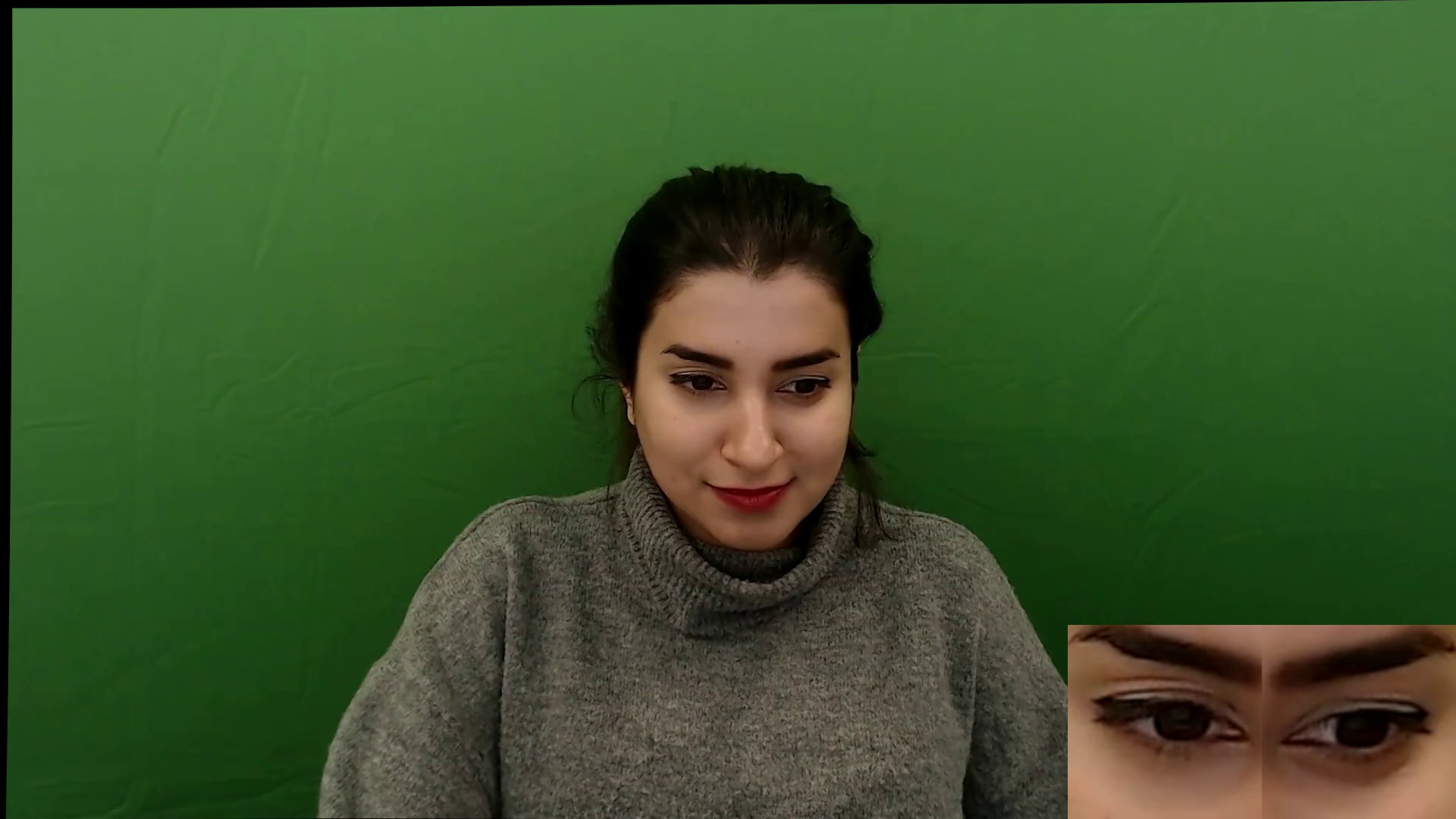}
            {\tiny Webcam (Center)}
        \end{minipage}
        \hfill
        \begin{minipage}[t]{0.48\columnwidth}\centering
            \includegraphics[width=\columnwidth]{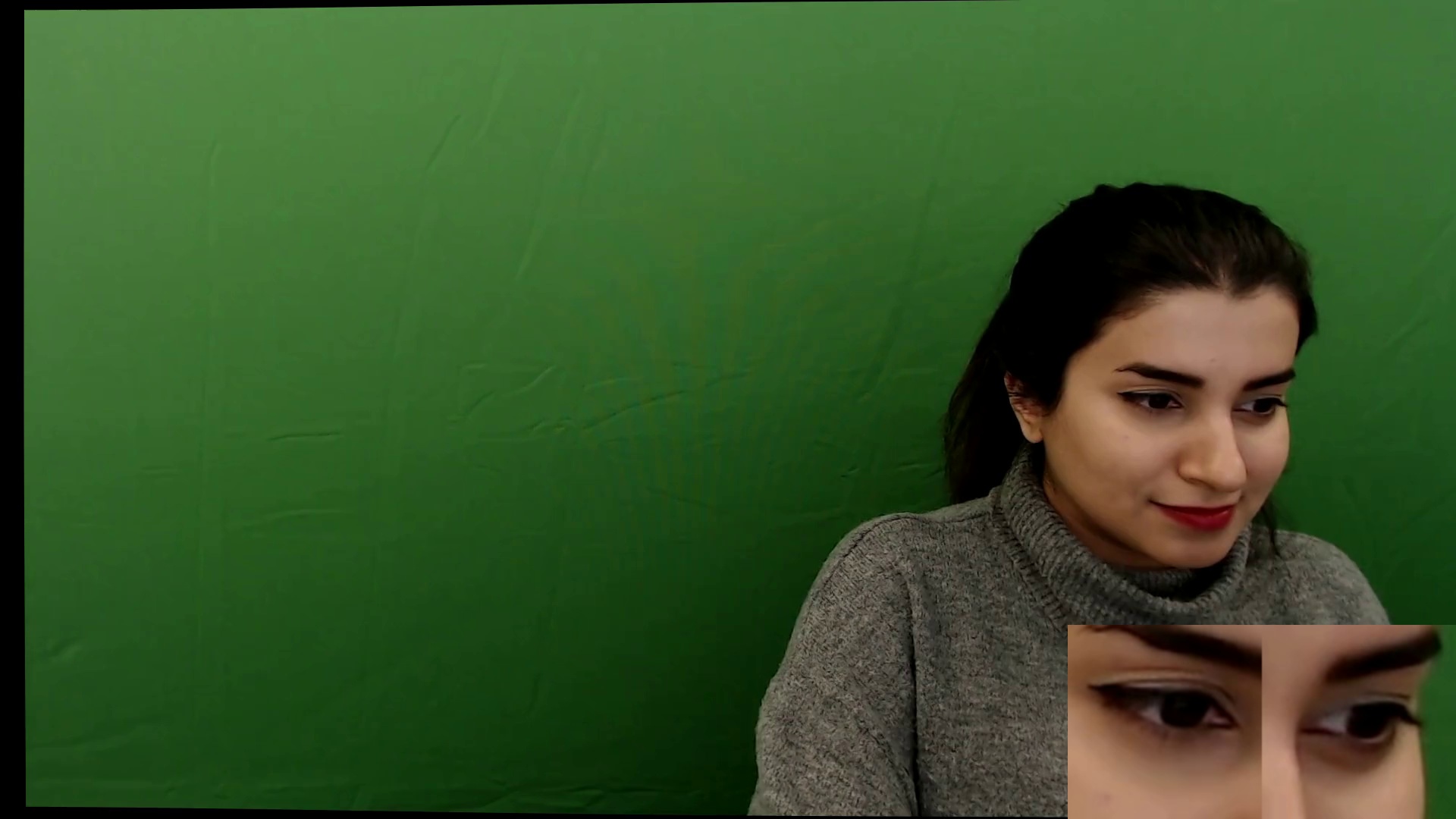}
            {\tiny Webcam (Right)}
        \end{minipage}
        \caption{Sample frames (each 1080p)}\label{fig:sample_frames}
    \end{subfigure}
    \caption{\datasetname data collection setup and example of (undistorted) frames collected from the 4 camera views with example eye patches shown as insets.}
\end{figure}

To study the semantic relations and temporal dynamics between eye gaze and visual content,
we identify a need for a new gaze dataset that:
\begin{enumerate}
    \item allows for the training and evaluation of temporal models on natural eye movements (including fixations, saccades, and smooth pursuits),
    \item enables the training of models that can process full camera frame inputs to yield screen-space Point-of-Gaze (PoG) estimates,
    \item and provide a community-standard benchmark for a good understanding of the generalization capabilities of upcoming methods.
\end{enumerate}

Furthermore, we consider the fact that the distribution of visual saliency on a computer screen at a given time is indicative of likely gaze positions. In line with this observation, prior work reports difficulty in generalization when considering saliency estimation and gaze estimation as separate components \cite{Sugano2010CVPR,Sugano2015UIST}. Thus, we define following further requirements for our new dataset:
\begin{enumerate}
    \item a video of the screen content synchronized with eye gaze data,
    \item a sufficiently large set of visual stimuli must be presented to allow for algorithms to generalize better without over-fitting to a few select stimuli,
    \item and lastly, gaze data must be collected over time without instructing participants to gaze at specific pin-point targets such that they act naturally, like behaviours in a real-world setting.
\end{enumerate}

We present in this section the methodologies we adopt to construct such a dataset, and briefly describe its characteristics. We call our proposed dataset \emph{``\datasetname''}, which stands for \emph{``a dataset for enabling progress towards truly \textbf{E}nd-to-end \textbf{V}ideo-based \textbf{E}ye-tracking algorithms''}.

\subsection{Captured Data}

The minimum requirements for constructing our proposed dataset is the captured video from a webcam, gaze ground truth data from a commercial eye tracker, and screen frames from a given display.
Furthermore, we:
\begin{itemize}
    \item use the Tobii Pro Spectrum eye tracker, which reports high accuracy and precision in predicted gaze\footnote{See \scriptsize https://www.tobiipro.com/pop-ups/accuracy-and-precision-test-report-spectrum/?v=1.1} even in the presence of natural head movements,
    \item add a high performance Basler Ace acA1920-150uc machine vision camera with global shutter, running at 60Hz,
    \item install three Logitech C922 webcams (30Hz) for a wider eventual coverage of head orientations, assuming that the final user will not only be facing the screen in a fully-frontal manner (see Fig.~\ref{fig:sample_frames}),
    \item and apply MidOpt BP550 band-pass filters to all webcams and machine vision camera to remove reflections and glints on eyeglass and cornea surfaces due to the powerful near-infra-red LEDs used by the Tobii eye tracker.
\end{itemize}

All video camera frames are captured at $1920\times 1080$ pixels resolution, but the superior timestamp-reliability and image quality of the Basler camera is expected to yield better estimates of gaze compared to the webcams.

The data captured by the Tobii Pro Spectrum eye tracker can be of very high quality which
is subject to participant and environment effects.
Hence to ensure data quality and reliability, an experiment coordinator is present during every data collection session to qualitatively assess eye tracking data via a live-stream of camera frames and eye movements.
Additional details on our hardware setup and steps we take to ensure the best possible eye tracking calibration and subsequent data quality are described in the supplementary materials.

\subsection{Presented Visual Stimuli}

A large variety of visual stimuli are presented to our participants. Specifically, we present image, video, and wikipedia page stimuli (shown later in Fig.~\ref{fig:qualitative}).

For static image stimuli, we select the widely used MIT1003 dataset \cite{Judd2009ICCV} originally created for the task of image saliency estimation. 
Most images in the dataset span $1024$ pixels in either horizontal or vertical dimensions. We randomly scale the image between 1320 and 1920 pixels in width or 480 to 1080 pixels in height, to be displayed on our 25-inch screen (with a resolution of 1080p).

All video stimuli are displayed in 1080p resolution (to span the full display),
and taken from the DIEM~\cite{Mital2011}, VAGBA~\cite{Li2011IVC}, and Kurzhals~\etal~\cite{Kurzhals2014} datasets. These datasets consist of 720p, 1080p, and 1080p videos respectively, and thus are of high-resolution compared to other video-based saliency datasets. 
DIEM consists of various videos sampled from public repositories such as trailers and documentaries. VAGBA includes human movement or interactions in everyday scenes, and Kurzhals~\etal contain purposefully designed video sequences with intentionally-salient regions. To increase the variety of the final set of video stimuli further, we select 23 videos from Wikimedia (at 1080p resolution).

Wikipedia pages are randomly selected on-the-fly by opening the following link in a web browser: \url{https://en.m.wikipedia.org/wiki/Special:Random#/random} and participants are then asked to freely view and navigate the page, as well as to click on links. Links leading to pages outside of Wikipedia are automatically removed using the GreaseMonkey web browser extension.  

In our data collection study, we randomly sample the image and video stimuli from the mentioned datasets. We ensure that each participant observes 60 image stimuli (for three seconds each), at least 12 minutes of video stimuli, and six minutes of wikipedia stimulus (three 2-minute sessions). At the conclusion of data collection, we found that each image stimulus has been observed $3.35$ times ($SD=0.73$), and each video stimulus has been observed $9.36$ times ($SD=1.28$).

\begin{figure}[t]
    \centering
    \rotatebox[origin=c]{90}{\scriptsize\hspace{4mm}Gaze Direction\hspace{13mm}Head Pose}
    \,
    \begin{subfigure}{0.25\columnwidth}
        \centering
        \includegraphics[width=0.9\columnwidth]{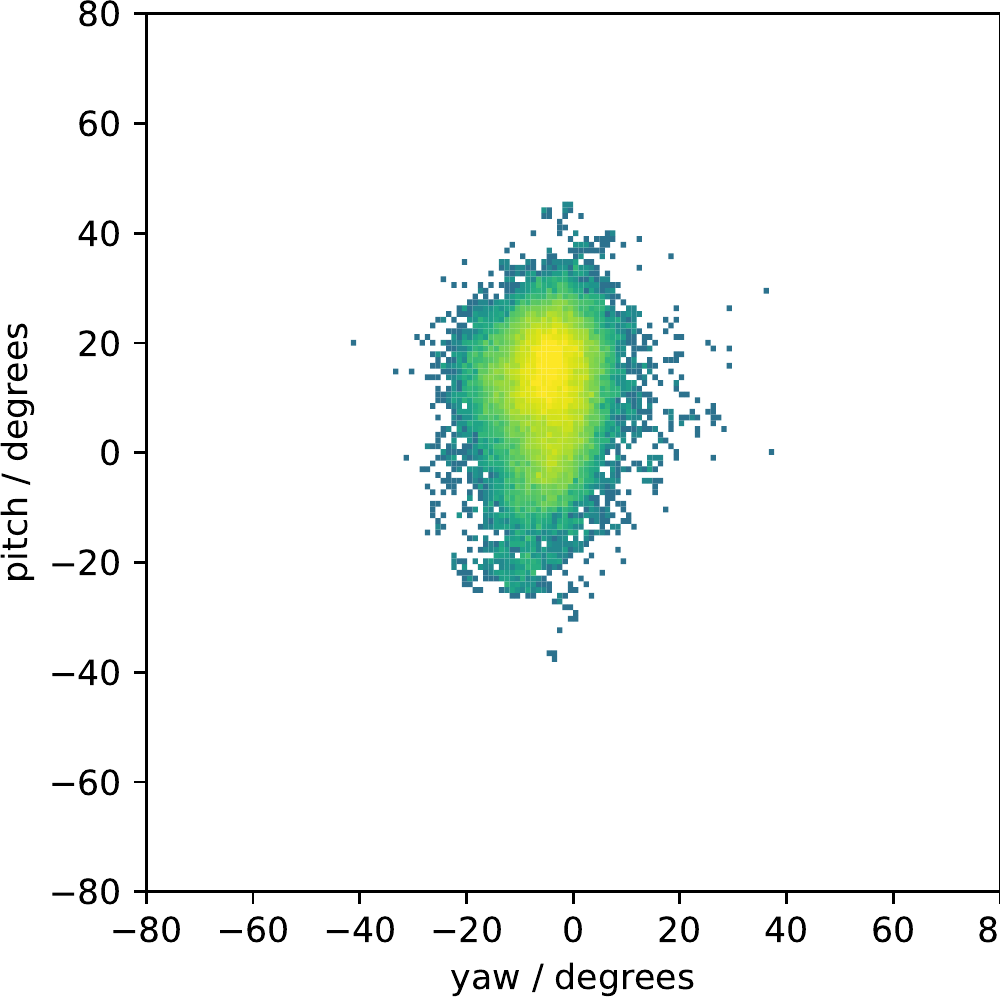} \\[1mm]
        \includegraphics[width=0.9\columnwidth]{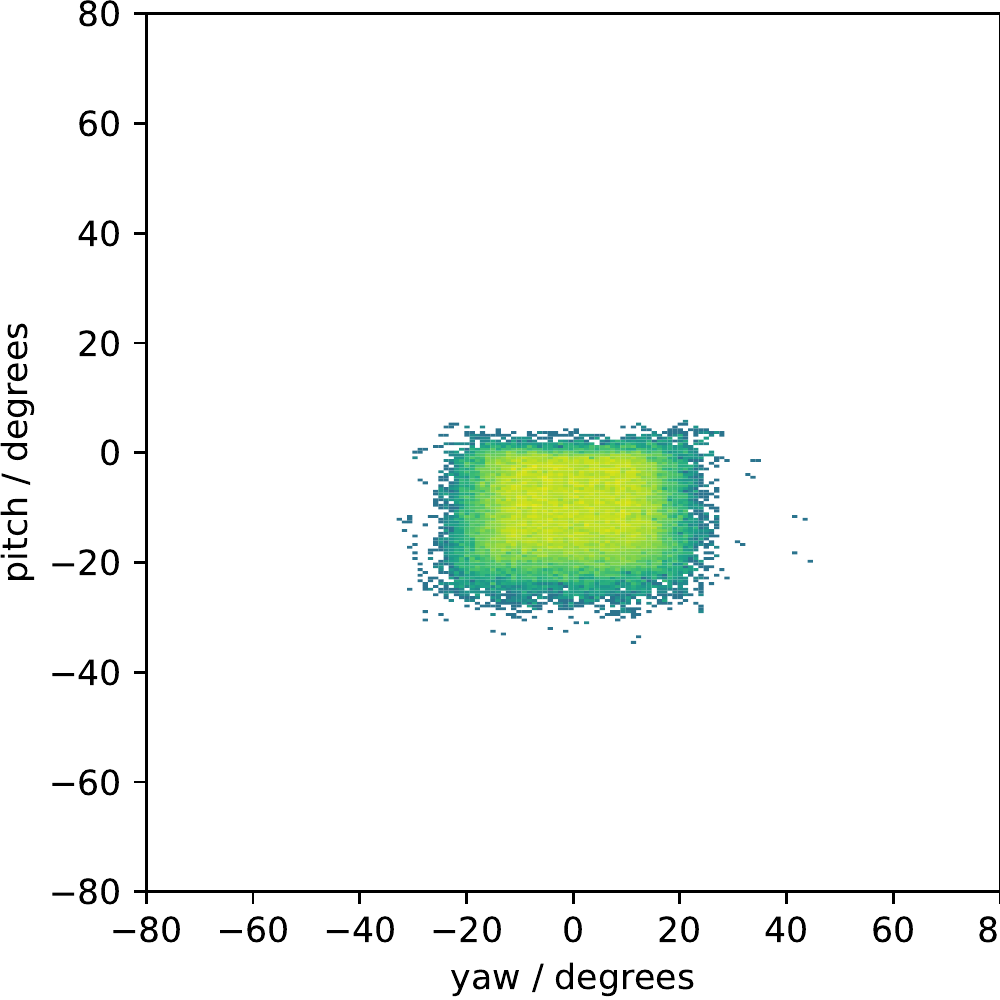} \\[1mm]
        \hspace{3mm}MPIIFaceGaze~\cite{Zhang2017CVPRW}
    \end{subfigure}
    \begin{subfigure}{0.25\columnwidth}
        \centering
        \includegraphics[width=0.9\columnwidth]{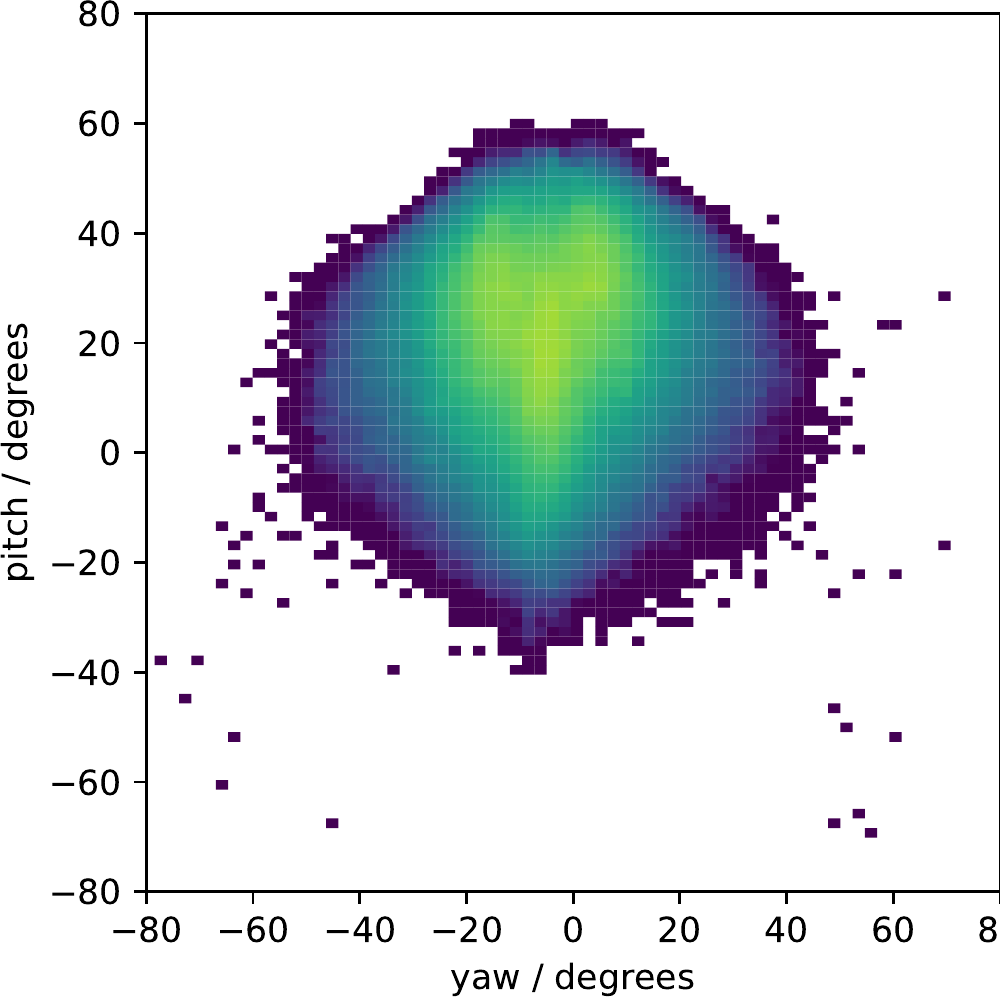} \\[1mm]
        \includegraphics[width=0.9\columnwidth]{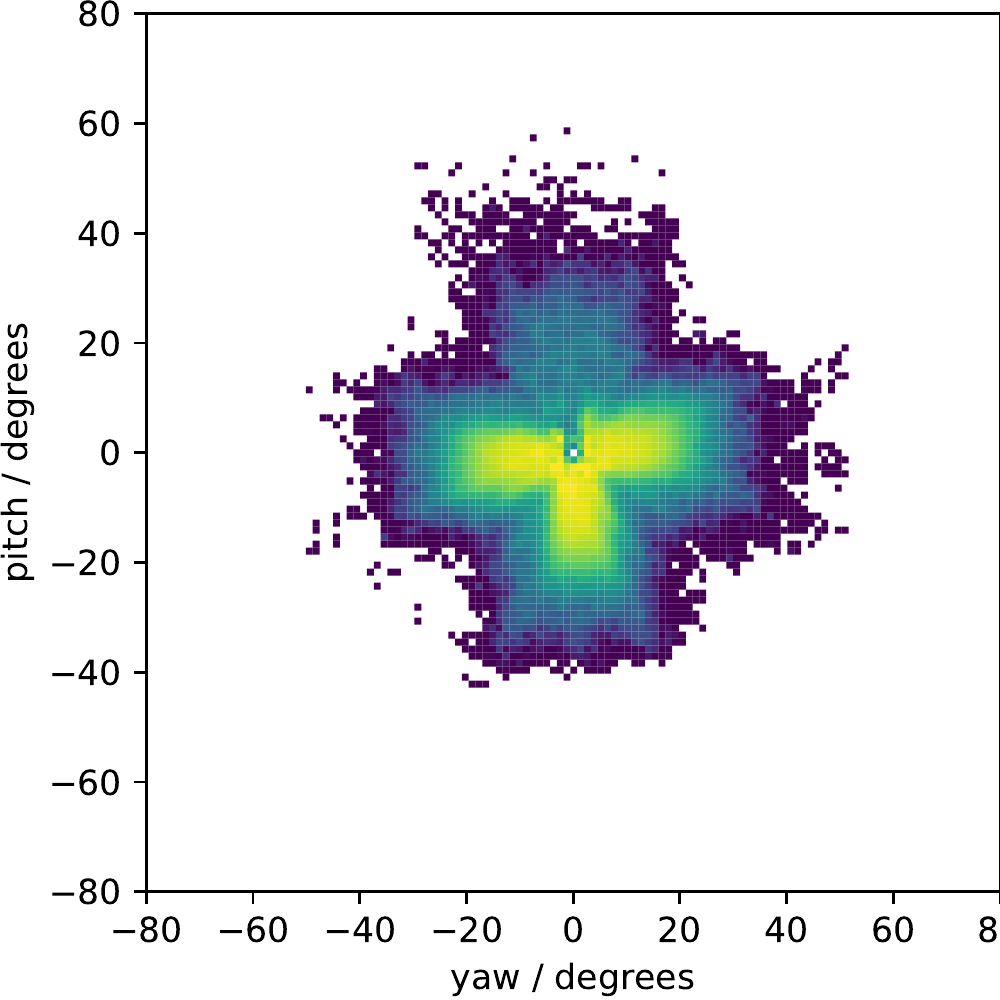} \\[1mm]
        \hspace{4mm}GazeCapture~\cite{Krafka2016CVPR}
    \end{subfigure}
    \begin{subfigure}{0.25\columnwidth}
        \centering
        \includegraphics[width=0.9\columnwidth]{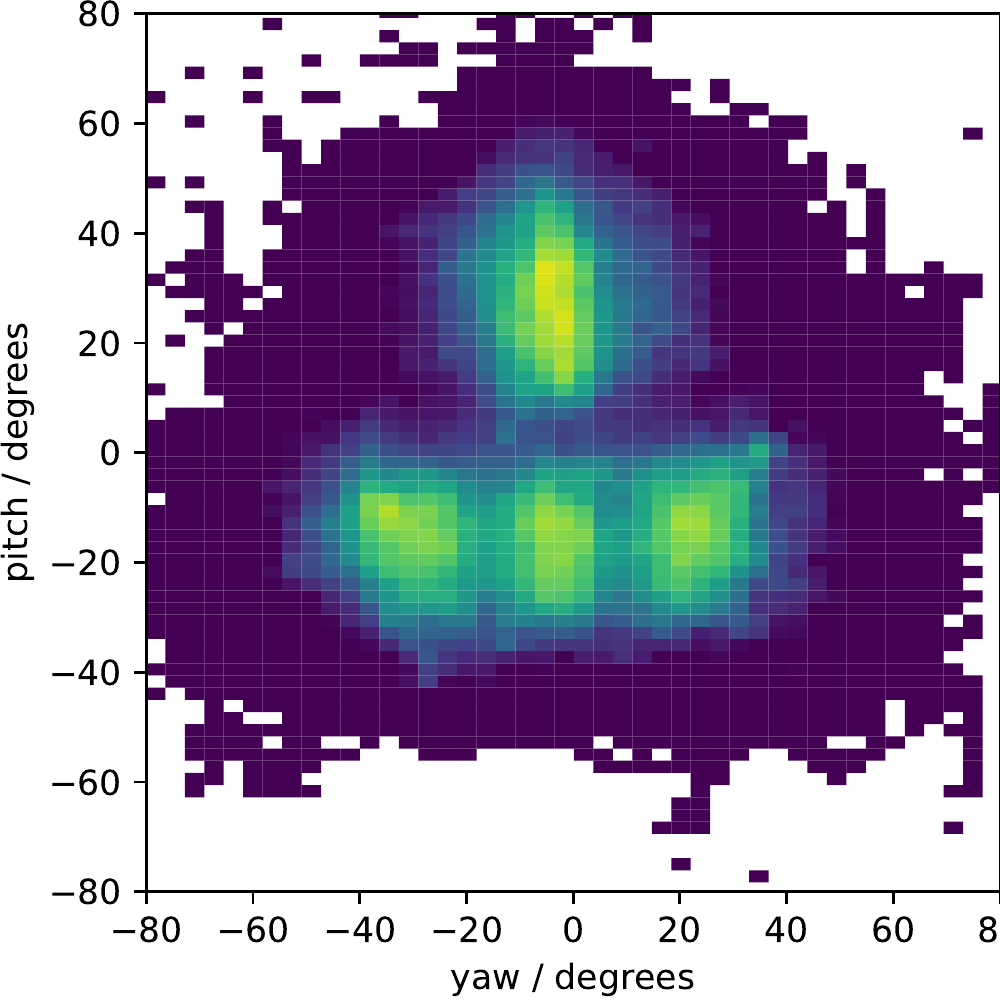} \\[1mm]
        \includegraphics[width=0.9\columnwidth]{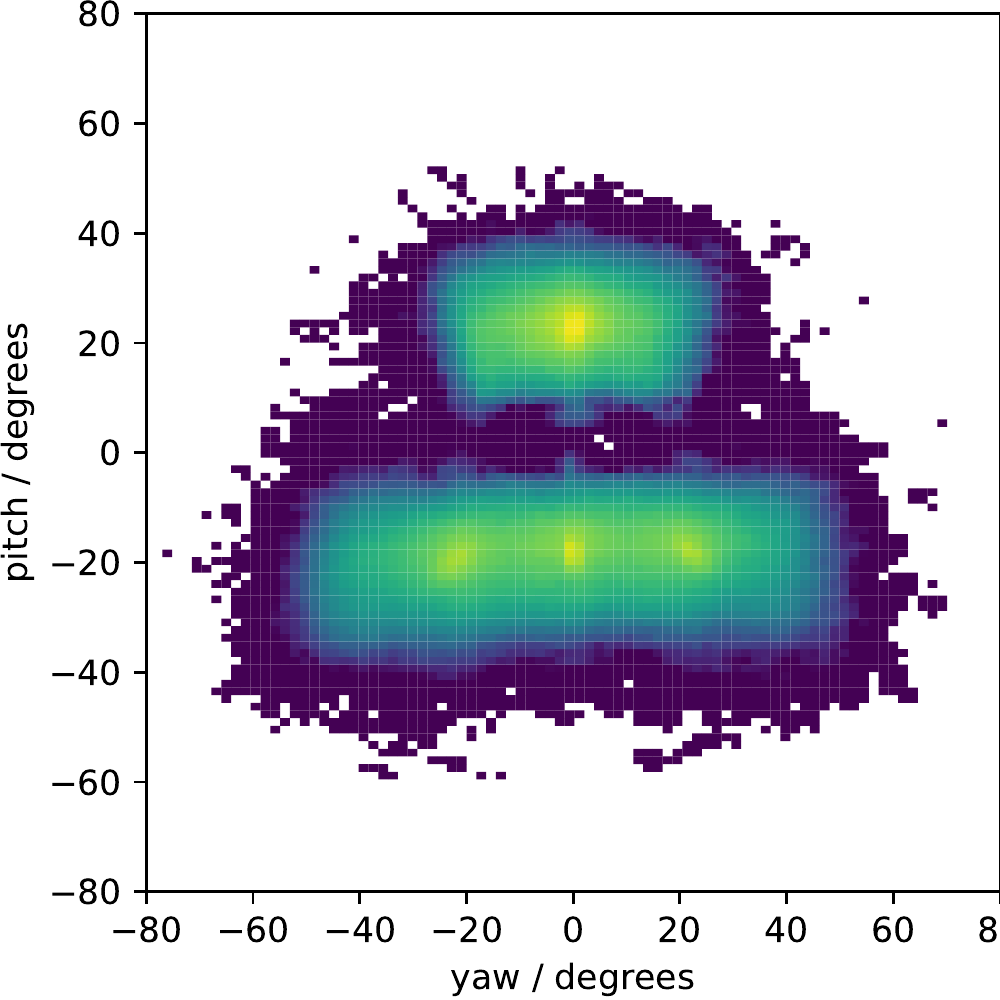} \\[1mm]
        \hspace{3mm}Ours~(\datasetname)
    \end{subfigure}
    \hspace{6mm}
    \\[-1mm]
    \caption{Head orientation and gaze direction distributions are compared with existing screen-based gaze datasets \cite{Zhang2017CVPRW,Krafka2016CVPR}. We capture a larger range of parameter space due to a multi-view camera setup and 25-inch display.
    2D histogram plot values are normalized and colored with log-scaling.
    }
    \label{fig:distributions}
\end{figure}

\subsection{Dataset Characteristics}

The final dataset is collected from 54 participants (30 male, 23 female, 1 unspecified). The details of responses to our demographics questionnaire can be found in our supplementary materials along with how we pre-process the dataset.
We ensure that the subjects in both training and test sets exhibit diverse gender, age, and ethnicity, some with and some without glasses.

In terms of gaze direction and head orientation distributions, \datasetname compares favorably to popular screen-based datasets such as MPIIFaceGaze~\cite{Zhang2017CVPRW} and GazeCapture \cite{Krafka2016CVPR}. Figure~\ref{fig:distributions} shows that we cover a larger set of gaze directions and head poses. 
This is likely due to the 4 camera views that we adopt, together with a large screen size of 25 inches (compared to the other datasets).

 \section{Method}
\begin{figure}[t]
    \centering
    \includegraphics[width=\columnwidth]{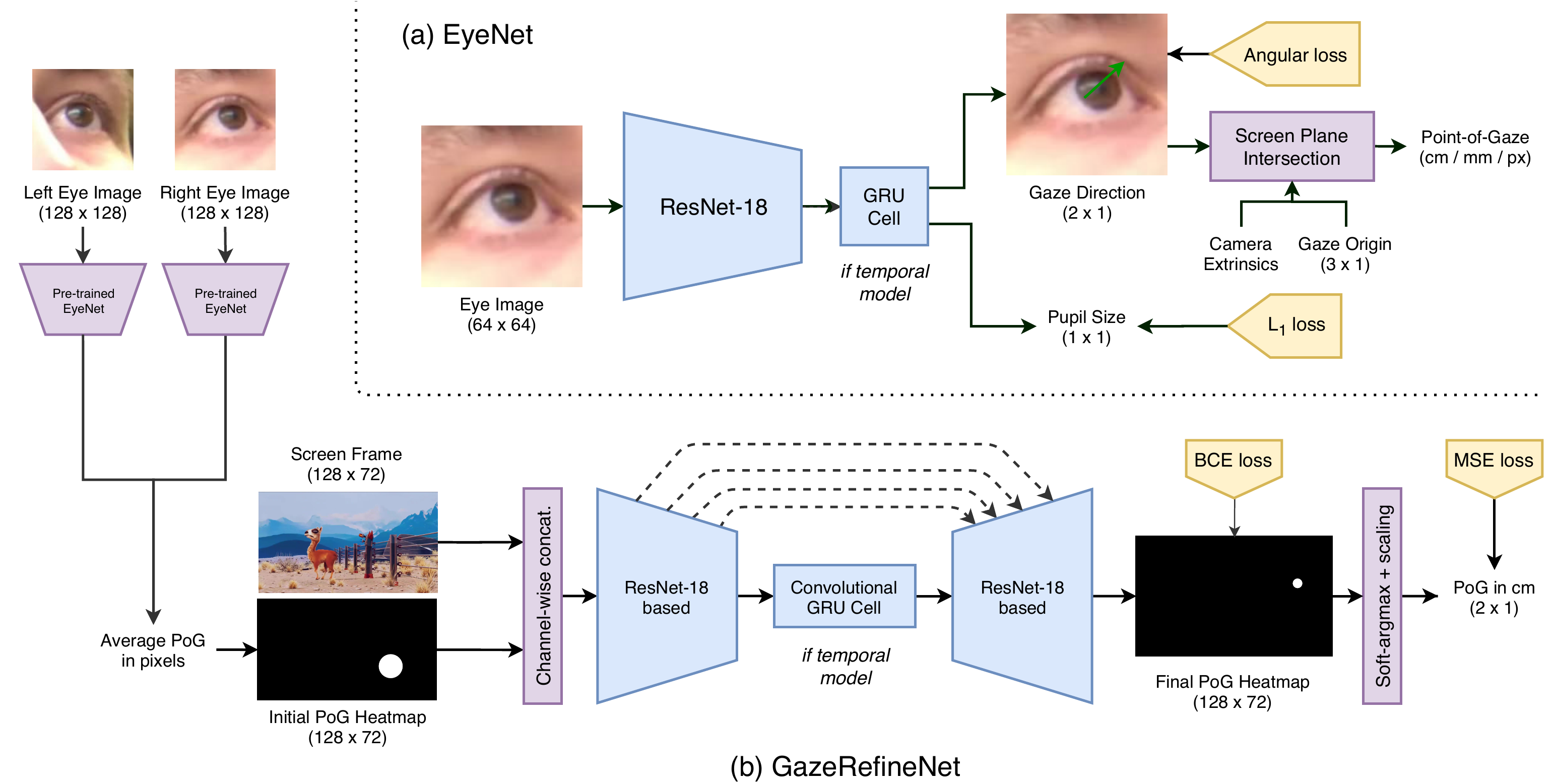}
    \caption{We adopt (a) a simple \eyenetname architecture for gaze direction and pupil size estimation with an optional recurrent component, and propose (b) a novel \refinenetname architecture for label-free PoG refinement using screen content. 
    }
    \label{fig:architectures}
\end{figure}
We now discuss a novel architecture designed to exploit the various sources of information in datasets and to serve as baseline for follow-up work. 
We first introduce a simple eye gaze estimation network (\eyenetstaticname) and its recurrent counterparts (\eyenetrnnname, \eyenetlstmname, \eyenetgruname) for the task of per-frame or temporal gaze and pupil size estimation (see Fig.~\ref{fig:architectures}a).
As the \datasetname dataset contains synchronized visual stimuli, we propose a novel technique to process these initial eye-gaze predictions further by taking the raw screen content directly into consideration. To this end, we propose the \refinenetname architecture (Fig.~\ref{fig:architectures}b), and describe its details in the second part of this section.

\subsection{\eyenetname Architecture}

Learning-based eye gaze estimation models typically output their predictions as a unit direction vector or in Euler angles in the spherical coordinate system. The common metric to evaluate the goodness of predicted gaze directions is via an angular distance error metric in degrees. Assuming that the predicted gaze direction is represented by a 3-dimensional unit vector $\bvh{g}$, the calculation of the angular error loss when given ground-truth $\bv{g}$ is then:
\begin{equation}
    \mathcal{L}_\mathrm{gaze}\left(\bv{g},\,\bvh{g}\right) =
    \frac{1}{NT}\sum^N\sum^T
    \frac{180}{\pi}
    \arccos\left(\frac{\bv{g}\cdot\bvh{g}}{\|\bv{g}\|\|\bvh{g}\|}\right)
\end{equation}
where a mini-batch consists of $N$ sequences each of length $T$.

To calculate PoG, the predicted gaze direction must first be combined with the 3D gaze origin position $\bv{o}$ (determined during data pre-processing), yielding a gaze ray with 6 degrees of freedom.
We can then intersect this ray with the screen plane to calculate the PoG by using the camera transformation with respect to the screen plane.
Pixel dimensions (our $1920\times 1080$ screen is 553mm wide and 311mm tall) can be used to convert the PoG to pixel units for an alternative interpretation.
We denote the predicted PoG in centimeters as $\bvh{s}$.

Assuming that the pupil size can be estimated, we denote it as $\bvh{p}$ and define an $\ell_1$ loss given ground-truth $\bv{p}$ as:
\begin{equation}
    \mathcal{L}_\mathrm{pupil}\left(\bv{p},\,\bvh{p}\right) = 
    \frac{1}{NT}\sum^N\sum^T
    {\|\bv{p}-\bvh{p}\|}_{1}
\end{equation}

The two values of gaze direction and pupil size are predicted by a ResNet-18 architecture \cite{He2015ICCV}. To make the network recurrent, we optionally incorporate a RNN \cite{Sutskever2014NeurIPS}, LSTM \cite{Hochreiter1997}, or GRU \cite{Chung2014NeurIPSW} cell.

\subsection{\refinenetname Architecture}\label{sec:refinenet}
Given the left and right eye images $\bv{x}_l$ and $\bv{x}_r$ of a person, we hypothesize that incorporating the corresponding screen content can improve the initial PoG estimate. 
Provided that an initial estimate of PoG $\tilde{\bv{s}} = f\left(\bv{x}\right)$ can be made for the left and right eyes $\tilde{\bv{s}}_l$ and $\tilde{\bv{s}}_r$ respectively, we first take the average of the predicted PoG values with $\tilde{\bv{s}}=\frac{1}{2}\left(\tilde{\bv{s}}_l+\tilde{\bv{s}}_r\right)$ to yield a single estimate of gaze. Here $f$ denotes the previously described \eyenetname.
We define and learn a new function, $\bv{s} = g\left(\bv{x}_S, \tilde{\bv{s}} \right)$, to refine the \eyenetname predictions by incorporating the screen content and temporal information. The function $g$ is parameterized by a fully convolutional neural network (FCN) to best preserve spatial information. Following the same line of reasoning, we represent our initial PoG estimate $\tilde{\bv{s}}$ as a confidence map. More specifically, we use an isotropic 2D Gaussian function centered at the estimated gaze position on the screen. The inputs to the FCN are concatenated channel-wise.

To allow the model to better exploit the temporal information, we use an RNN cell in the bottleneck. Inspired by prior work in video-based saliency estimation, we adopt a convolutional recurrent cell \cite{Linardos2019BMVC} and evaluate RNN \cite{Sutskever2014NeurIPS}, LSTM \cite{Hochreiter1997}, and GRU \cite{Chung2014NeurIPSW} variants.

The network optionally incorporate concatenative skip connections between the encoder and decoder layers, as this is shown to be helpful in FCNs.
We train the \refinenetname by using pixel-wise binary cross-entropy loss on the output heatmap and MSE loss on the final numerical estimate of the PoG. It is calculated in a differentiable manner via a soft-argmax layer \cite{Chapelle2010,Honari2018CVPR}. The PoG is converted to centimeters to keep the loss term from exploding (due to its magnitude).
Please refer to Fig.~\ref{fig:architectures}b for the full architecture diagram, and our supplementary materials for implementation details.

\subsubsection{Offset augmentation}
In the task of cross-person gaze estimation, it is common to observe high discrepancies between the training and validation objectives.
This is not necessarily due to overfitting or non-ideal hyperparameter selections but rather due to the inherent nature of the problem.
Specifically, every human has a person-specific offset between their optical and visual axes in each eye, often denoted by a so-called Kappa parameter.
While the optical axis can be observed by the appearance of the iris, the visual axis cannot be observed at all as it is defined by the position of the fovea at the back of the eyeball.

During training, this offset is absorbed into the neural network's parameters, limiting generalization to unseen people.
Hence, prior work typically incur a large error increase in cross-person evaluations ($\sim5^\circ$) in comparison to person-specific evaluations ($\sim3^\circ$).
Our insight is that we are now posing a gaze refinement problem, where an initially incorrect assessment of offset could actually be corrected by additional signals such as that of screen content.
This is in contrast with the conventional setting, where no such corrective signal is made available.
Therefore, the network should be able to learn to overcome this offset when provided with randomly sampled offsets to a given person's gaze.

This randomization approach can intuitively be understood as learning to undo all possible inter-personal differences rather than learning the corrective parameters for a specific user, as would be the case in traditional supervised personalization (e.g., \cite{Park2019ICCV}).  
We dub our training data augmentation approach as an \emph{``offset augmentation''}, and provide further details of its implementation in our supplementary materials.

 \section{Results}

In this section, we evaluate the variants of \eyenetname and find that temporal modelling can aid in gaze estimation. Based on a pre-trained \eyenetgruname, we then evaluate the effects of our contributions in refining an initial estimate of PoG using variants of \refinenetname. We demonstrate large and consistent performance improvements even across camera views and visual stimulus types.

\subsection{Eye Gaze Estimation}

We first consider the task of eye gaze estimation purely from a single eye image patch.
Tab.~\ref{tab:initial_gaze_errors} shows the performance of the static \eyenetstaticname and its temporal variants (\eyenetrnnname, \eyenetlstmname, \eyenetgruname) on predicting gaze direction, PoG, and pupil size. The networks are trained on the training split of \datasetname.
Generally, we find our gaze direction error values to be in line with prior works in estimating gaze from single eye images \cite{Zhang2015CVPR}, and see that the addition of recurrent cells improve gaze estimation performance modestly.
This makes a case for training gaze estimators on temporal data, using temporally-aware models, and corroborates observations from a prior learning-based gaze estimation approach on natural eye movements \cite{Wang2019CVPR_Temporal}.

Pupil size errors are presented in terms of mean absolute error. Considering that the size of pupils in our dataset vary from 2mm to 4mm, the presented errors of $0.3$mm should allow for meaningful insights to be made in fields such as the cognitive sciences. 
We select the GRU variant (\eyenetgruname) for the next steps as it shows consistently good performance for both eyes.

\begin{table}[t]
    \centering
    \caption{Cross-person gaze estimation and pupil size errors of \eyenetname variants, evaluated on the test set of \datasetname.
    The GRU variant performs best in terms of both gaze and pupil size estimates 
    }
    \label{tab:initial_gaze_errors}
    \renewcommand{\arraystretch}{1.2}
    \renewcommand\theadset{\def\arraystretch{1.2}}
    \begin{tabular}{|l|c|c|c|c|c|c|c|c|}
        \hline
        \multirow{2}{*}{Model} & 
        \multicolumn{4}{c|}{Left Eye} & 
        \multicolumn{4}{c|}{Right Eye} \\
        \cline{2-9}
        & \thead{Gaze Dir.\\[-1mm]\scriptsize($^\circ$)} & \thead{\enspace PoG\enspace\\[-1mm]\scriptsize(cm)} & \thead{PoG\\[-1mm]\scriptsize(px)} & \thead{Pupil Size\\[-1mm]\scriptsize(mm)} 
        & \thead{Gaze Dir.\\[-1mm]\scriptsize($^\circ$)} & \thead{\enspace PoG\enspace\\[-1mm]\scriptsize(cm)} & \thead{PoG\\[-1mm]\scriptsize(px)} & \thead{Pupil Size\\[-1mm]\scriptsize(mm)} \\
        \hline
        \eyenetstaticname     
        &         4.54  &         5.10  &            172.7    &  \textbf{0.29} 
        &         4.75  &         5.29  &            181.0    &  \textbf{0.29} \\
        \eyenetrnnname      
        &         4.33  &         4.86  &            166.7    &  \textbf{0.29} 
        &         4.91  &         5.48  &            186.5    &  \textbf{0.28} \\
        \eyenetlstmname     
        &         4.17  &         4.66  &            161.0    &          0.32  
        & \textbf{4.71} & \textbf{5.25} &  \,\textbf{180.5}\, &          0.33  \\
        \eyenetgruname  
        & \textbf{4.11} & \textbf{4.60} &  \,\textbf{158.5}\, &  \textbf{0.28} 
        &         4.80  &         5.33  &            183.9    &  \textbf{0.29} \\
        \hline
    \end{tabular}
\end{table}

\begin{table}[t]
    \centering
    \caption{An ablation study of our contributions in \refinenetname, where a frozen and pre-trained \eyenetgruname is used for initial gaze predictions. 
    Temporal modelling and our novel offset augmentation both yield large gains in performance.
    }
    \label{tab:temporal_ablation}
    \renewcommand{\arraystretch}{1.2}
    \begin{tabular}{|lccc|c|c|c|}
        \hline
        Model & 
        \thead{Screen\\[-1mm]Content} & 
        \thead{Offset\\[-1mm]Augmen.} & 
        \thead{Skip\\[-1mm]Conn.} & 
        Gaze Dir. ($^\circ$) & PoG (cm) & PoG (px) \\
        \hline
        \hline
        Baseline (\eyenetgruname)       
        &&&&            3.48  &         3.85  &        132.56  \\
        \hline
        &o&&&           3.33  &         3.67  &        127.59  \\[-1mm]
        &o&o&&          2.80  &         3.09  &        107.42  \\[-1mm]
        \refinenetstaticname
        &o&o&o&         2.87  &         3.16  &        109.85  \\
        \hline       
        &o&o&&          2.67  &         2.95  &        102.36  \\[-1mm]
        \refinenetrnnname
        &o&o&o&         2.57  &         2.83  &         98.38  \\
        \hline       
        &o&o&&  \textbf{2.49} & \textbf{2.75} & \textbf{95.43} \\[-1mm]
        \refinenetlstmname
        &o&o&o&         2.53  &         2.79  &         96.97  \\
        \hline        
        &o&o&&          2.51  &         2.77  &         96.24  \\[-1mm]
        \refinenetgruname
        &o&o&o& \textbf{2.49} & \textbf{2.75} & \textbf{95.59} \\
        \hline
    \end{tabular}
\end{table}

\subsection{Screen Content based Refinement of PoG}

\refinenetname consists of a fully-convolutional architecture which takes as input a screen content frame, and an offset augmentation procedure at training time. 
Our baseline performance for this experiment is different to Tab.~\ref{tab:initial_gaze_errors} as gaze errors are improved when averaging the PoG from the left and right eyes, with according adjustments to the label (averaged in screen space).
Even with the new competitive baseline from PoG averaging, we find in Tab.~\ref{tab:temporal_ablation} that each of our additional contributions yield large performance improvements, amounting to a 28\% improvement in gaze direction error, reducing it to $2.49^\circ$. 
While not directly comparable due to differences in setting, this value is lower even than recently reported performances of supervised few-shot adaptation approaches on in-the-wild datasets \cite{Park2019ICCV,Linden2019ICCVW}.
Specifically, we find that the offset augmentation procedure yields the greatest performance improvements, with temporal modeling further improving performance.
Skip connections between the encoder and decoder do not necessarily help (except in the case of \refinenetrnnname), presumably because the output relies mostly on information processed at the bottleneck.
We present additional experiments of \refinenetname in the following paragraphs, and describe their setup details in our supplementary materials.
\\

\begin{table}[t]
    \centering
    \caption{Improvement in PoG prediction (in px) of our method in comparison with two saliency-based alignment methods, as evaluated on the EVE dataset.}
    \label{tab:saliency_results}
    \renewcommand{\arraystretch}{1.2}
    \begin{tabular}{|l|c|c|c|}
        \hline
        \diagbox{Method}{Stimulus Type} & Image & Video & Wikipedia \\
        \hline
        Saliency-based (scale + bias) & \result{78.4}{36.3} & \result{116.7}{12.0} & \result{198.3}{-43.6} \\
        Saliency-based (kappa) & \result{75.0}{39.2} & \result{110.9}{17.0} & \result{258.0}{-84.4} \\
        \refinenetgruname (Ours) & \resultb{48.7}{60.4} & \resultb{96.7}{27.1} & \resultb{116.3}{15.8} \\
        \hline
    \end{tabular}
\end{table}

\begin{table}[t]
    \caption{Final gaze direction errors (in degrees, lower is better) from the output of \refinenetgruname, evaluated on the \datasetname test set in cross-stimuli settings. Indicated improvements are with respect to initial PoG predictions (mean of left+right) from \eyenetgruname trained on specified source stimuli types.}
    \label{tab:cross_stimuli_refinement}
    \centering
    \renewcommand{\arraystretch}{1.3}
    \begin{tabular}{|l|l|l|l|}
       \hline
       \diagbox{Source}{Target} & Images & Videos & Wikipedia \\
       \hline
        Images    & \resultb{1.30}{60.55} & \result{3.60}{-4.10}   & \result{4.74}{-30.13}  \\
        Videos    & \result{1.97}{40.09}  & \resultb{2.60}{24.88} & \result{3.71}{-1.94}   \\
        Wikipedia & \result{2.12}{35.75}  & \result{3.32}{3.84}   & \resultb{3.04}{16.62} \\
       \hline
    \end{tabular}
\end{table}

\noindent\textbf{Comparison to Saliency-based Methods}.
In order to assess how our \refinenetname approach compares with existing saliency-based methods, we implement two up-to-date methods loosely based on \cite{Alnajar2013ICCV} and \cite{Wang2016ETRA}.
First, we use the state-of-the-art UNISAL approach \cite{drostejiao2020eccv} to attain high quality visual saliency predictions.
We accumulate these predictions over time for the full exposure duration of each visual stimulus in \datasetname (up to 2 minutes), which should provide the best context for alignment (as opposed to our online approach, which is limited to 3 seconds of history).
Standard back propagation is then used to optimize for either scale and bias in screen-space (similar to \cite{Alnajar2013ICCV}) or the visual-optical axis offset, kappa (similar to \cite{Wang2016ETRA}) using a KL-divergence objective between accumulated visual saliency predictions and accumulated heatmaps of refined gaze estimates in the screen space.
Tab.~\ref{tab:saliency_results} shows that while both saliency-based baselines perform respectably on the well-studied image and video stimuli, they fail completely on wikipedia stimuli despite the fact that the saliency estimation model was provided with full 1080p frames (as opposed to the $128\times 72$ input used by \refinenetgruname).
Furthermore, our direct approach takes raw screen pixels and gaze estimations up to the current time-step as explicit conditions and thus is a simpler yet explicit solution for live gaze refinement that can be learned end-to-end.
Both the training of our approach and its large-scale evaluation is made possible by the \datasetname, which should allow for insightful comparisons in the future.
\\

\begin{figure}[t]
    \centering
    \begin{tabular}{m{3mm}ccc}
        \raisebox{18mm}{\rotatebox{90}{\scriptsize Image}} &
        \includegraphics[width=0.3\columnwidth]{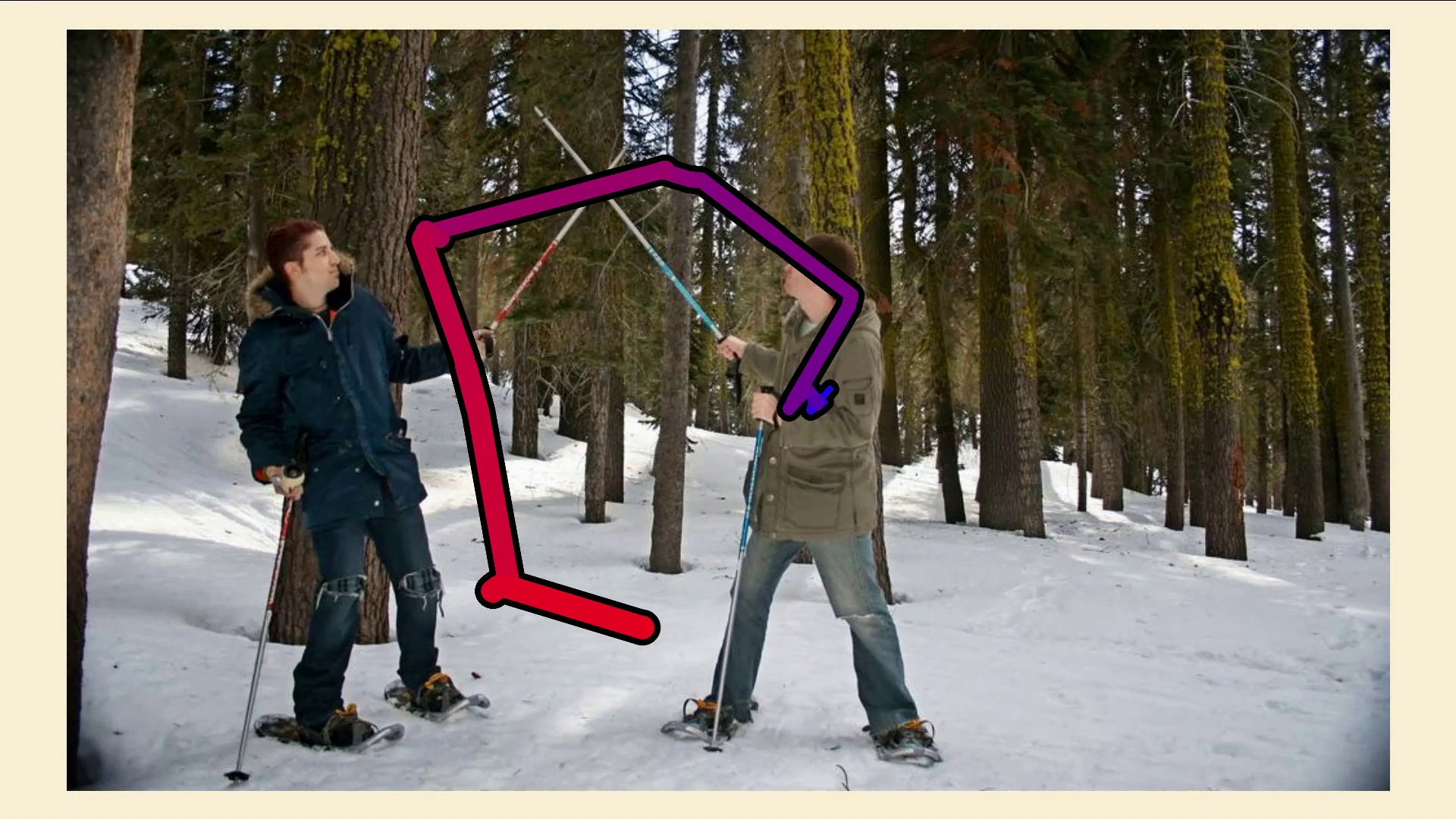} &
        \includegraphics[width=0.3\columnwidth]{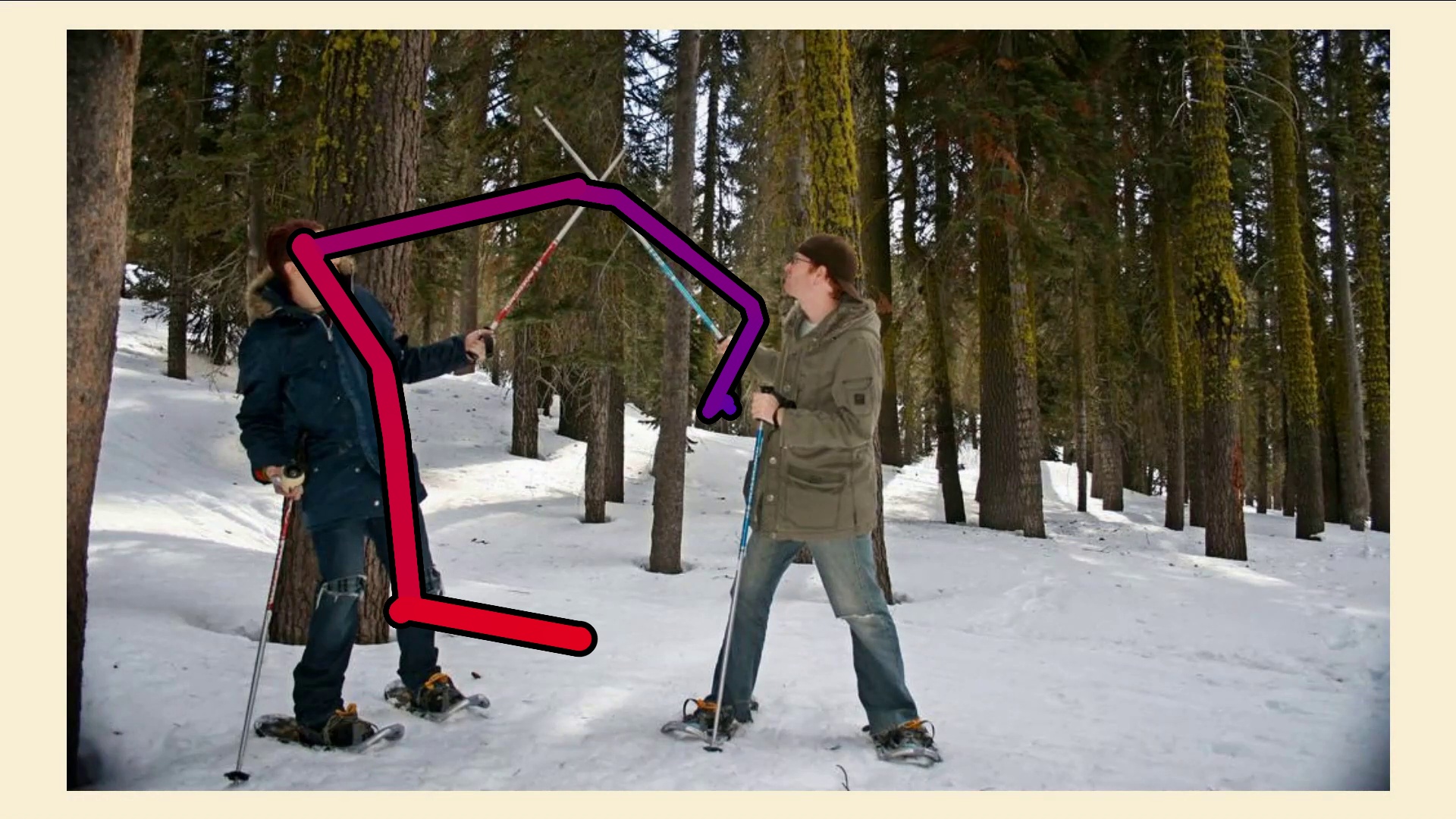} &
        \includegraphics[width=0.3\columnwidth]{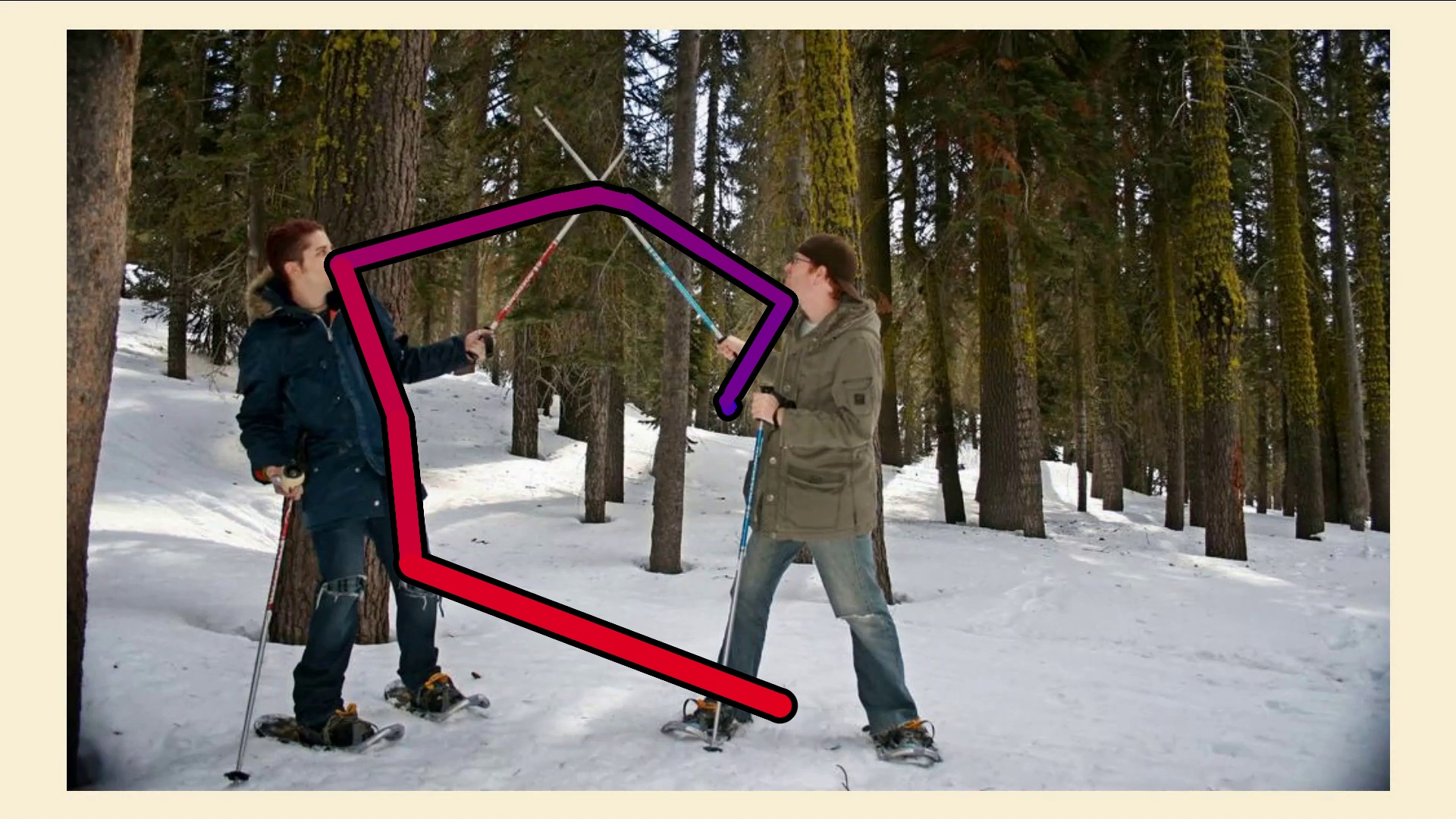} \\[-11mm]
        \raisebox{18mm}{\rotatebox{90}{\scriptsize Video}} &
        \includegraphics[width=0.3\columnwidth]{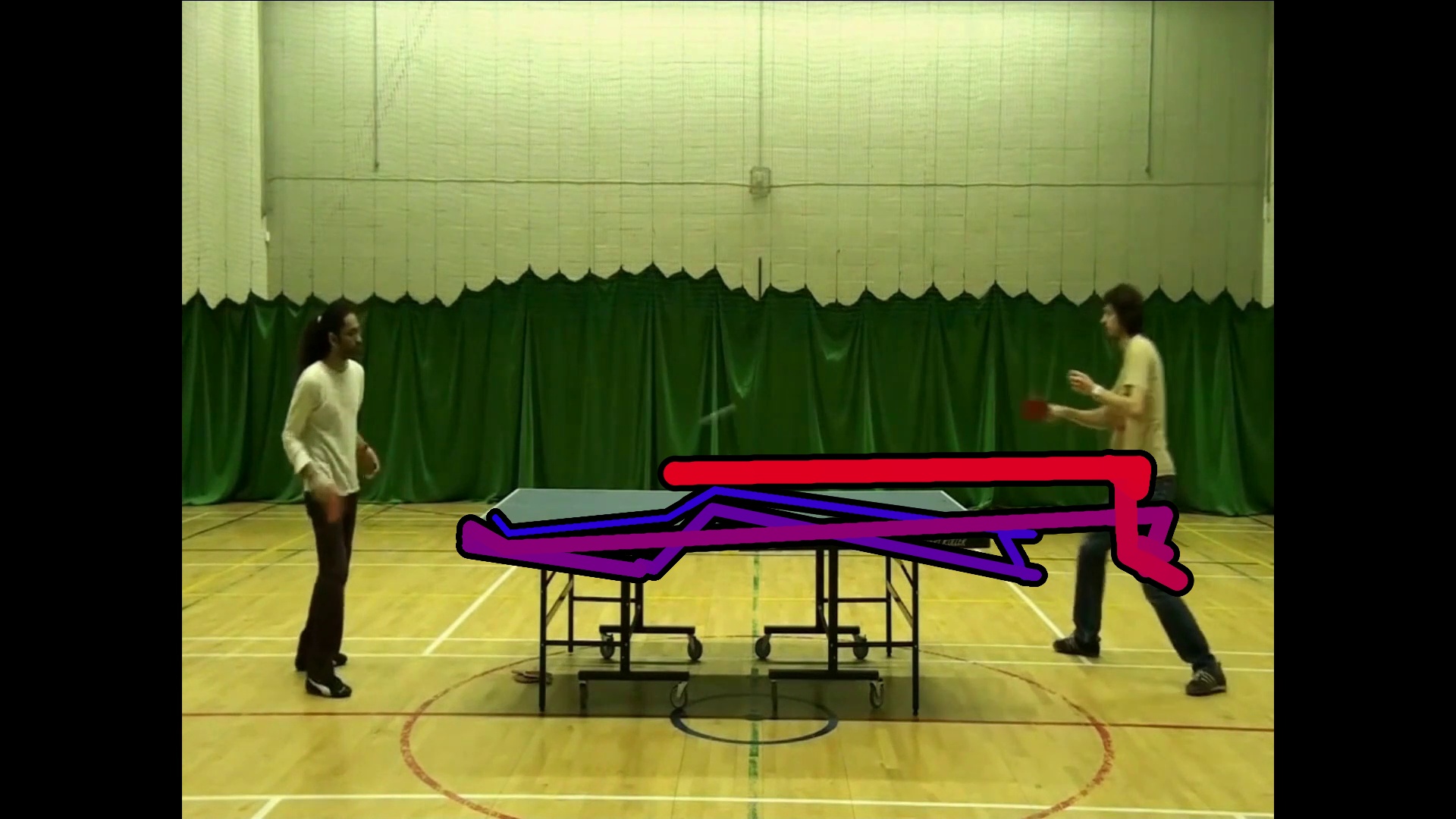} &
        \includegraphics[width=0.3\columnwidth]{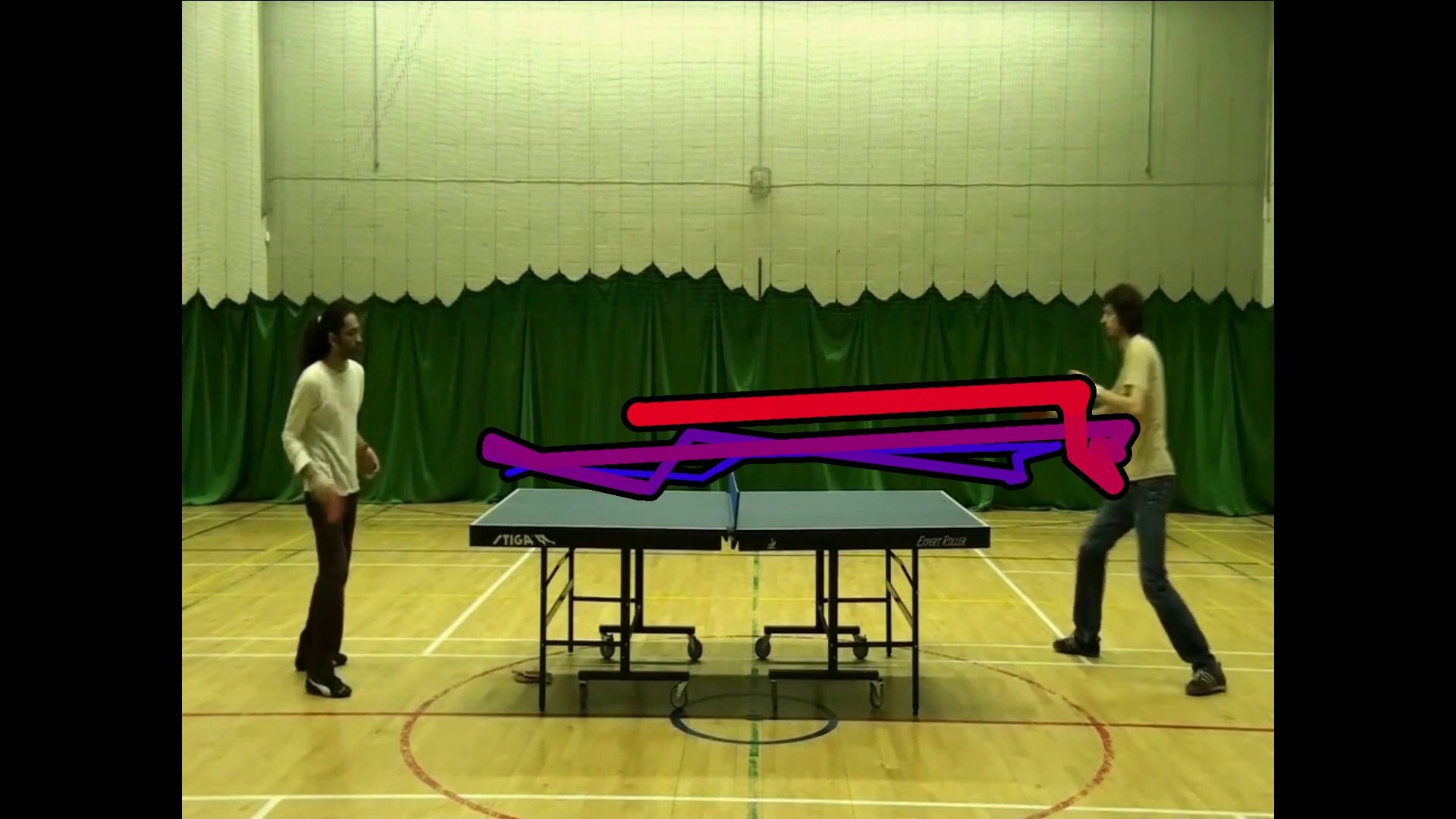} &
        \includegraphics[width=0.3\columnwidth]{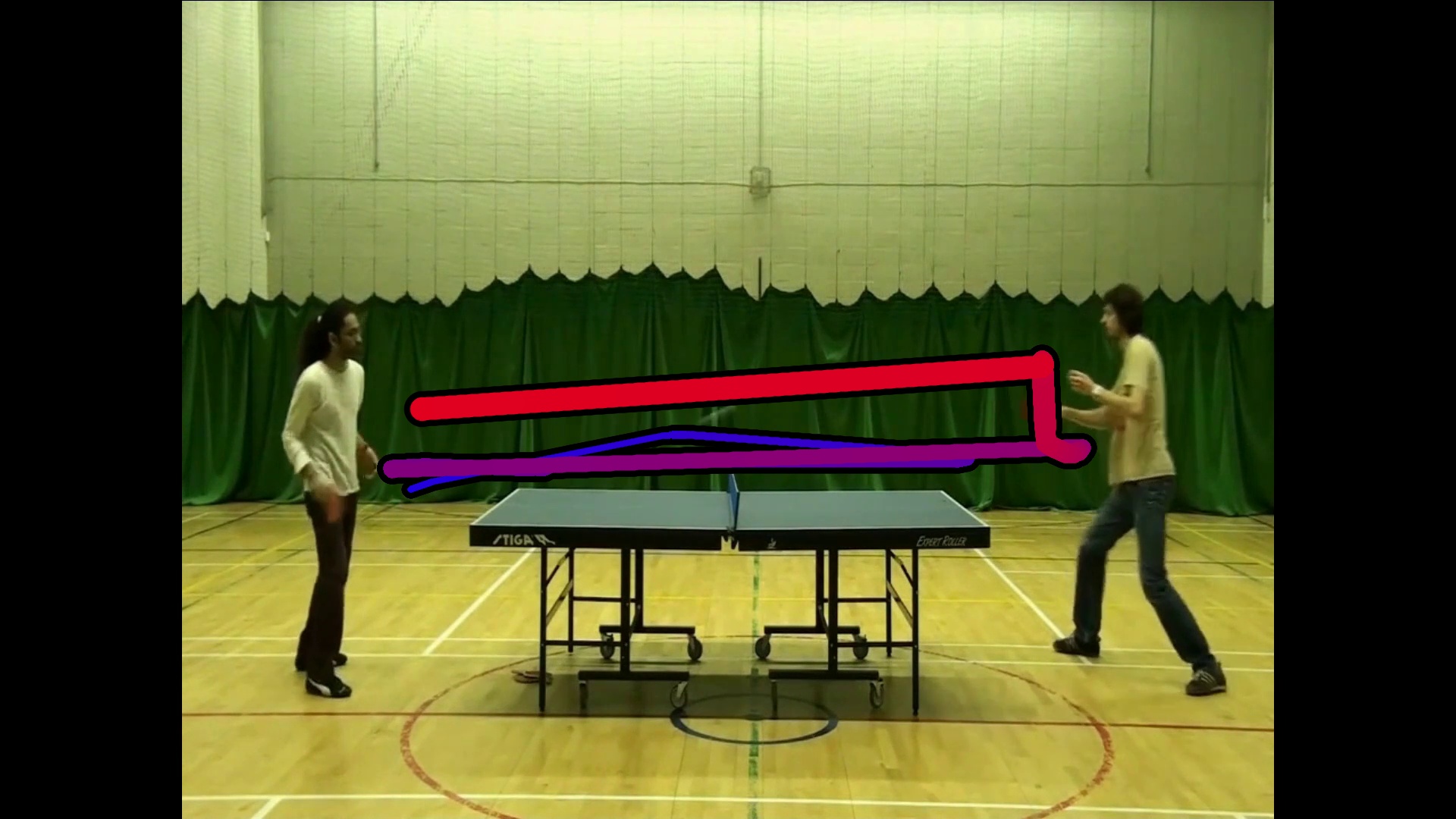} \\[-11mm]
        \raisebox{18mm}{\rotatebox{90}{\scriptsize Wikipedia}} &
        \includegraphics[width=0.3\columnwidth]{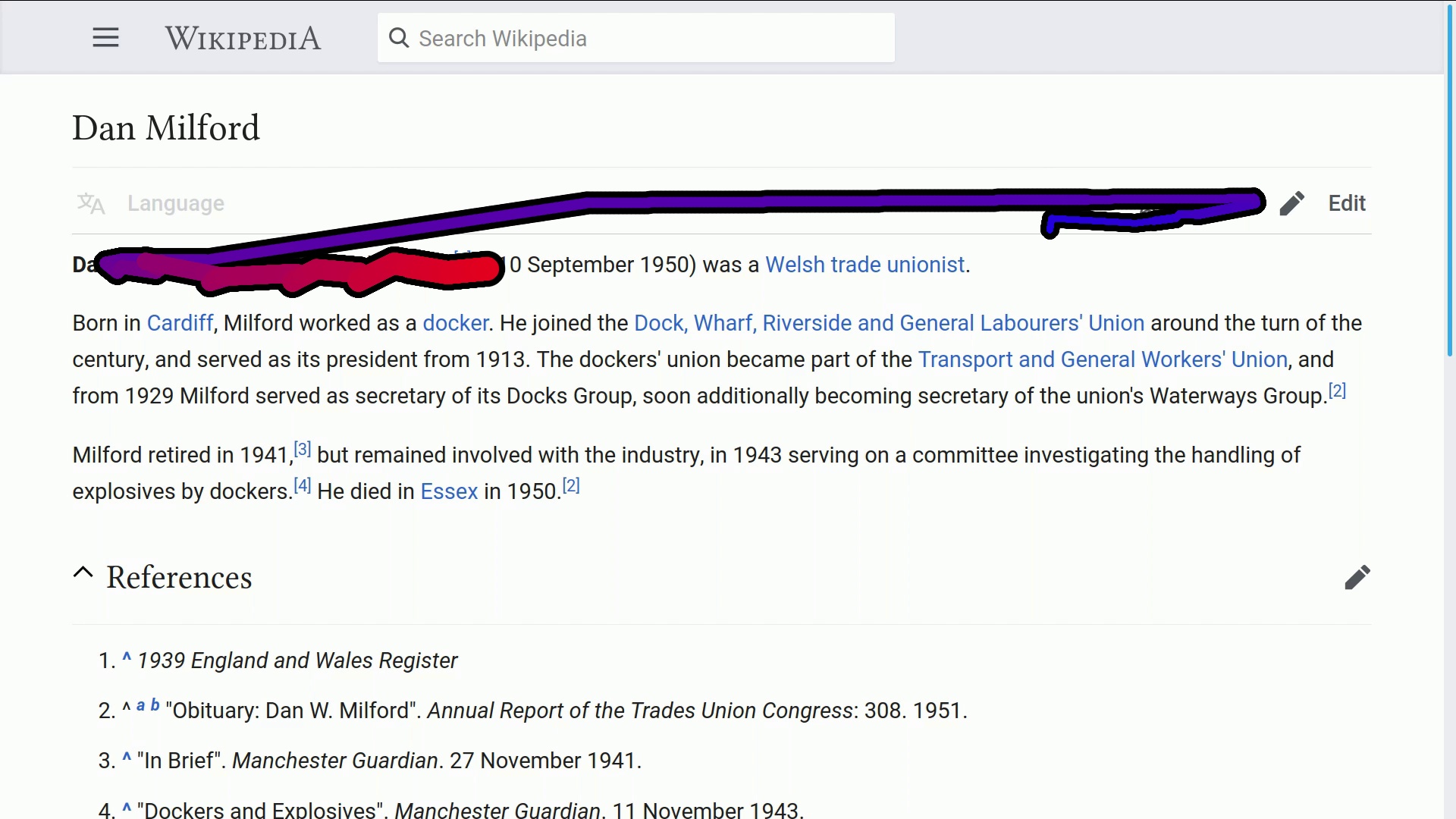} &
        \includegraphics[width=0.3\columnwidth]{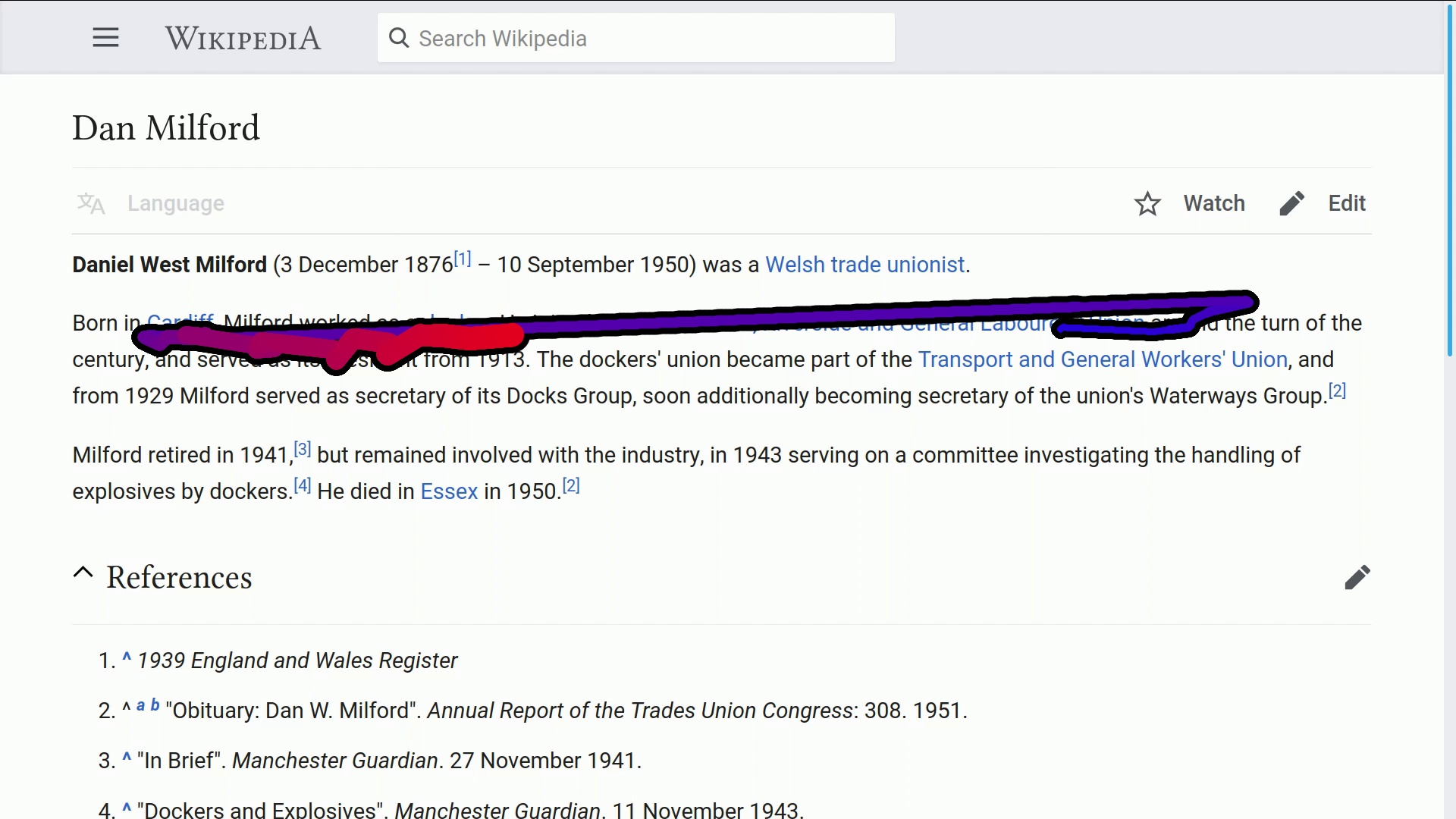} &
        \includegraphics[width=0.3\columnwidth]{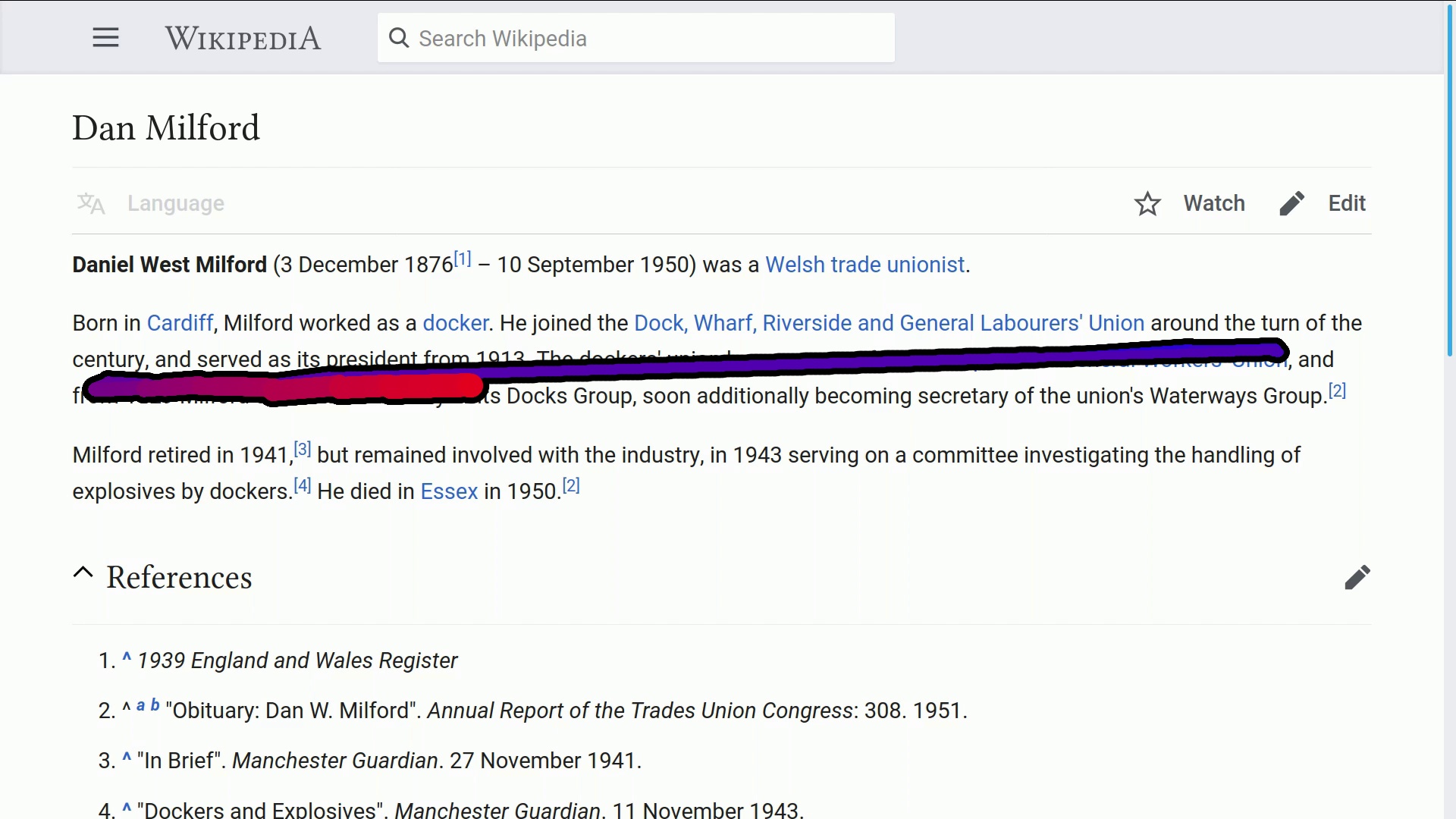} \\[-14mm]
        & {\scriptsize Initial Prediction} 
        & {\scriptsize Refined Prediction}
        & {\scriptsize Ground-truth} \\
    \end{tabular}
    \\[-2mm]
    \caption{Qualitative results of our gaze refinement method on our test set, where PoG over time are colored from blue-to-red (old-to-new). It can be seen that \refinenetname corrects offsets between the initial prediction and ground-truth.
    }
    \label{fig:qualitative}
\end{figure}

\noindent\textbf{Cross-Stimuli Evaluation}.
We study if our method generalizes to novel stimuli types, as this has previously been raised as in issue for saliency-based gaze alignment methods (such as in \cite{Sugano2015UIST}). In Tab.~\ref{tab:cross_stimuli_refinement}, we confirm that indeed training and testing on the same stimulus type yields the greatest improvements in gaze direction estimation (shown in diagonal of table).
We find in general that large improvements can be observed even when training solely on video or wikipedia stimuli types. One assumes that this is the case due to the existence of text in our video stimuli and the existence of small amounts of images in the wikipedia stimulus.
In contrast, we can see that training a model on static images only does not lead to good generalization on the stimuli types.
\\

\noindent\textbf{Qualitative Results}.
We visualize our results qualitatively in Fig.~\ref{fig:qualitative}. Specifically, we can see that when provided with initial estimates of PoG over time from \eyenetgruname (far-left column), our \refinenetgruname can nicely recover person-specific offsets at test time to yield improved estimates of PoG (center column).
When viewed in comparison with the ground-truth (far-right column), the success of \refinenetgruname in these example cases is clear.
In addition, note that the final operation is not one of pure offset-correction, but that the gaze signal is more aligned with the visual layout of the screen content post-refinement. \section{Conclusion}

In this paper, we introduced several effective steps towards increasing screen-based eye-tracking performance even in the absence of labeled samples or eye-tracker calibration from the final target user.
Specifically, we identified that eye movements and the change in visual stimulus have a complex interplay which previous literature have considered in a disconnected manner.
Subsequently, we proposed a novel dataset (\datasetname) for evaluating temporal gaze estimation models and for enabling a novel online PoG-refinement task based on raw screen content.
Our \refinenetname architecture performs this task effectively, and demonstrates large performance improvements of up to $28\%$.
The final reported angular gaze error of $2.49^\circ$ is achieved without labeled samples from the test set.

The \datasetname dataset is made publicly available\footnote{\dataseturl}, with a public web server implemented for consistent test metric calculations.
We provide the dataset and accompanying training and evaluation code in hopes of further progress in the field of remote webcam-based gaze estimation.
Comprehensive additional information regarding the capture, pre-processing, and characteristics of the dataset is made available in our supplementary materials. 
\section*{Acknowledgements}
\begin{wrapfigure}{r}{0.3\columnwidth}
    \raggedleft
    \vspace{-4mm}
    \includegraphics[width=0.3\columnwidth]{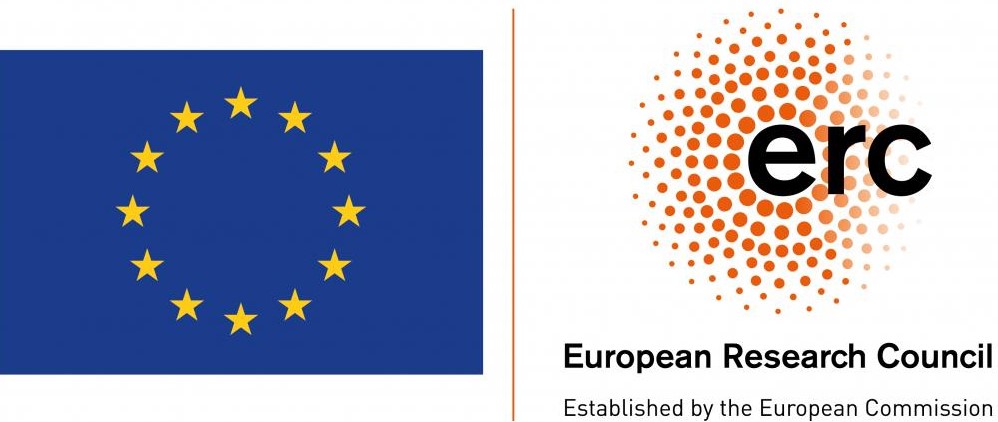}
\end{wrapfigure}
We thank the participants of our dataset for their contributions, our reviewers for helping us improve the paper, and Jan Wezel for helping with the hardware setup.
This project has received funding from the European Research Council (ERC) under the European Union’s Horizon 2020 research and innovation programme grant agreement No. StG-2016-717054.
 \clearpage
\bibliographystyle{splncs04}
\bibliography{egbib}

\clearpage
\setcounter{section}{0}
\renewcommand{\thesection}{\Alph{section}}
\section*{\Large Appendix}
\section{The \datasetname Dataset}

Much care was taken in capturing, pre-processing, and analyzing of the \datasetname dataset. We present a few additional details regarding these steps in this section.

\subsection{Ethics Approval}
The collection of this dataset and the procedure of the study was approved by the Ethics Commission of ETH Zurich (application no. 2019-N-103).
Before the beginning of a capture session, we clearly presented the risks (bodily and data-related) to our participants via information sheets and a comprehensive consent form.
Participants were recruited via a university job board\footnote{\url{https://marktplatz.uzhalumni.ch/}} and after the hour-long session, were paid a fee of 25 Swiss Francs in cash.

\subsection{Actual Capture}
\begin{figure}[t]
    \centering
    \begin{subfigure}{0.24\columnwidth}
        \includegraphics[width=\columnwidth]{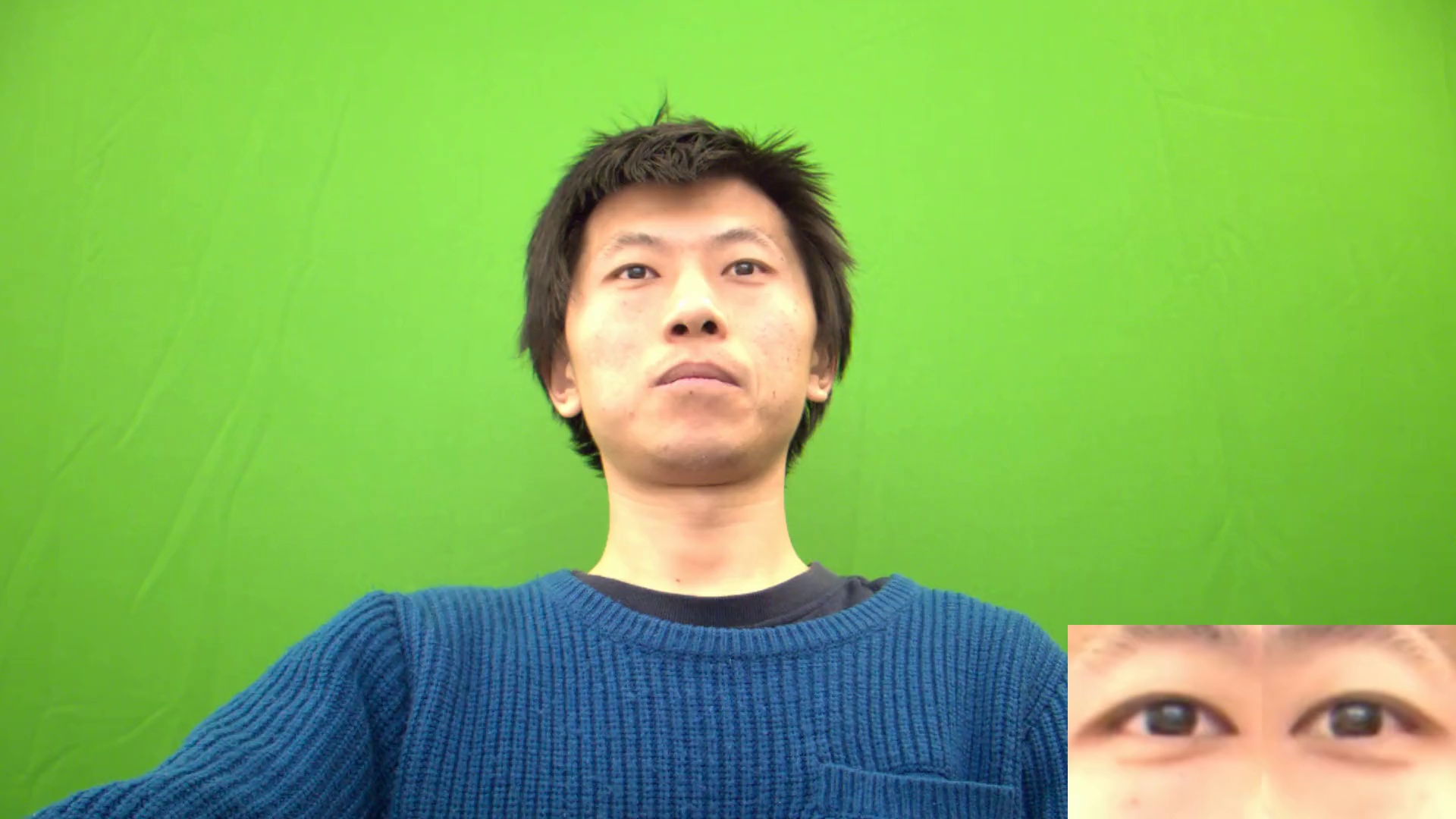}
        \includegraphics[width=\columnwidth]{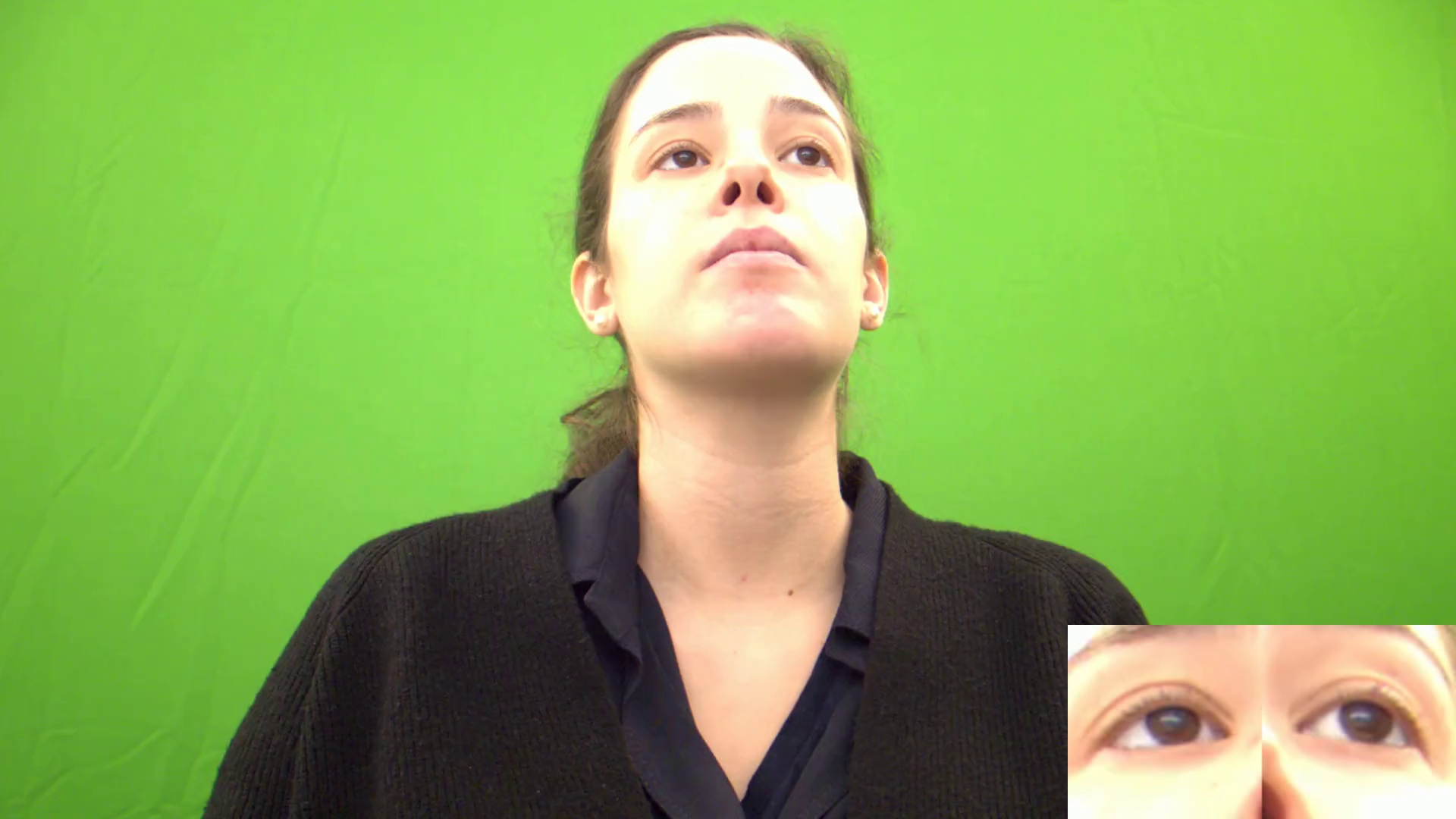}
        \includegraphics[width=\columnwidth]{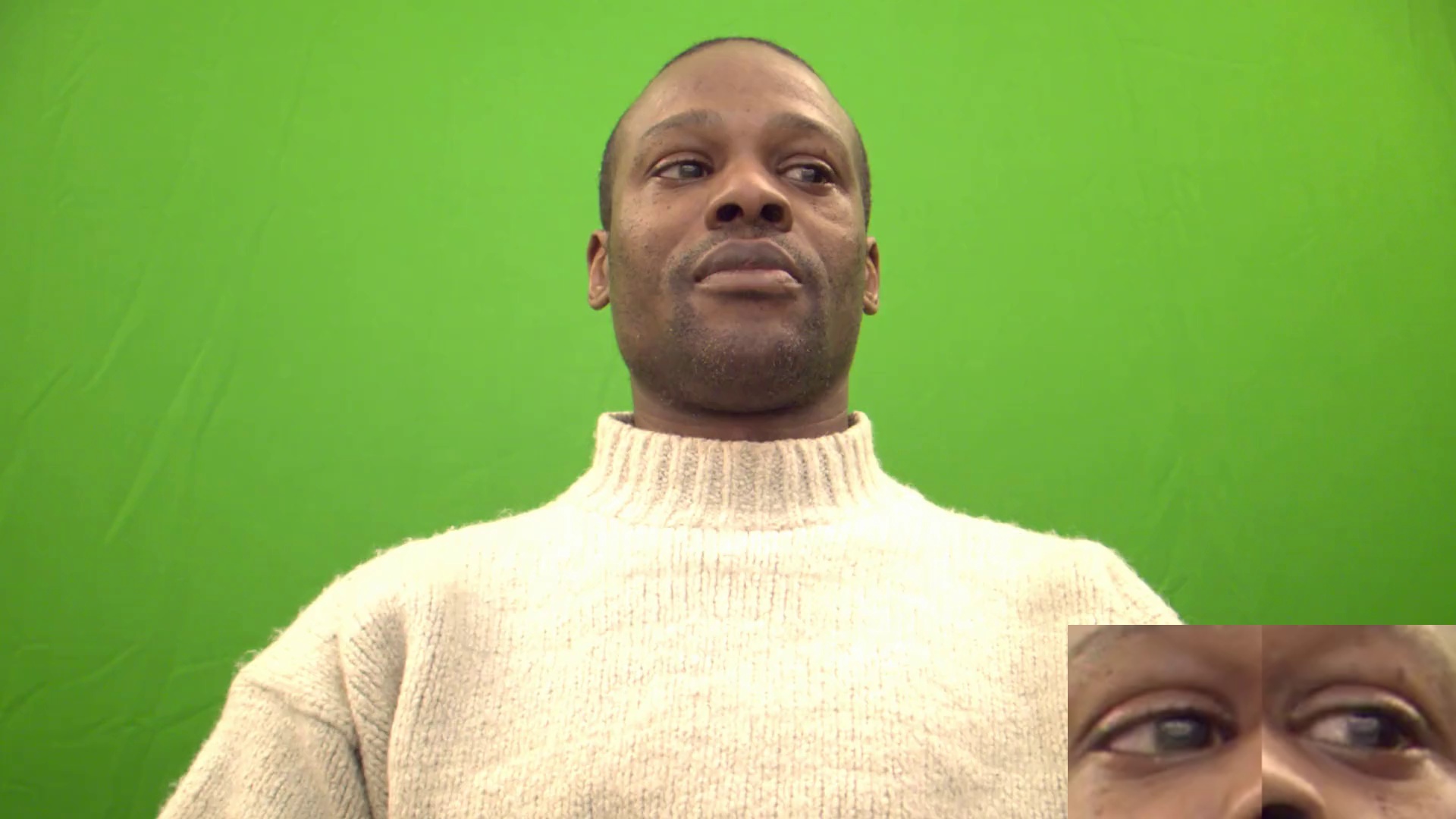}
        \includegraphics[width=\columnwidth]{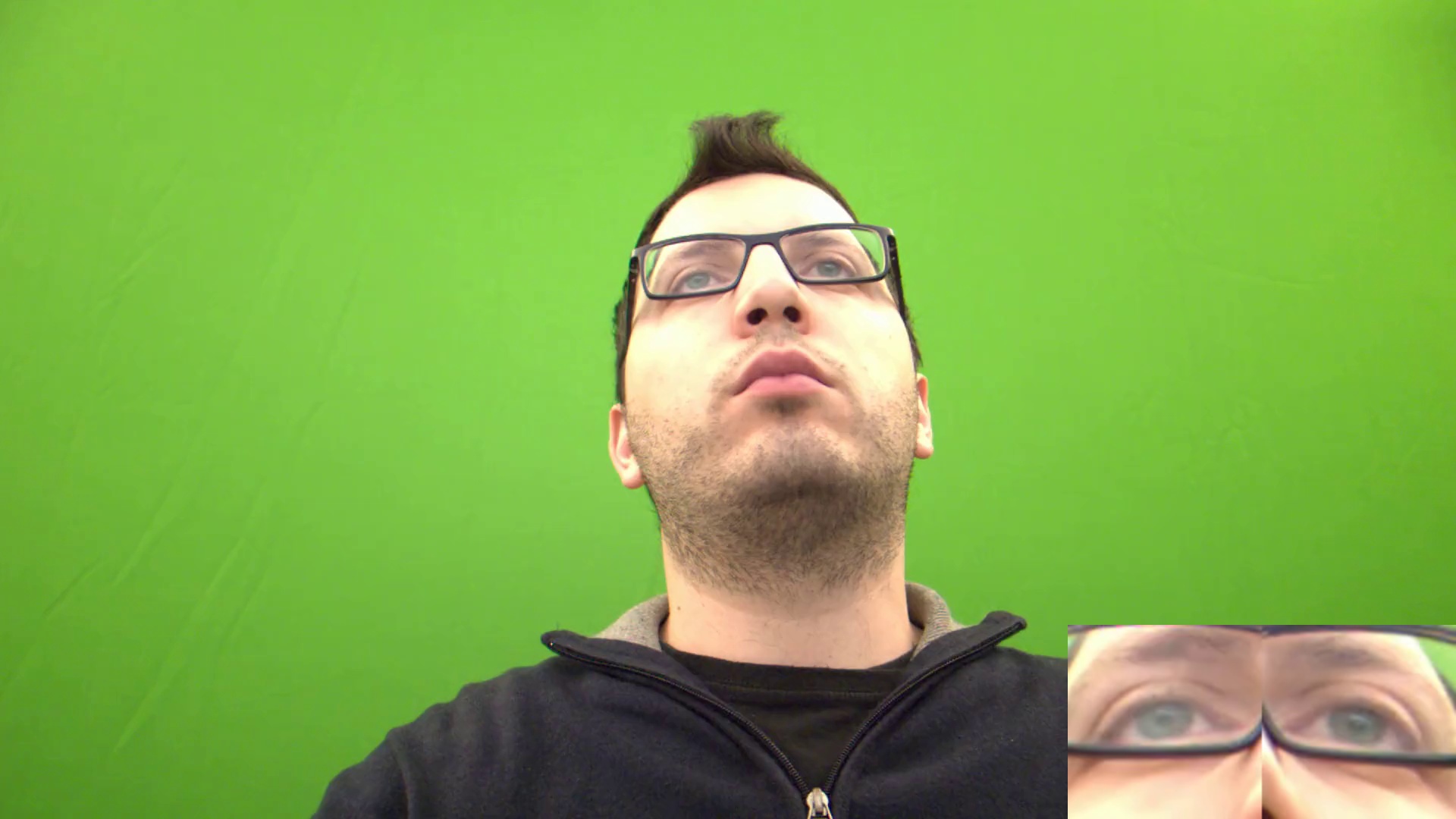}
        \includegraphics[width=\columnwidth]{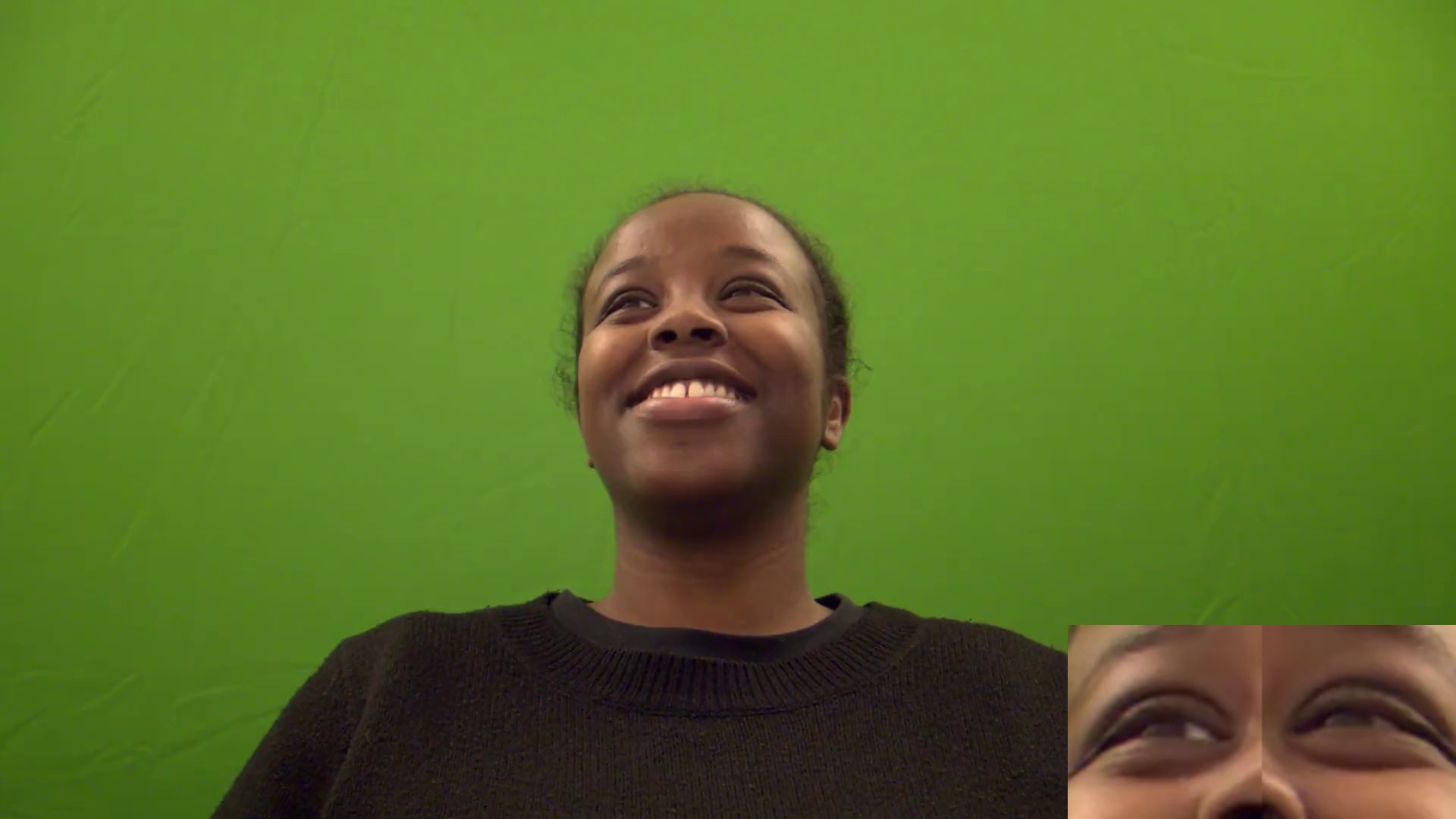}
        \includegraphics[width=\columnwidth]{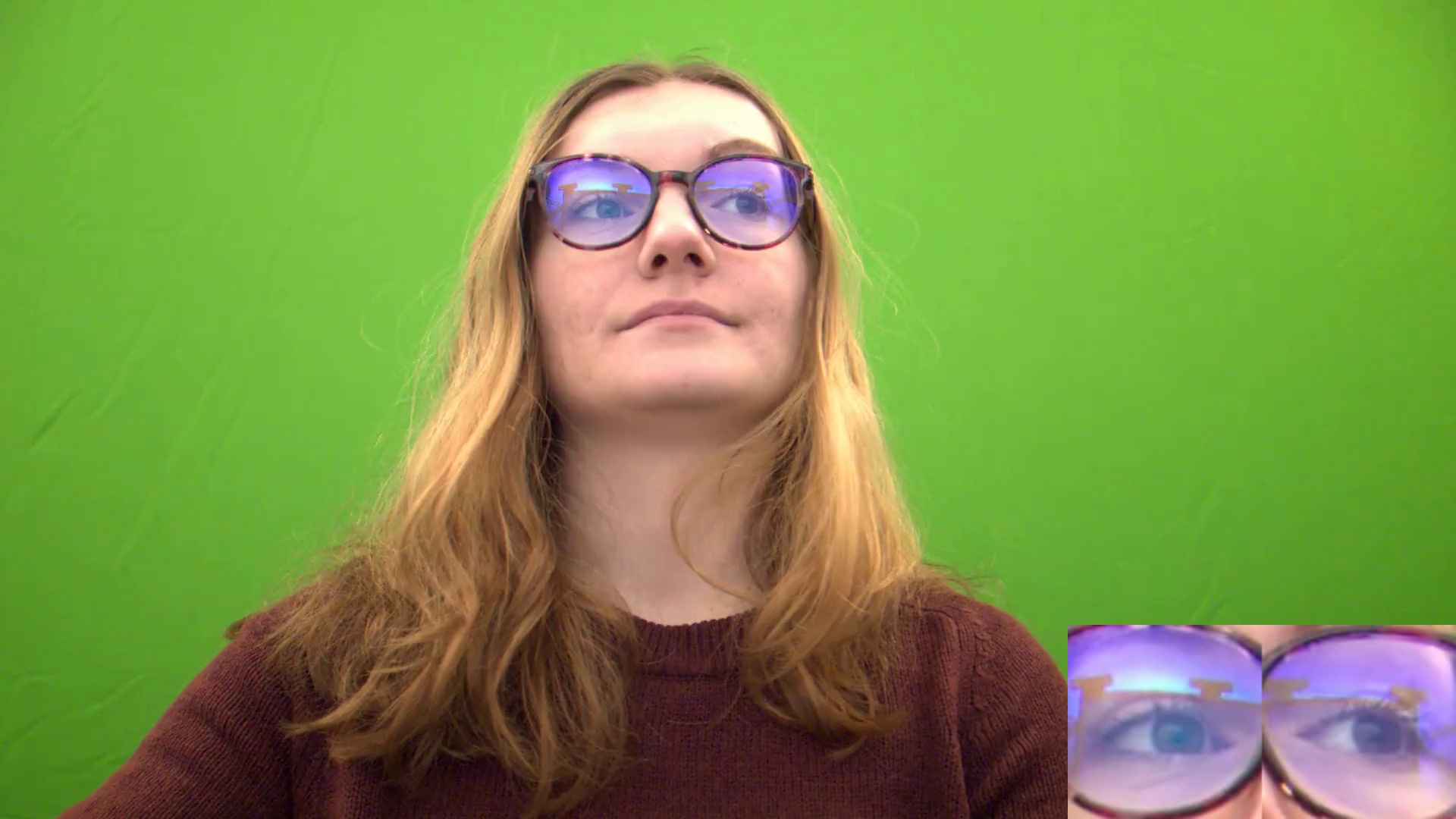}
        \caption{\scriptsize MVC}
    \end{subfigure}
    \begin{subfigure}{0.24\columnwidth}
        \includegraphics[width=\columnwidth]{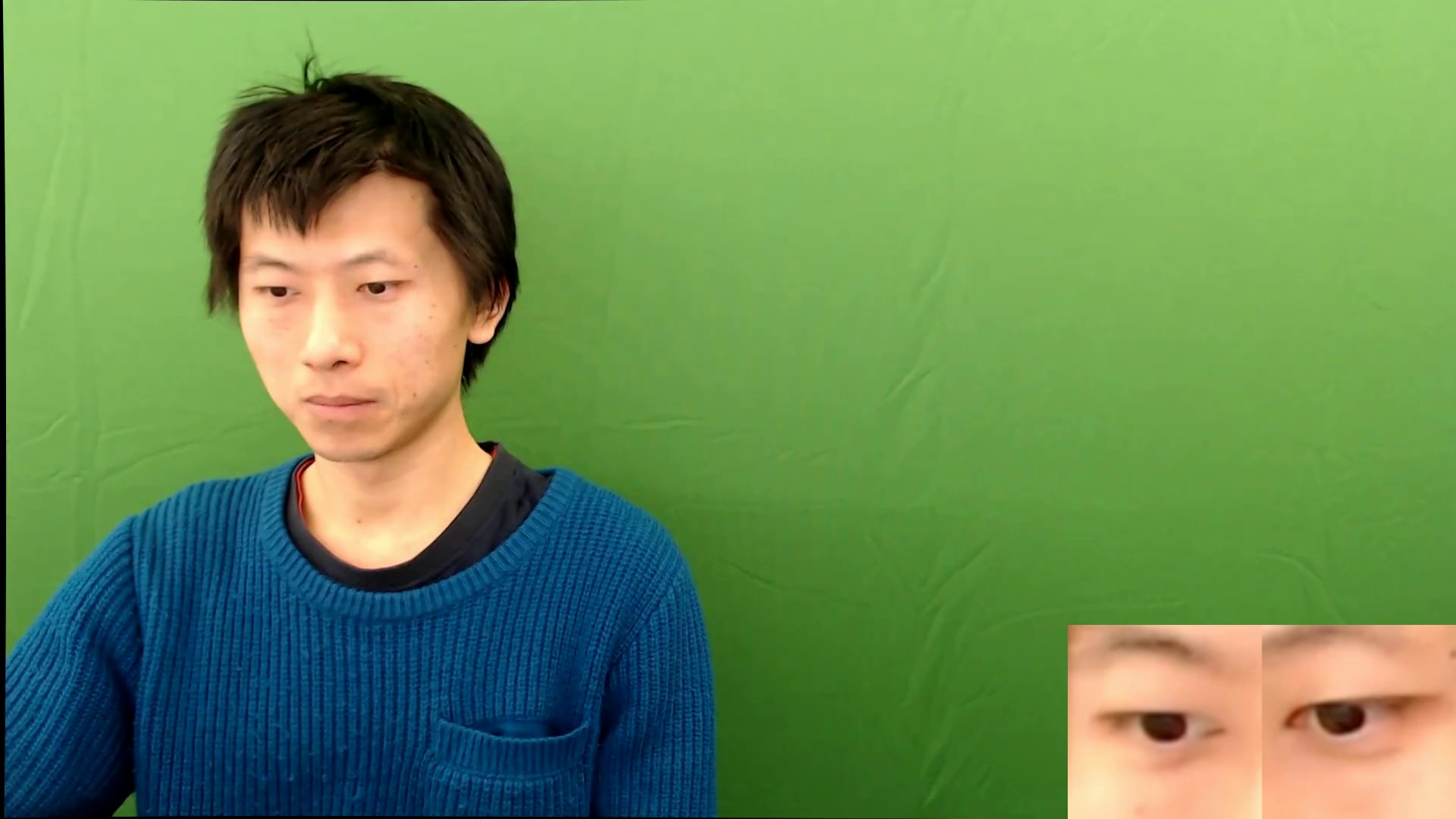}
        \includegraphics[width=\columnwidth]{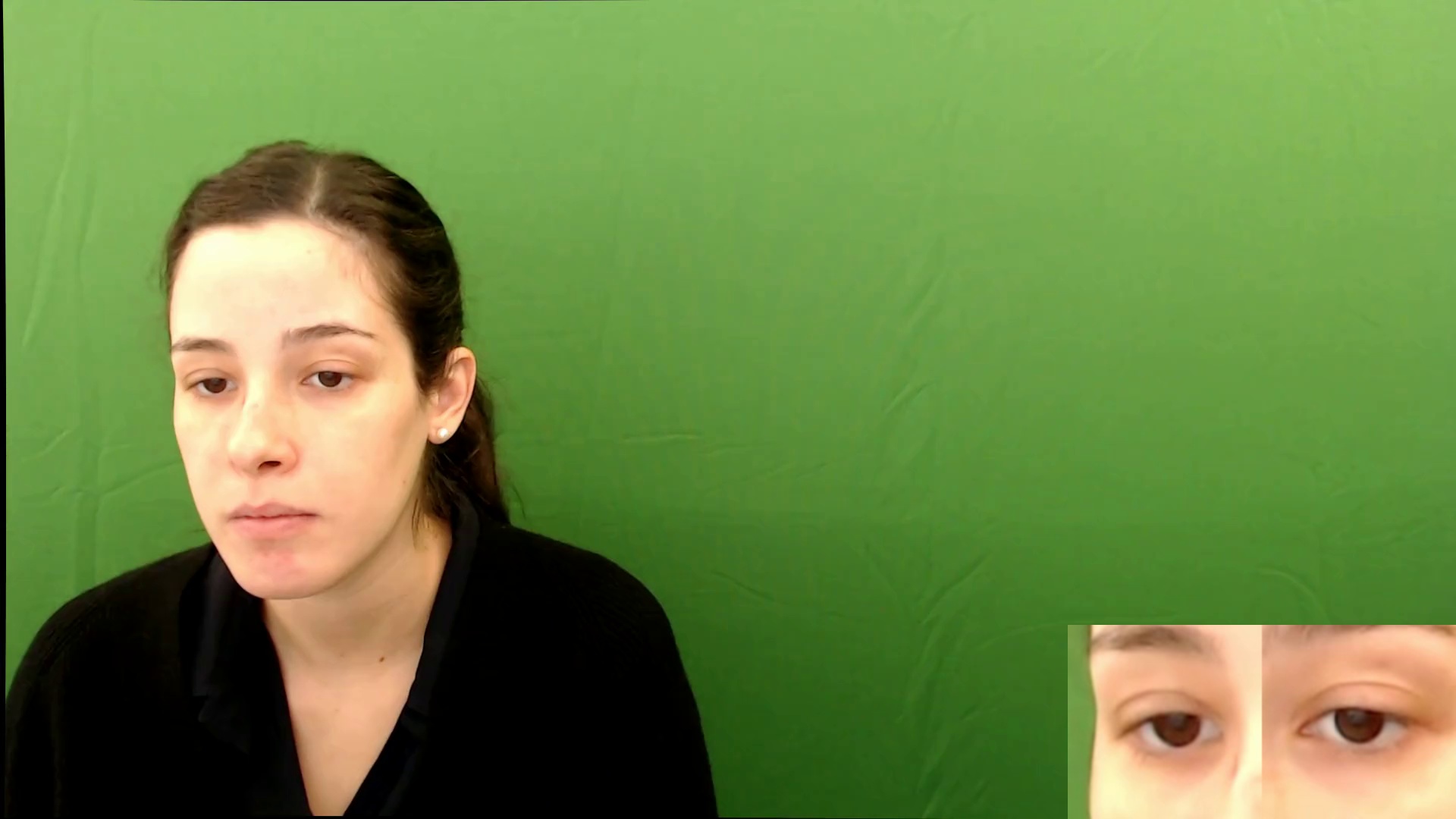}
        \includegraphics[width=\columnwidth]{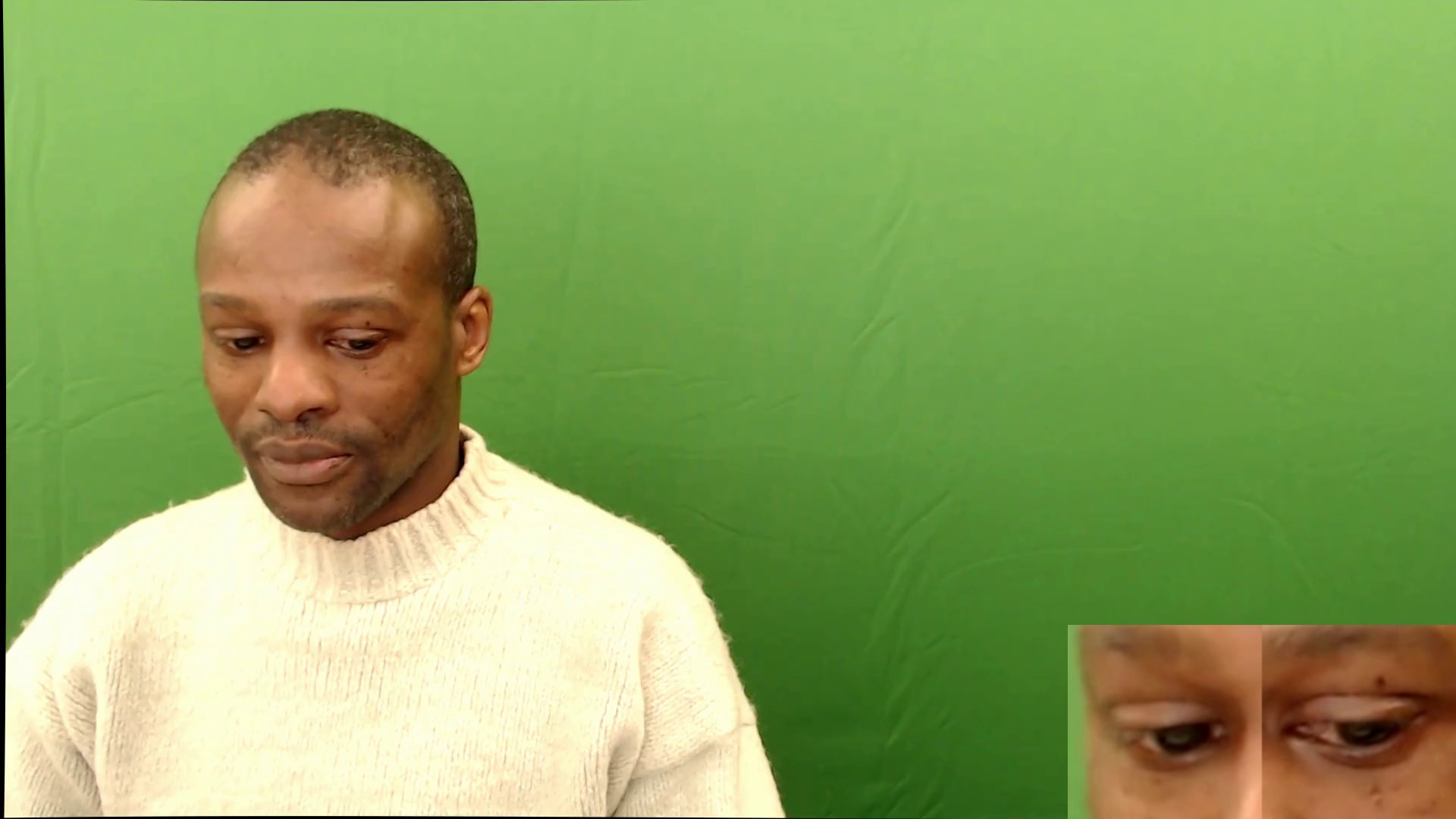}
        \includegraphics[width=\columnwidth]{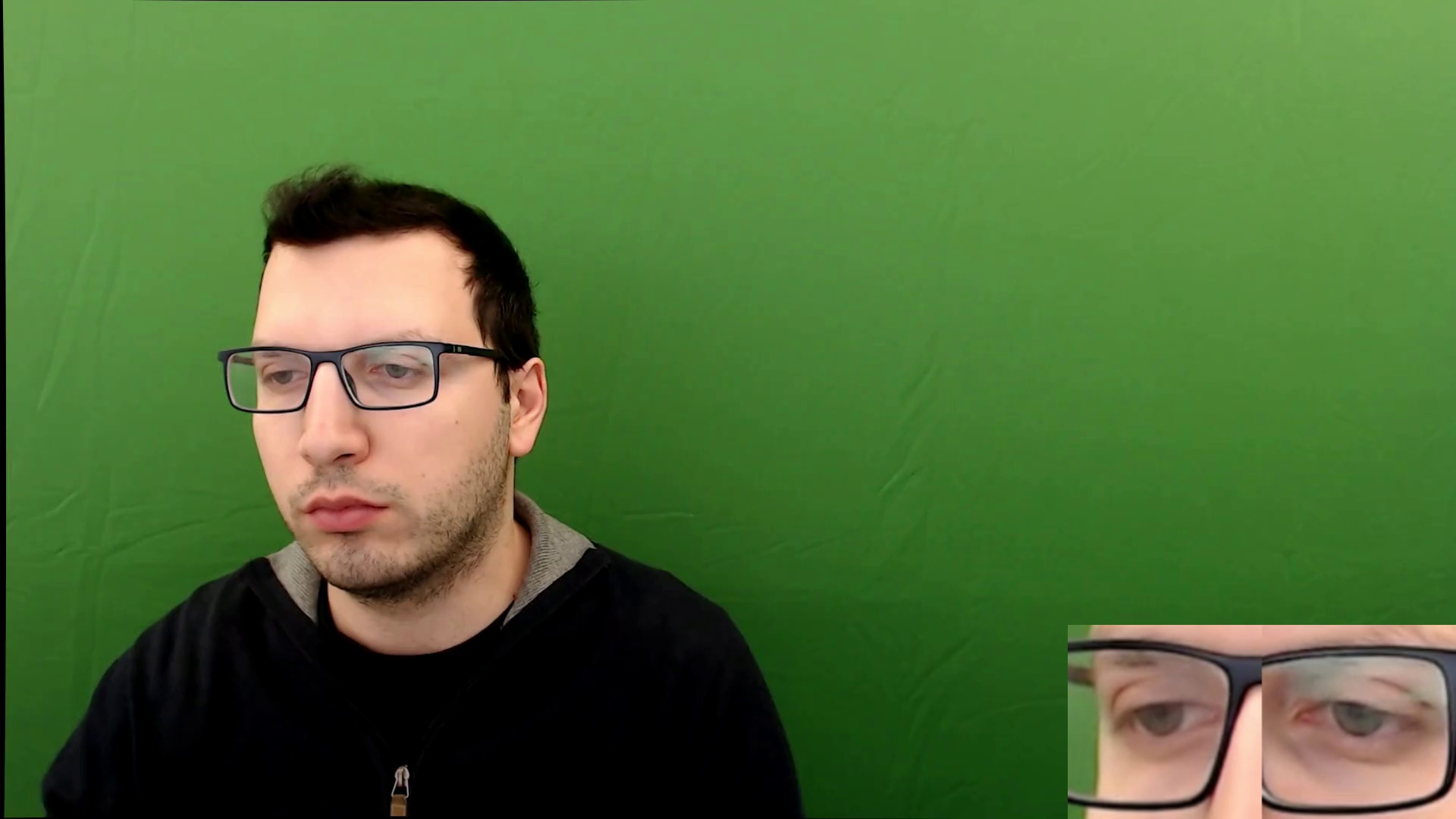}
        \includegraphics[width=\columnwidth]{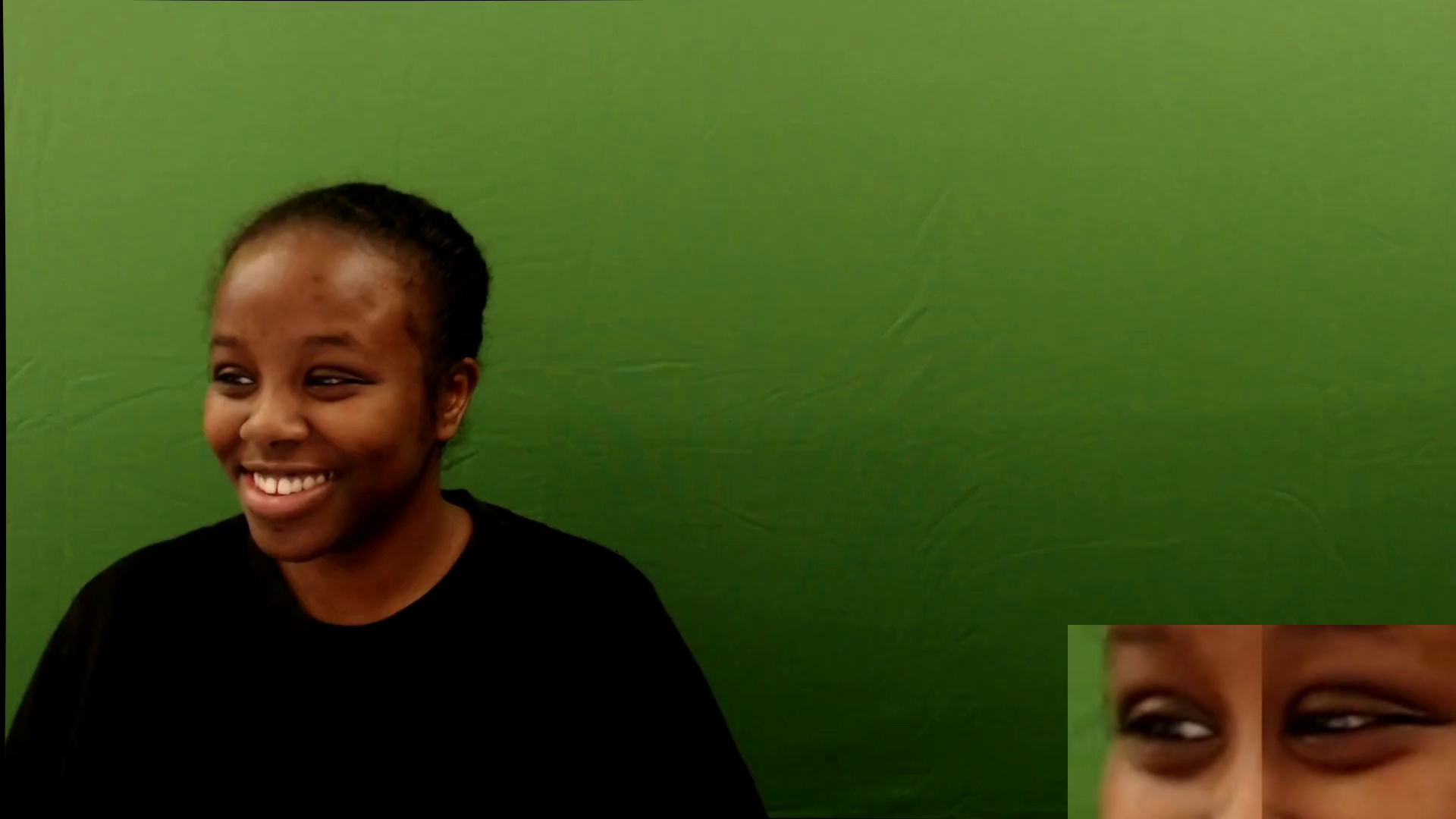}
        \includegraphics[width=\columnwidth]{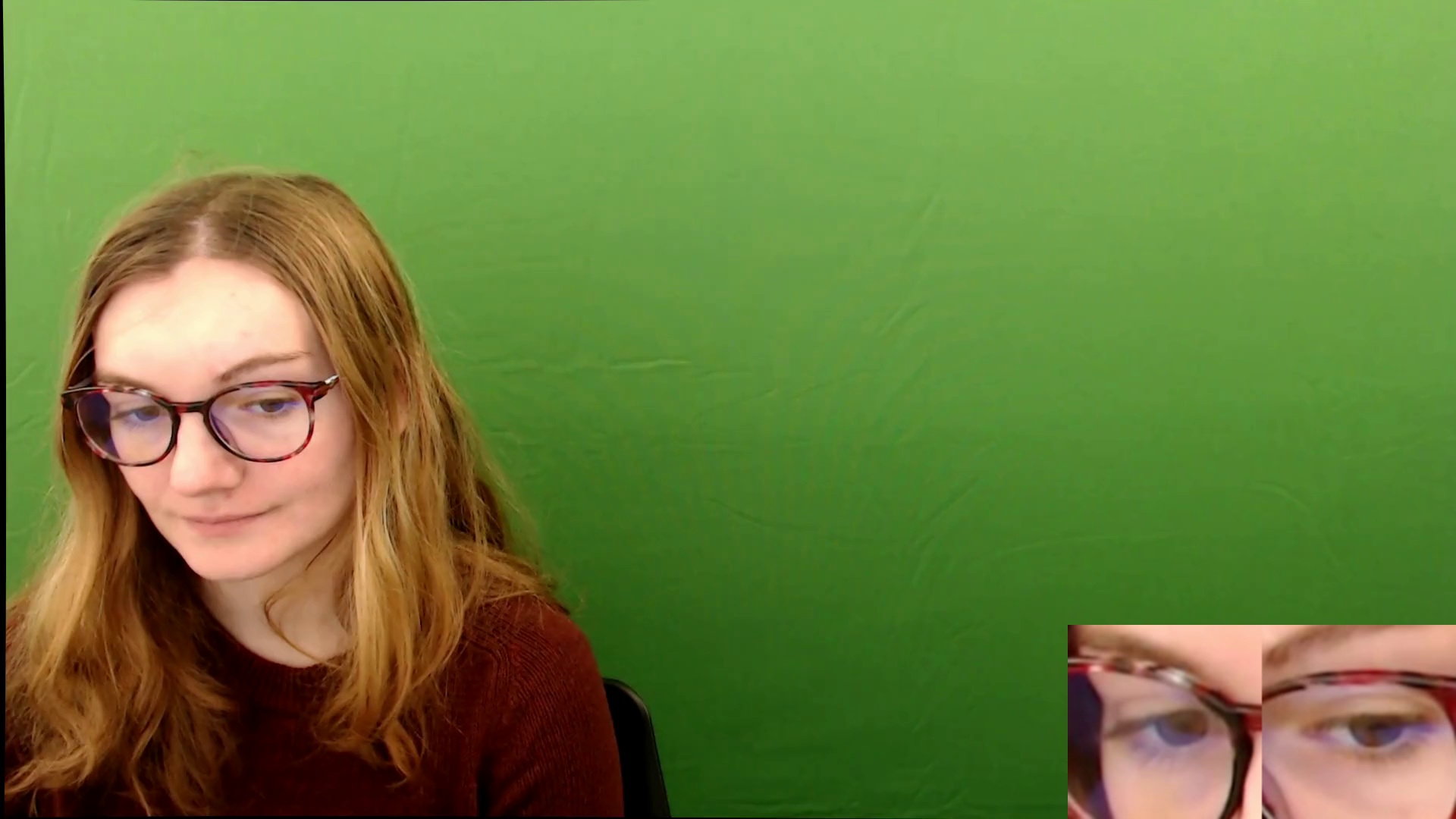}
        \caption{\scriptsize Webcam (Left)}
    \end{subfigure}
    \begin{subfigure}{0.24\columnwidth}
        \includegraphics[width=\columnwidth]{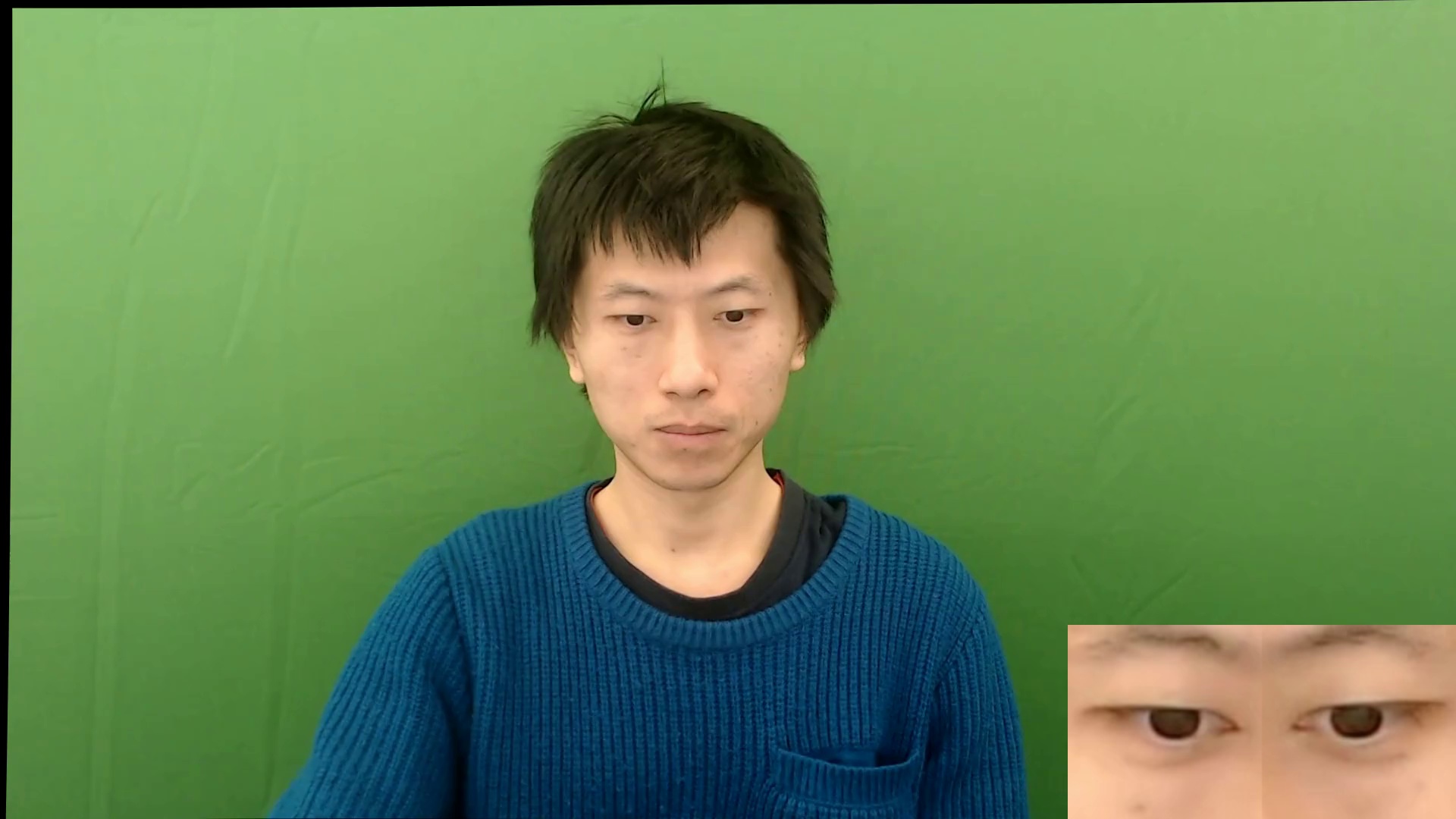}
        \includegraphics[width=\columnwidth]{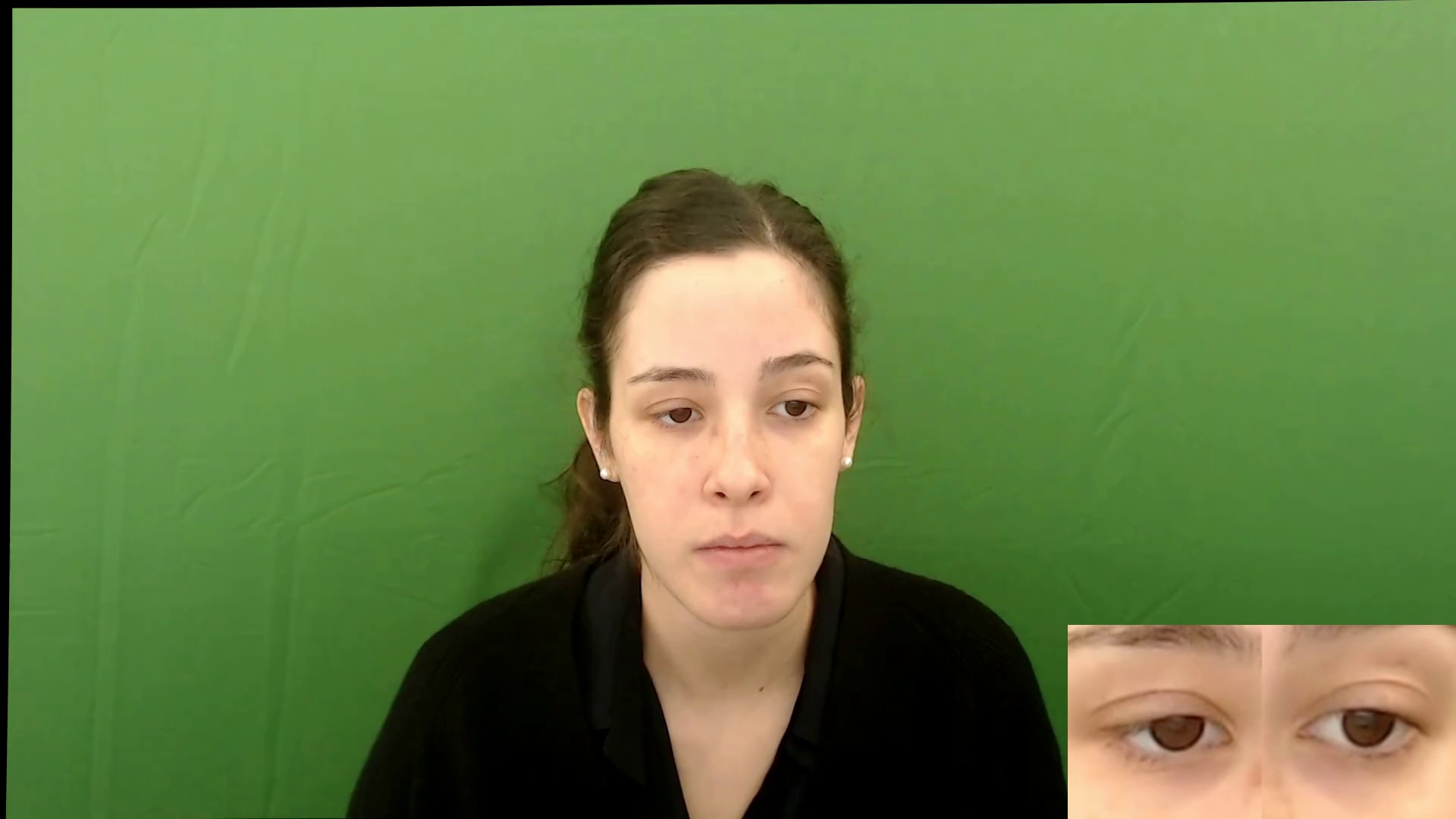}
        \includegraphics[width=\columnwidth]{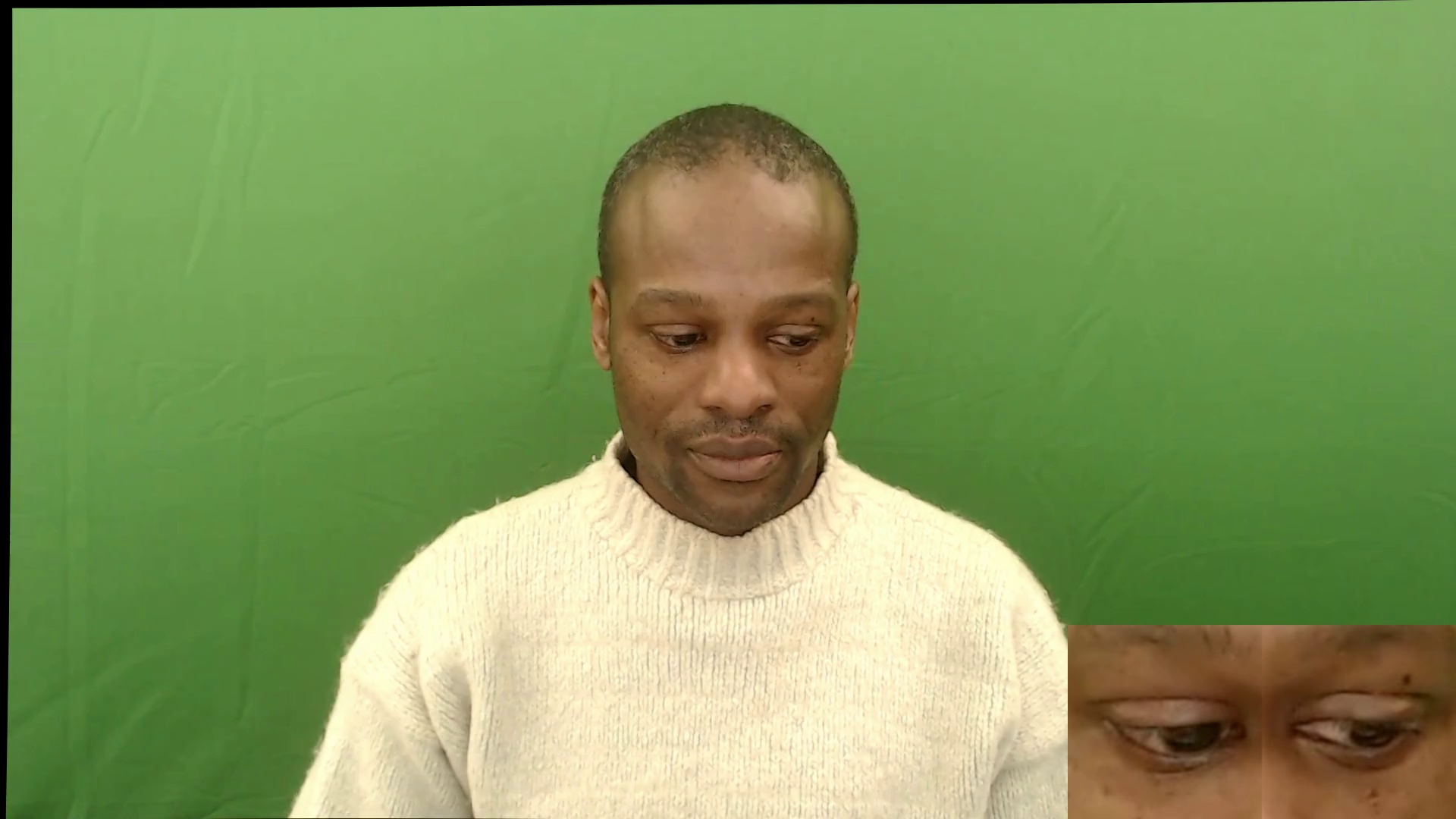}
        \includegraphics[width=\columnwidth]{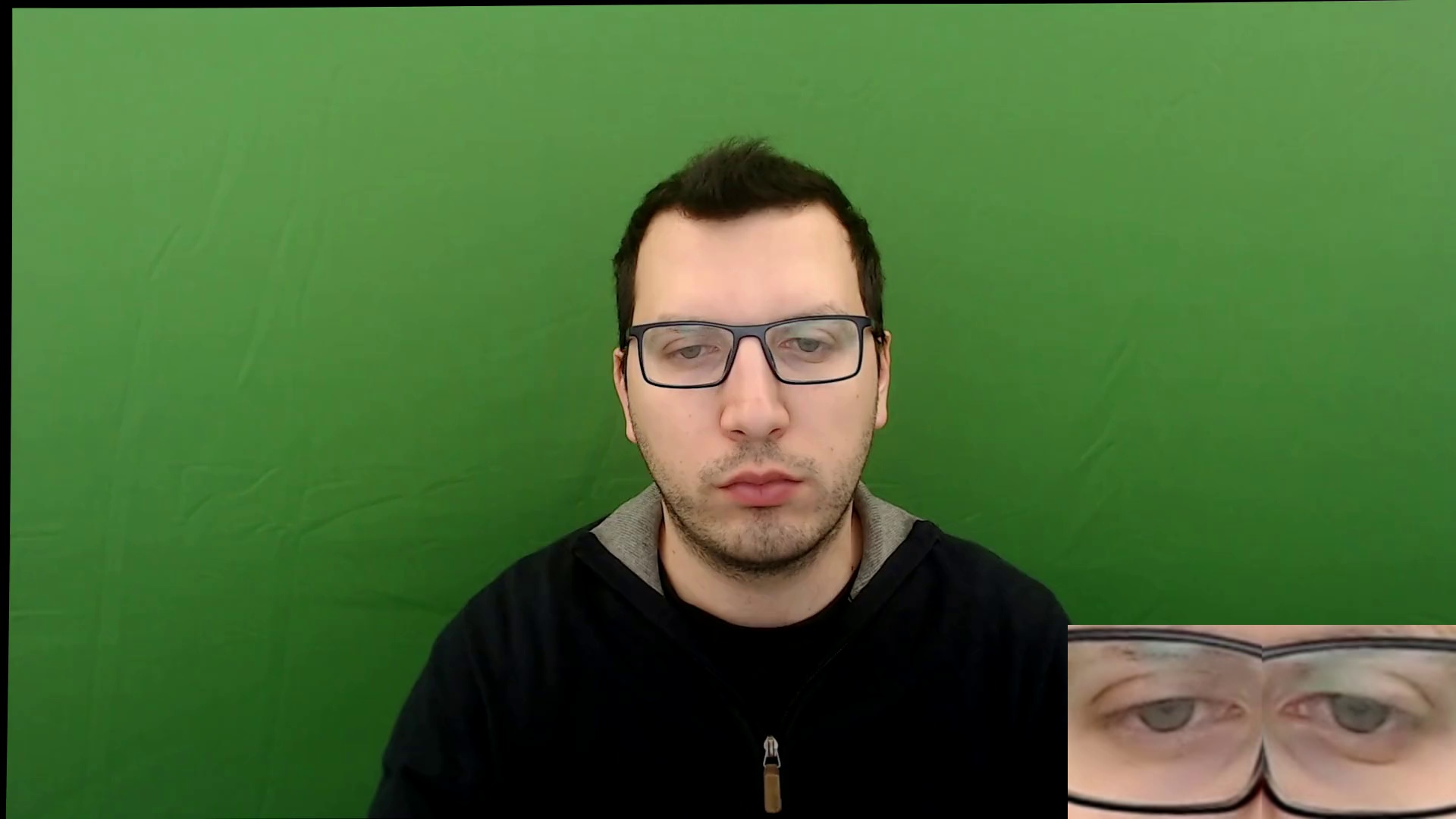}
        \includegraphics[width=\columnwidth]{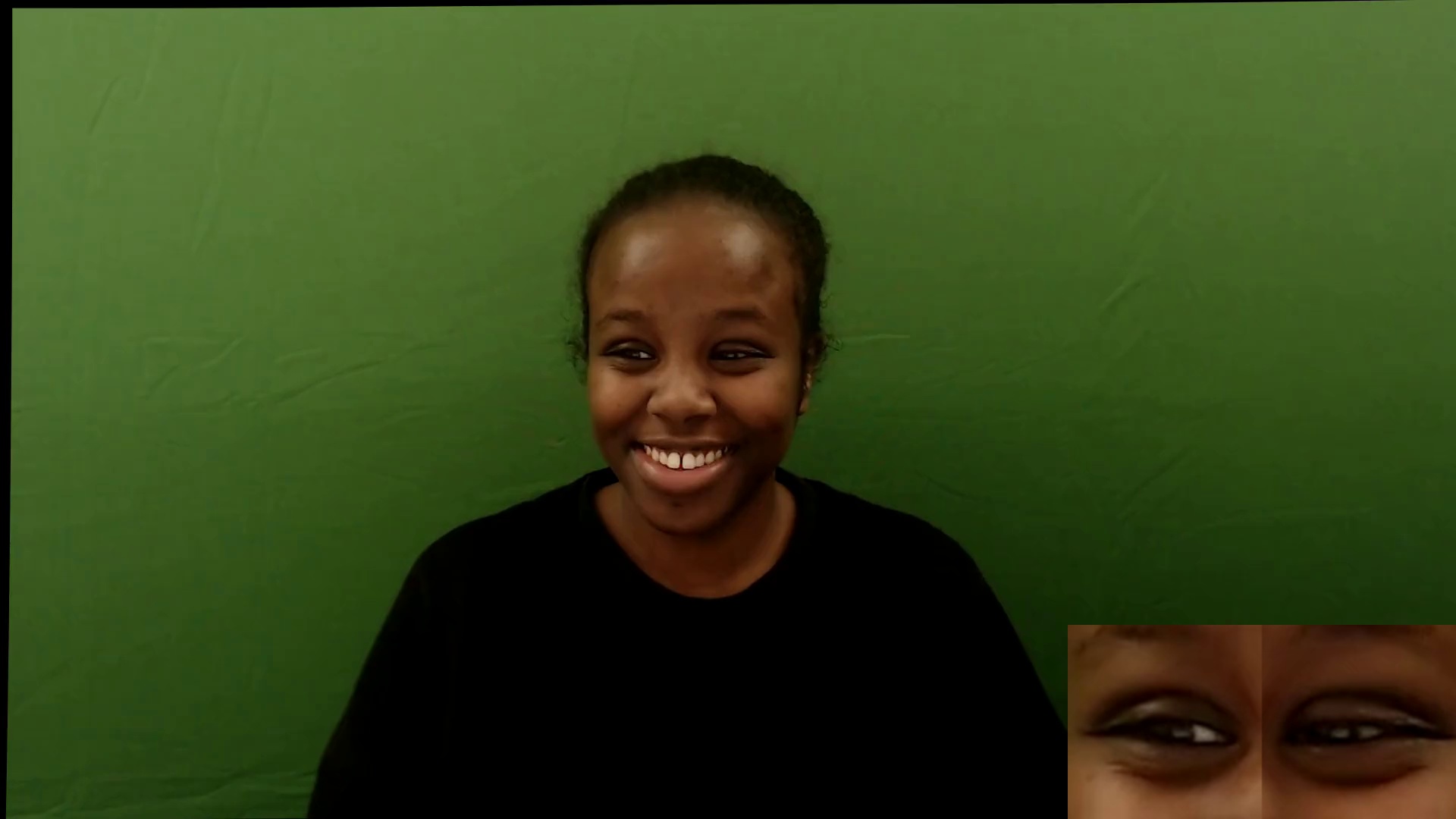}
        \includegraphics[width=\columnwidth]{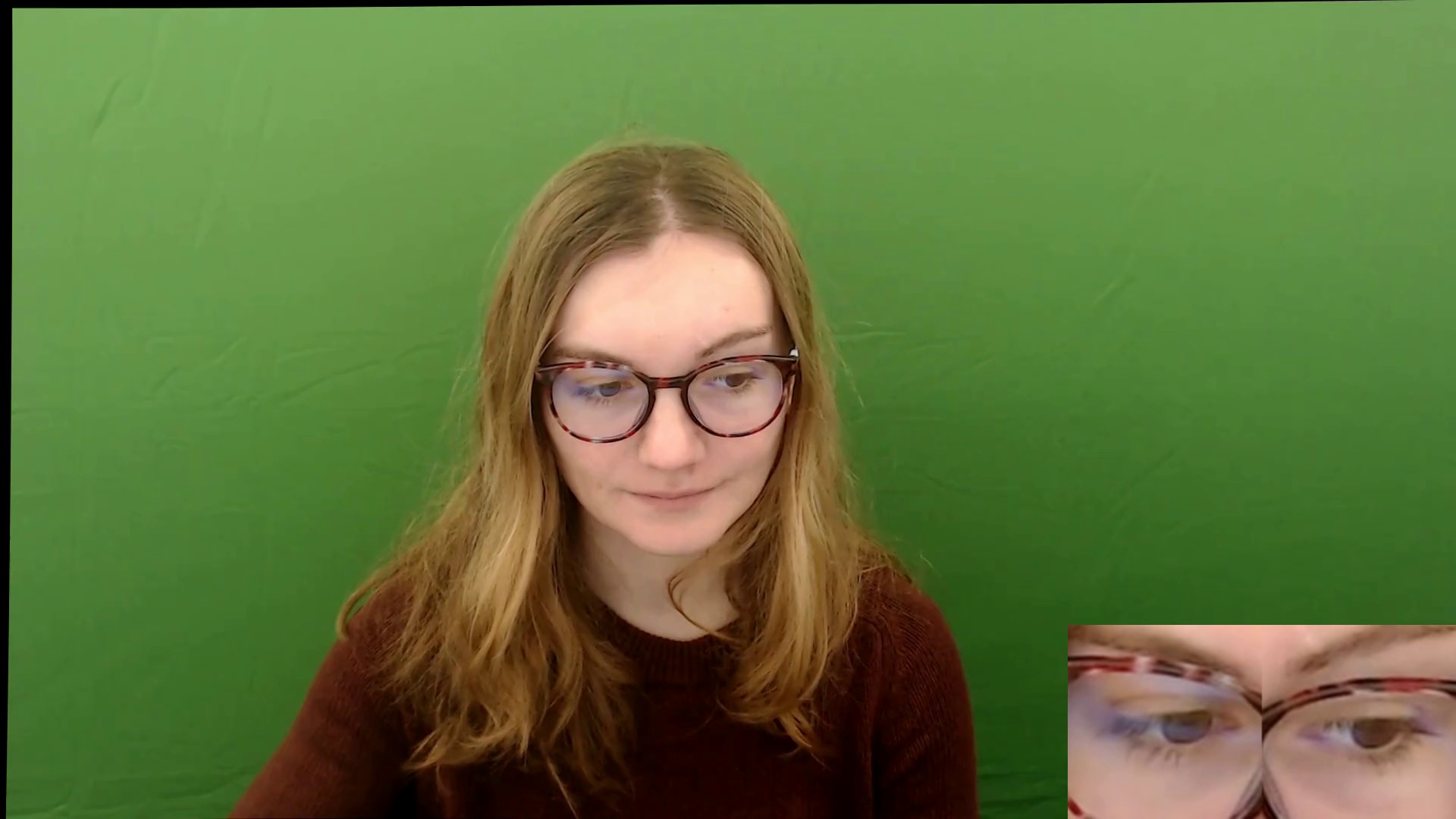}
        \caption{\scriptsize Webcam (Center)}
    \end{subfigure}
    \begin{subfigure}{0.24\columnwidth}
        \includegraphics[width=\columnwidth]{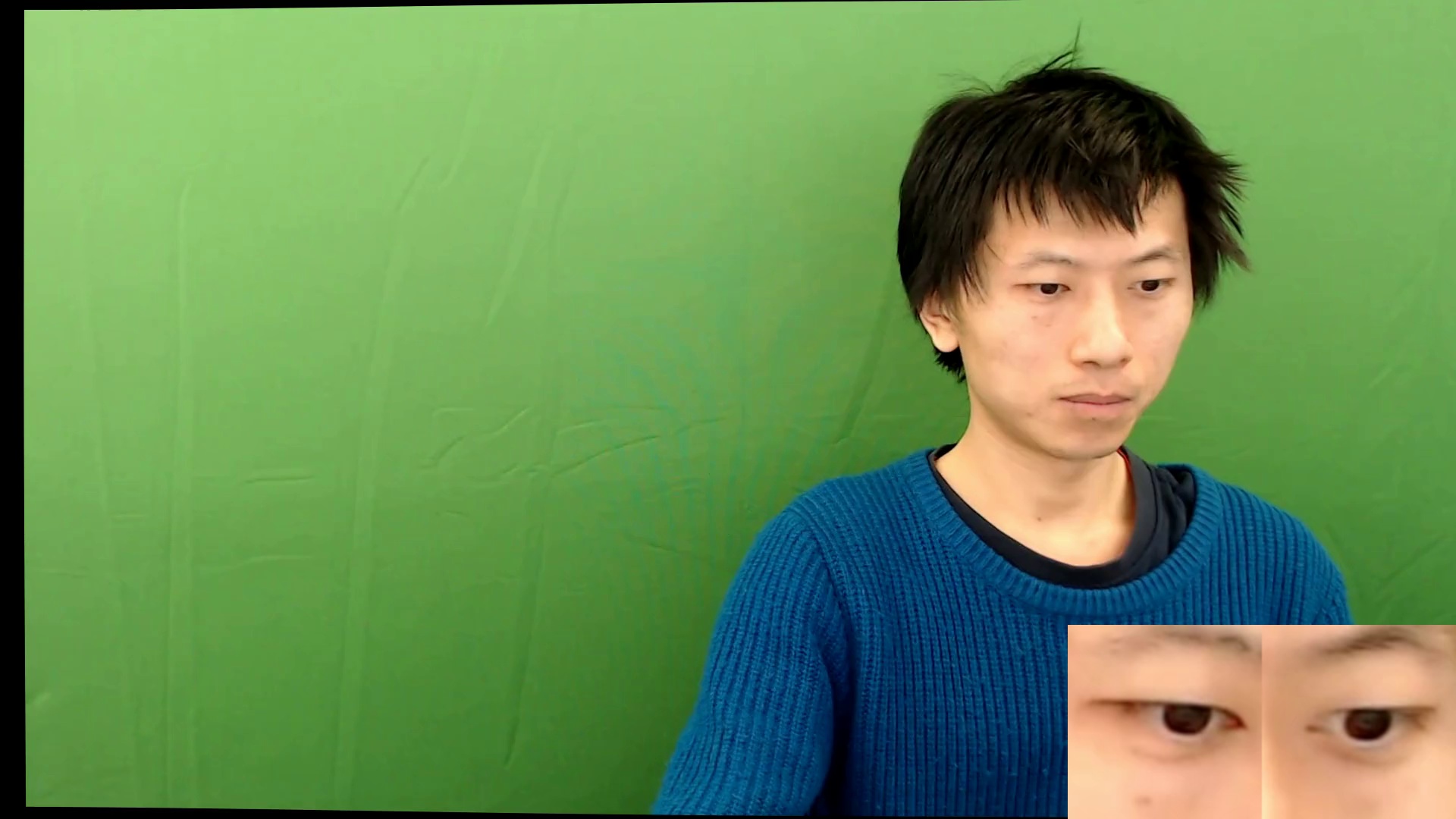}
        \includegraphics[width=\columnwidth]{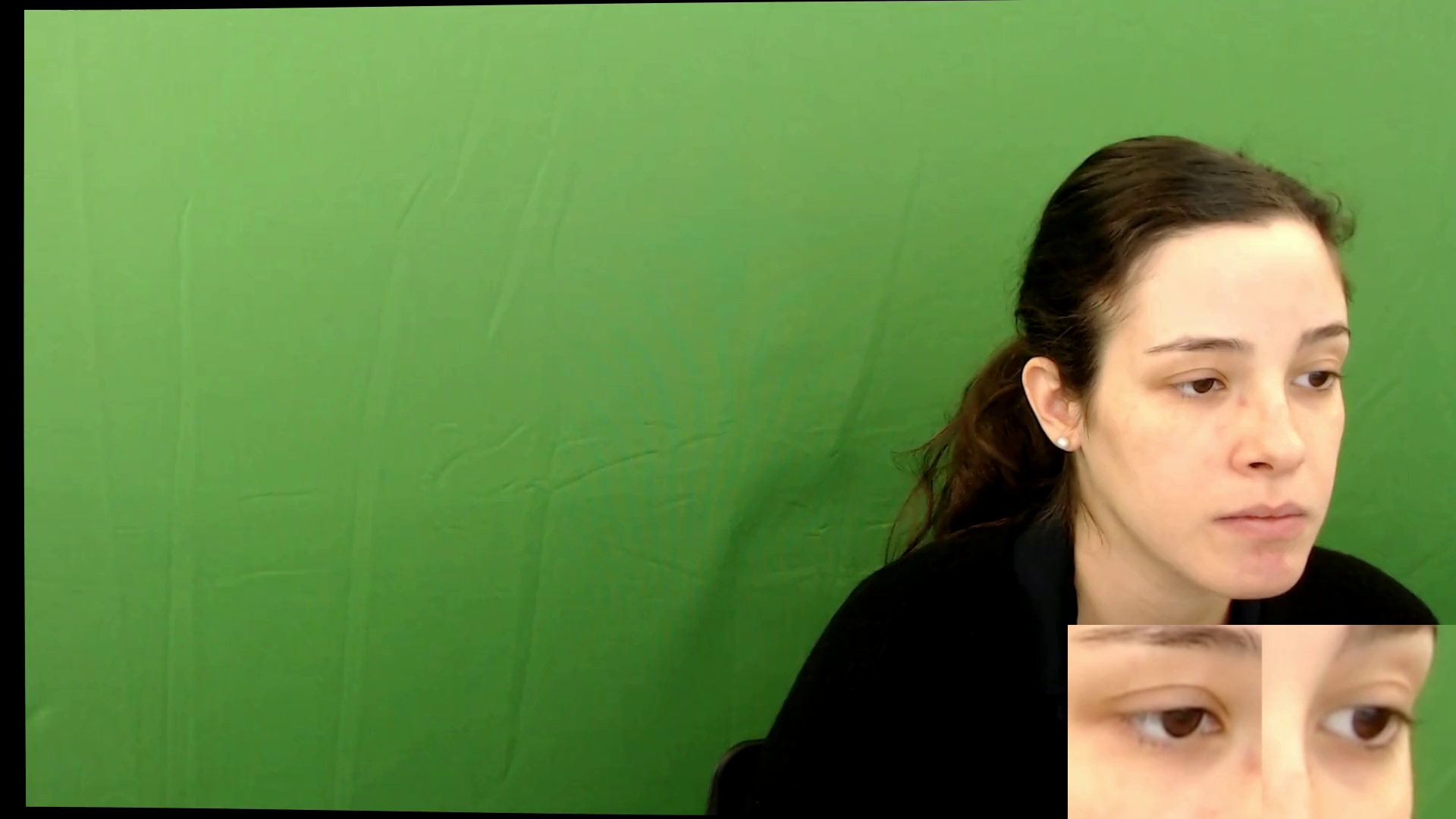}
        \includegraphics[width=\columnwidth]{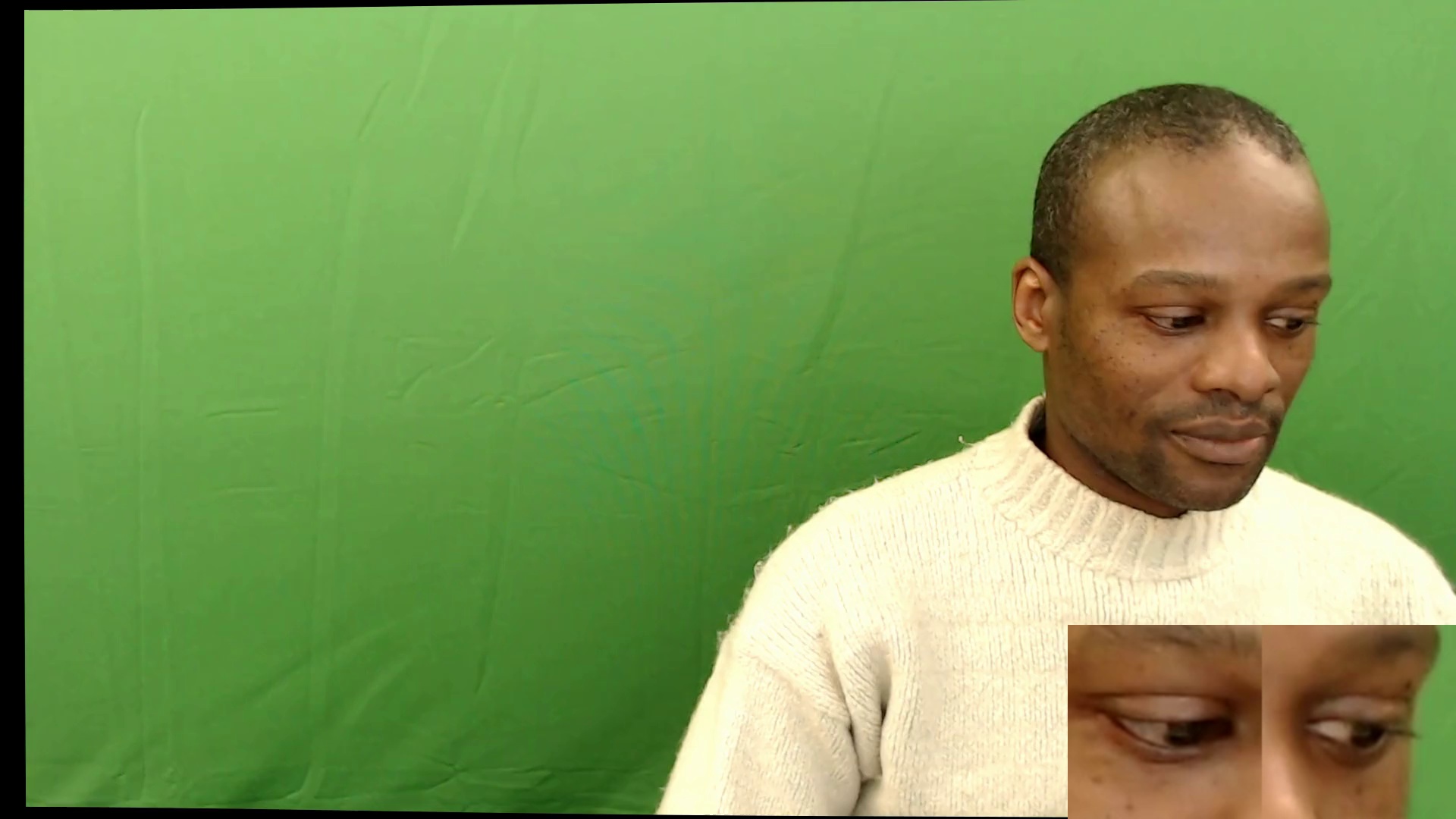}
        \includegraphics[width=\columnwidth]{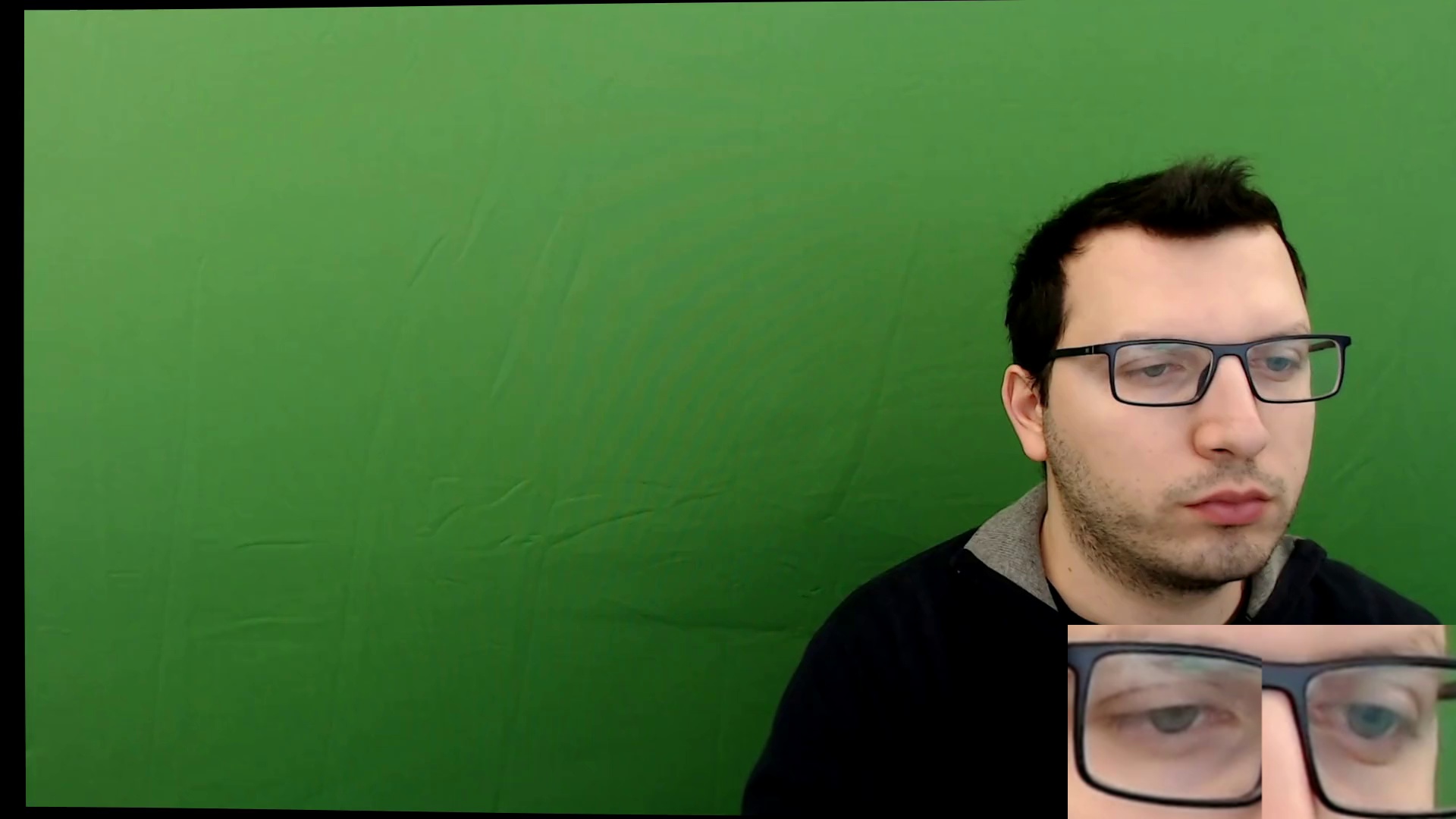}
        \includegraphics[width=\columnwidth]{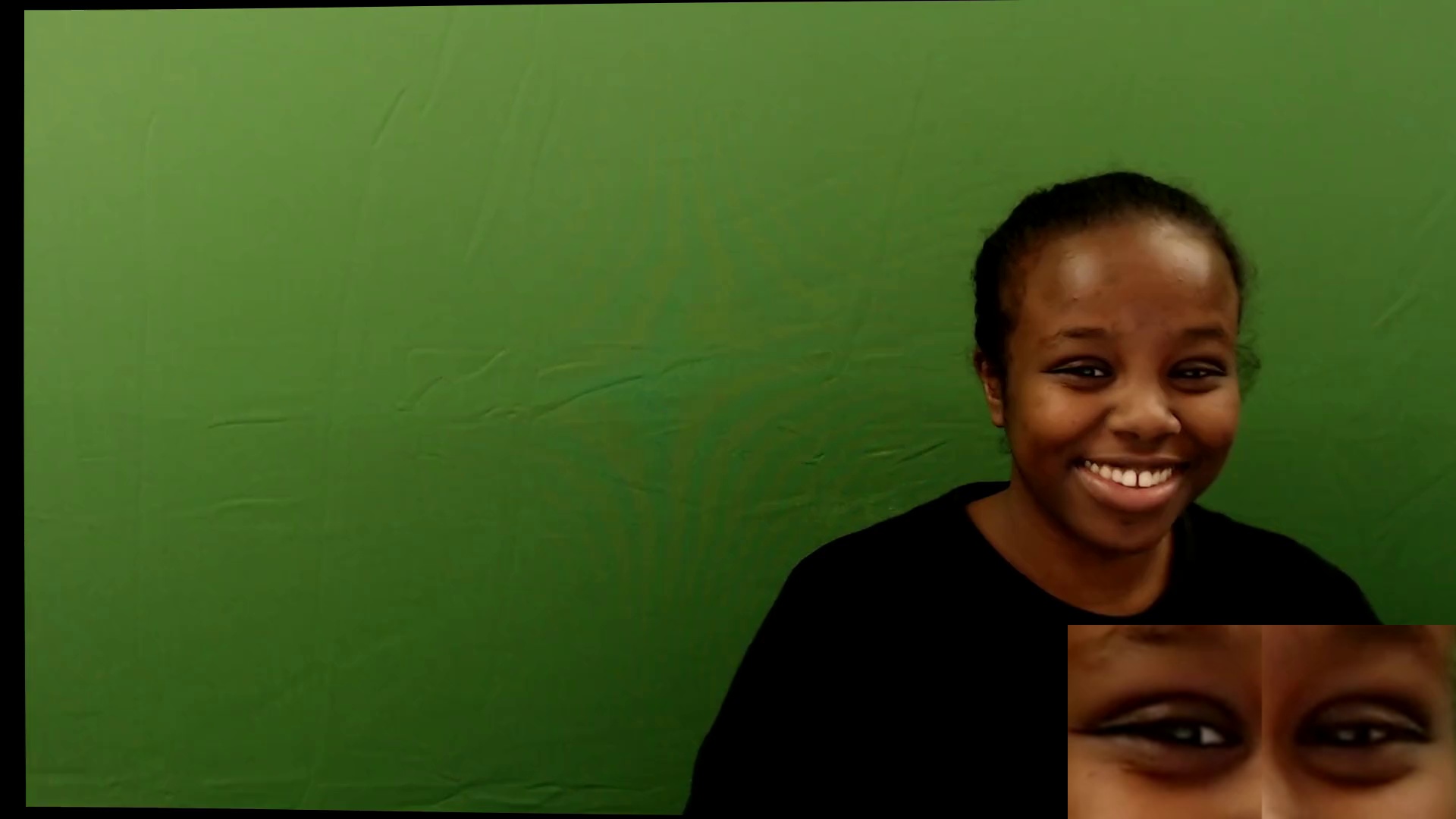}
        \includegraphics[width=\columnwidth]{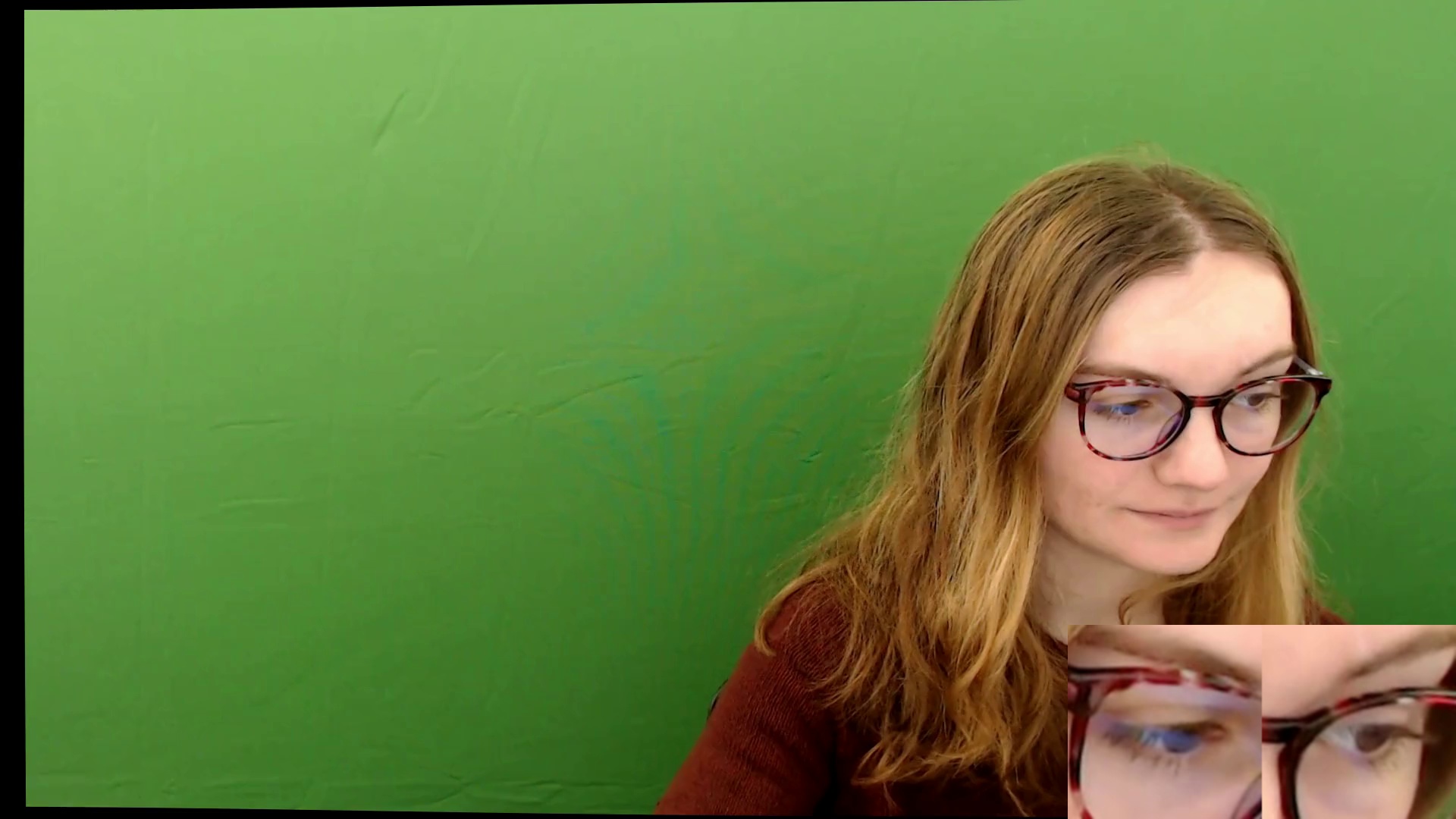}
        \caption{\scriptsize Webcam (Right)}
    \end{subfigure}
    \caption{Example frames from the \datasetname dataset, showing the 4 camera views (Machine Vision Camera or MVC from below, and 3 webcams mounted atop the monitor). Note that the outer webcams in particular capture relatively oblique head orientations. The green screen behind the participants should allow for future works to apply background augmentation for training neural networks.}
    \label{fig:more_sample_frames}
\end{figure}

The quality of eye tracking data can vary greatly depending on specific illumination conditions, ethnicity, gender, and other factors, and as such we placed much care in designing the data collection environment.
For example, we used two separate tables placed on top of a carpeted floor:
one for holding the eye tracker via a VESA-mount arm, and one for the participants to rest their arms or elbows on (cf. Fig.~2 in the main paper). This was done to minimize the transfer of vibrations due to the participants' movements.
We mainly adopted indirect illumination sources for better diffusion of light, and blocked any bright or direct sources of light with black tape or tissue paper. We provide additional samples of collected camera frames in Fig.~\ref{fig:more_sample_frames}.

Yet, not all nuisance factors can be anticipated and as such an experiment coordinator was present at every data collection session to monitor a live-stream of camera frames and eye movements.
We collected a qualitative analysis of gaze data quality in terms of accuracy, precision, and jitter, and provide these alongside the dataset.

\subsection{Dataset Pre-processing}
\label{sec:preprocessing}

To pre-process the collected data, we first performed camera intrinsics calibration using the OpenCV framework. Extrinsic camera transformation determination was done using a first-surface mirror (to avoid errors due to the refraction occurring in standard mirrors) and code released in \cite{Takahashi2016}, with reference points defined by a ChArUco board (flipped as appropriate).
Video was collected for every participant while moving the first-surface mirror around each camera such that the reflected ChArUco board was present across the span of the full camera frame with different inclinations.

In processing the video of participants, we first undistorted the frames' pixels and detected the face \cite{Zhang2017ICCV} and face-region landmarks \cite{Bulat2017ICCV}. We then performed a 3D morphable model (3DMM) fit to the detected 3D facial landmarks \cite{Huber2016} with the purpose of yielding better estimates of gaze ray origins in 3D space. For every participant, we determined a person-specific inter-ocular distance value by exploiting our knowledge of relative camera positions. This inter-ocular distance (defined as the Euclidean distance in millimeters between the outer eye corner landmarks) is then used as a target scale value for scaling every fitted 3DMM. In this way we attempted to further stabilize the yielded eye patches, which were later used as input to our gaze estimation model. The determination of person-specific head-scale was done over 10 randomly sampled frames per participant.

Finally, we applied the ``data normalization'' procedure for yielding eye patches for gaze estimation \cite{Sugano2014CVPR,Zhang2018ETRA}. The final eye patches are $128\times 128$ in size and created with the assumption that the virtual camera is located 60cm away from the defined gaze origin, with a focal length of 1800mm. The selected origin of gaze is an average of the 3D eye corner landmarks of the eye in consideration, taken from the fitted 3DMM found in the previous step.

\begin{figure}[t]
    \centering
    \begin{subfigure}[t]{0.222\columnwidth}
        \includegraphics[width=\columnwidth]{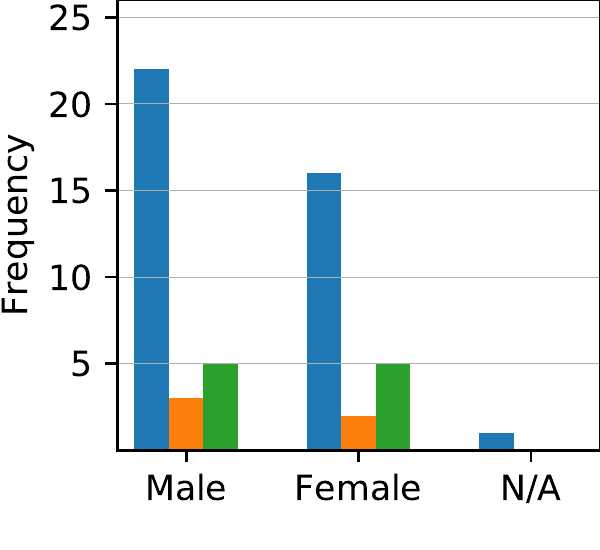}
        \vskip -1mm
        \caption{Biological Sex}
    \end{subfigure}
    \hfill
    \begin{subfigure}[t]{0.74\columnwidth}
        \includegraphics[width=\columnwidth]{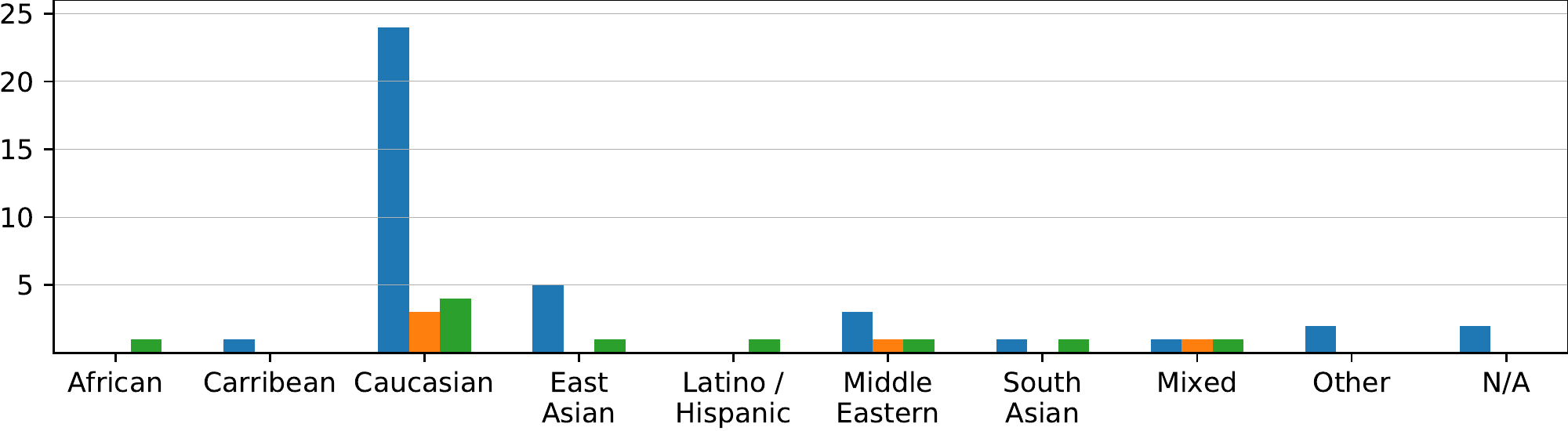}
        \vskip -1mm
        \caption{Association}
        \label{fig:race}
    \end{subfigure}
    \\[3mm]
    \begin{subfigure}[t]{0.222\columnwidth}
        \includegraphics[width=\columnwidth]{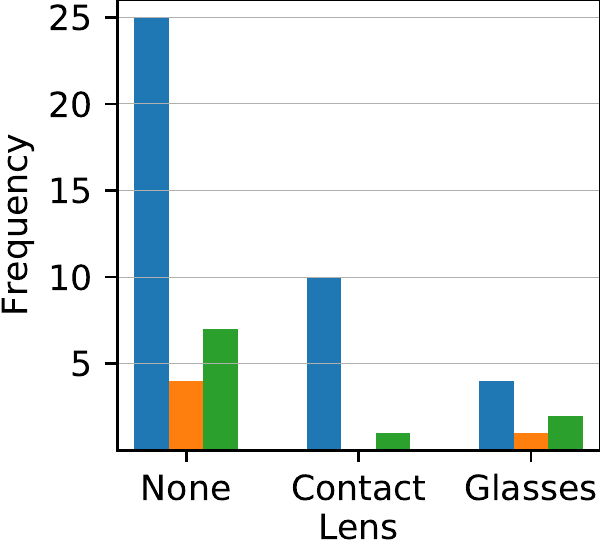}
        \vskip -1mm
        \caption{Visual Aid}
    \end{subfigure}
    \hfill
    \begin{subfigure}[t]{0.295\columnwidth}
        \includegraphics[width=\columnwidth]{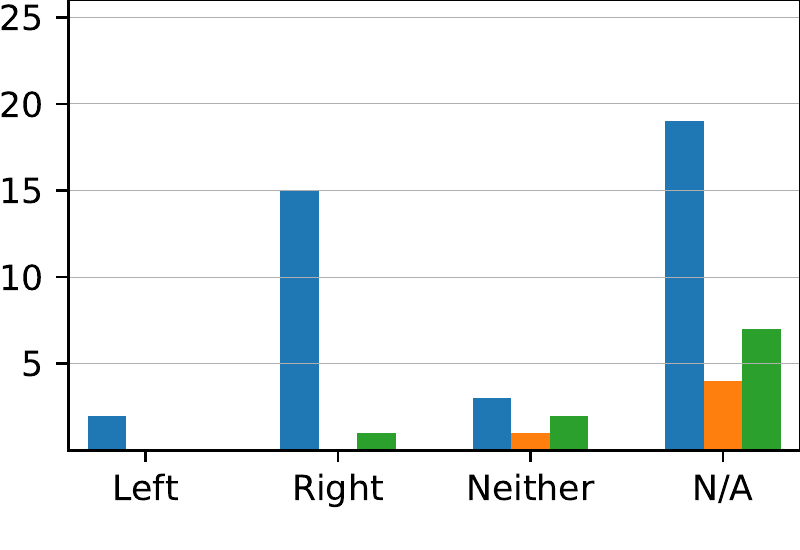}
        \vskip -1mm
        \caption{Dominant Eye}
    \end{subfigure}
    \hfill
    \begin{subfigure}[t]{0.44\columnwidth}
        \includegraphics[width=\columnwidth]{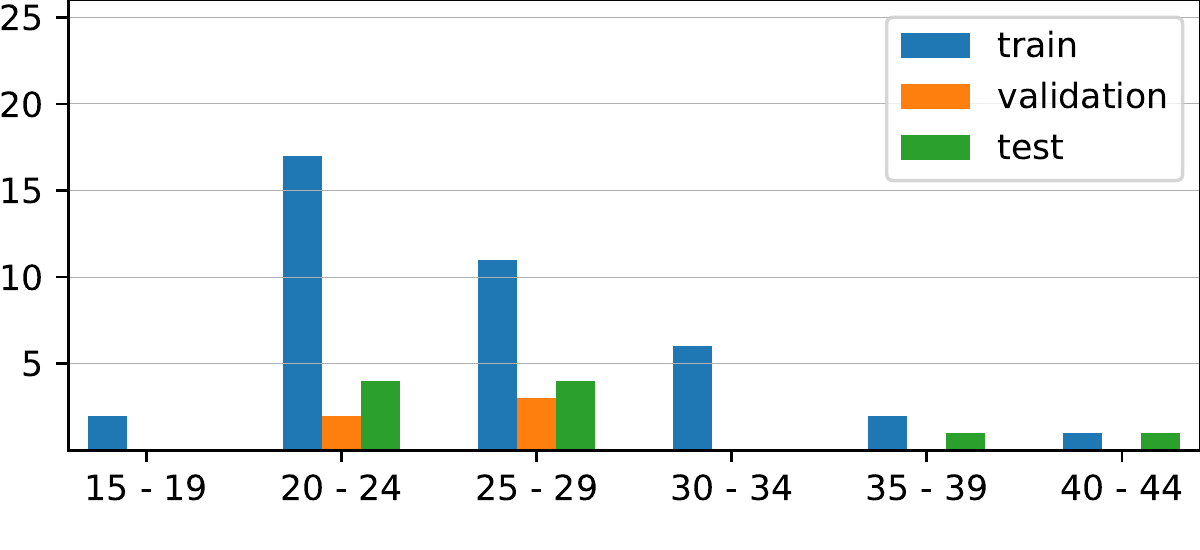}
        \vskip -1mm
        \caption{Age}
    \end{subfigure}
    \caption{Distribution of biological sex, ethnicity, adopted visual-aid, dominant eye, and age in the training, validation and test subsets of \datasetname, based on participants' self-reports. ``N/A'' marks cases where participants either did not know the answer or refused to provide one.}
    \label{fig:demographics}
\end{figure}

\subsection{Dataset Characteristics}

The final dataset is collected from 54 participants (30 male, 23 female, 1 unknown). The distribution in terms of answers to our demographics questionnaire can be seen in Fig.~\ref{fig:demographics}. While there are a few biases in the training data due to the available participant-pool in our local population, the careful selection of our final test set participants (10 participants in total) should allow for conclusions on generalization capabilities to be made.
In particular, it can be seen in Fig.~\ref{fig:race} that we attempted to sample our 10 test set participants from a variety of ethnicities.
More fine-grained per-participant-level information will not be published in order to preserve the participants' privacy.

\begin{figure}[t]
    \centering
    \rotatebox[origin=c]{90}{\scriptsize
        \hspace{10mm}
        Test
        \hspace{8mm}
        Validation
        \hspace{8mm}
        Train
        \hspace{7mm}
        Full \datasetname
    }
    \,
    \begin{subfigure}{0.28\columnwidth}
        \centering
        \includegraphics[width=0.9\columnwidth]{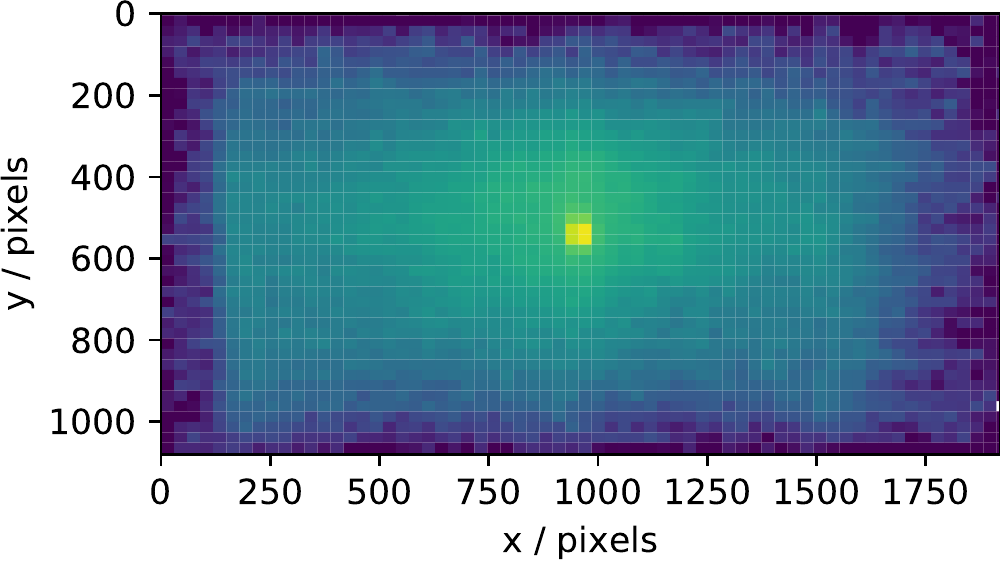} \\[1mm]
        \includegraphics[width=0.9\columnwidth]{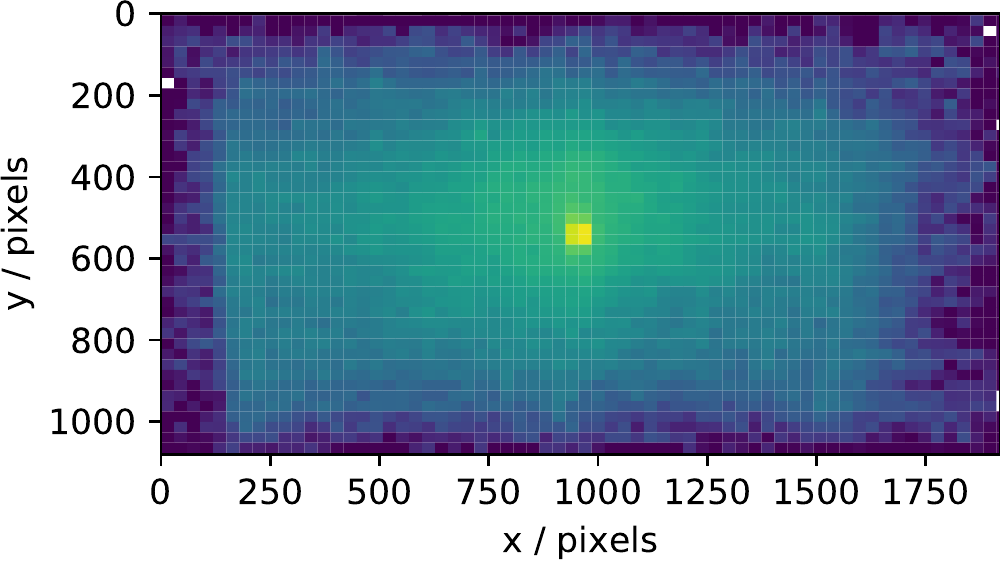} \\[1mm]
        \includegraphics[width=0.9\columnwidth]{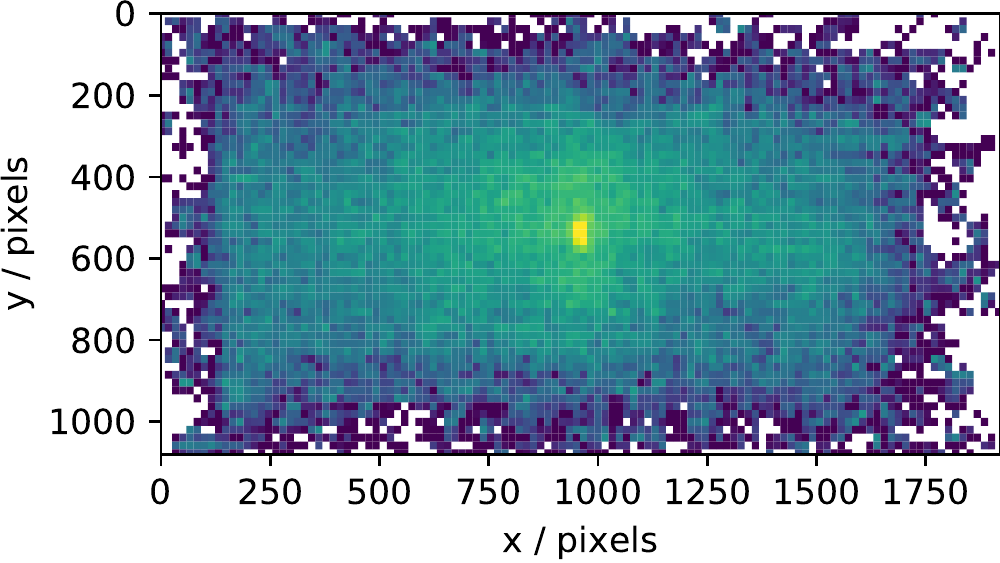} \\[1mm]
        \includegraphics[width=0.9\columnwidth]{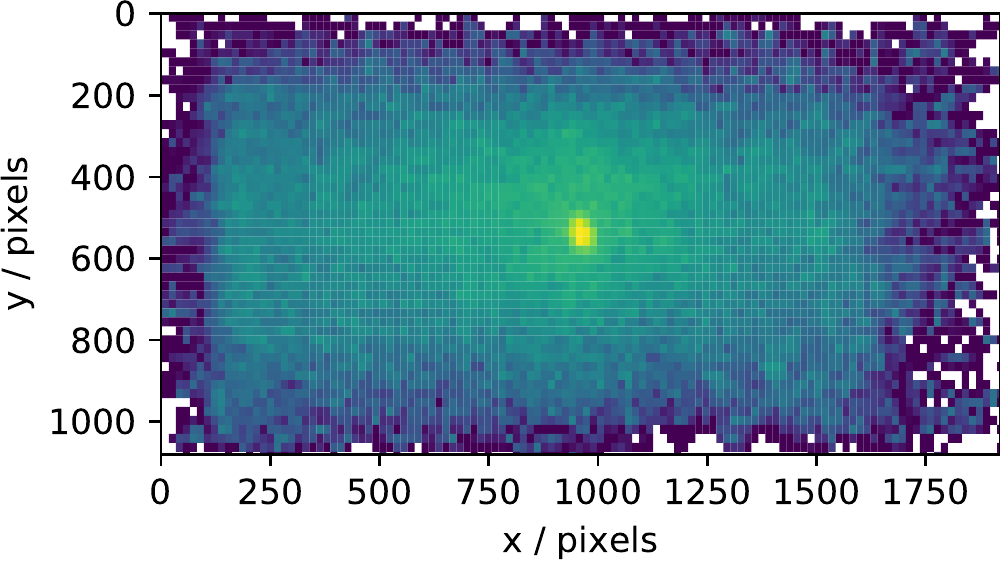} \\[1mm]
        \scriptsize\hspace{4mm}Point-of-Gaze (px)
    \end{subfigure}
    \begin{subfigure}{0.28\columnwidth}
        \centering
        \includegraphics[width=0.9\columnwidth]{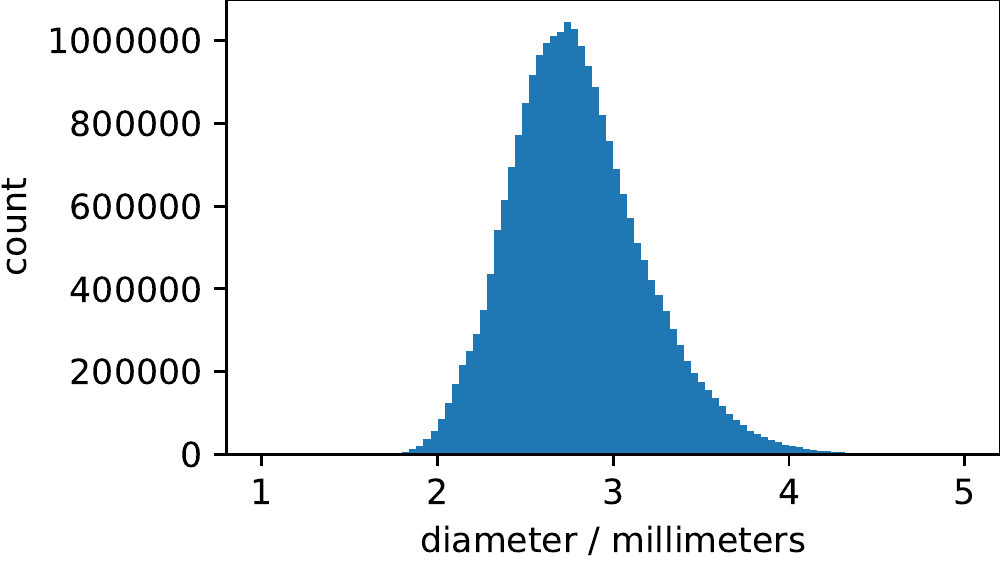} \\[1mm]
        \includegraphics[width=0.9\columnwidth]{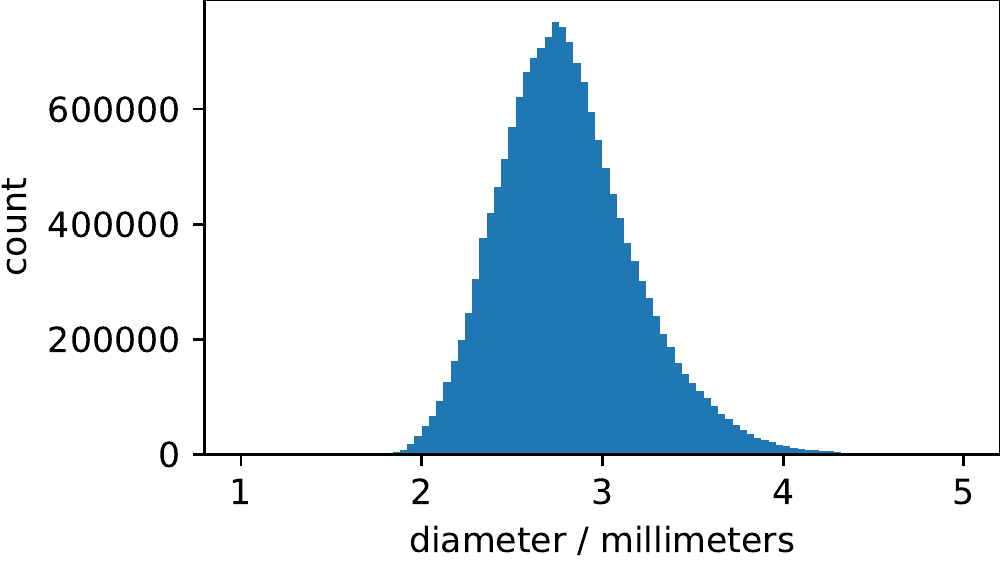} \\[1mm]
        \includegraphics[width=0.9\columnwidth]{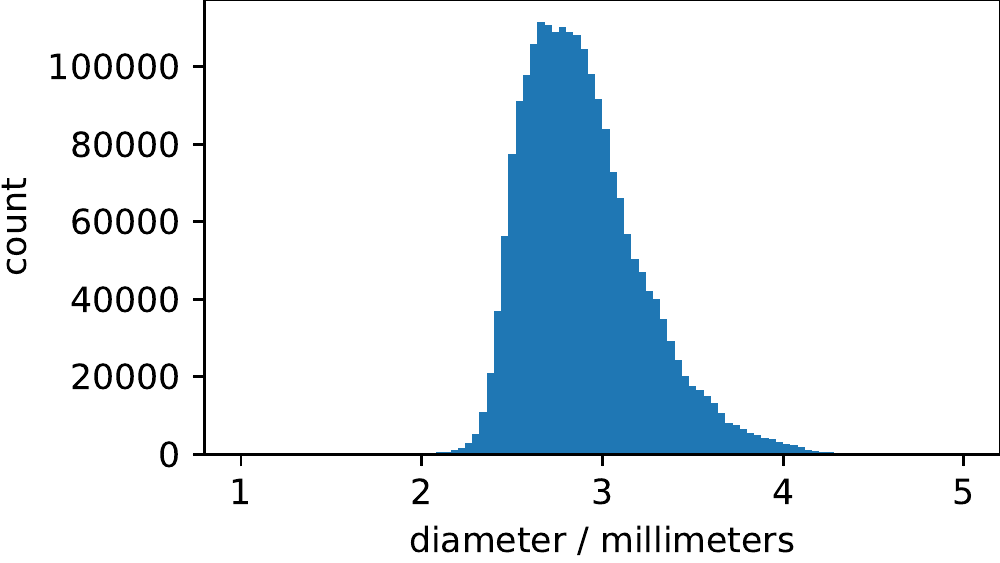} \\[1mm]
        \includegraphics[width=0.9\columnwidth]{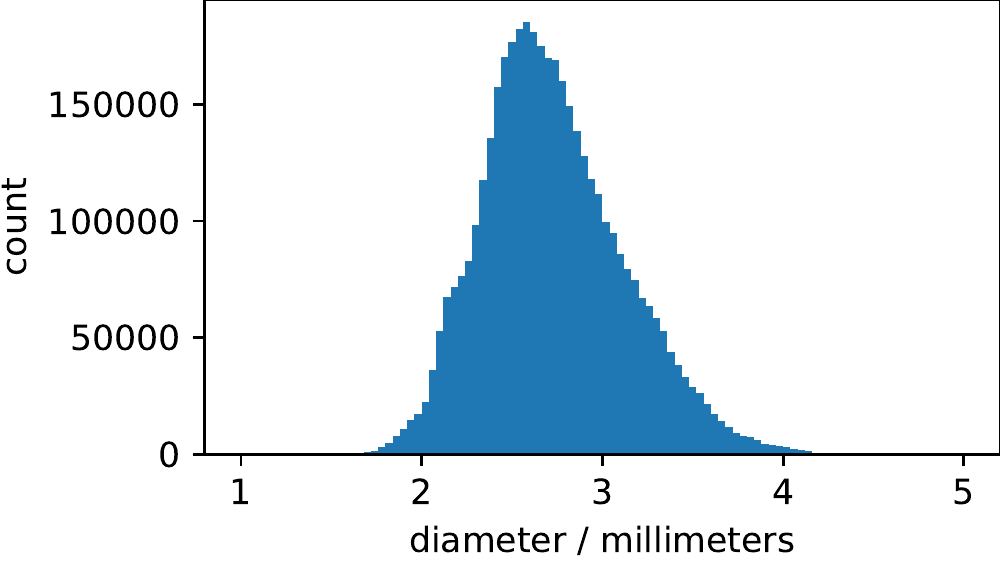} \\[1mm]
        \scriptsize\hspace{5mm}Pupil Size (mm)
    \end{subfigure}
    \begin{subfigure}{0.28\columnwidth}
        \centering
        \includegraphics[width=0.9\columnwidth]{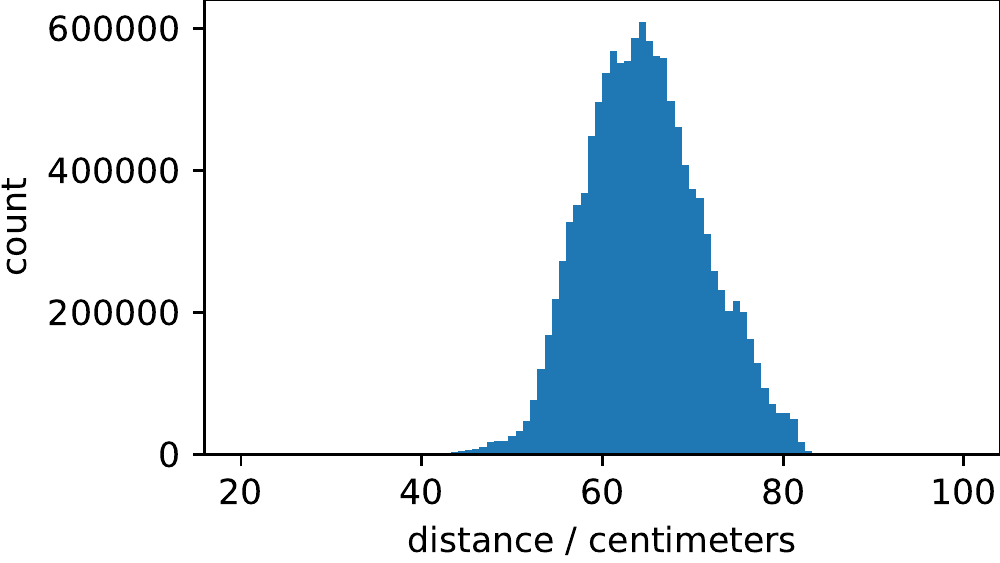} \\[1mm]
        \includegraphics[width=0.9\columnwidth]{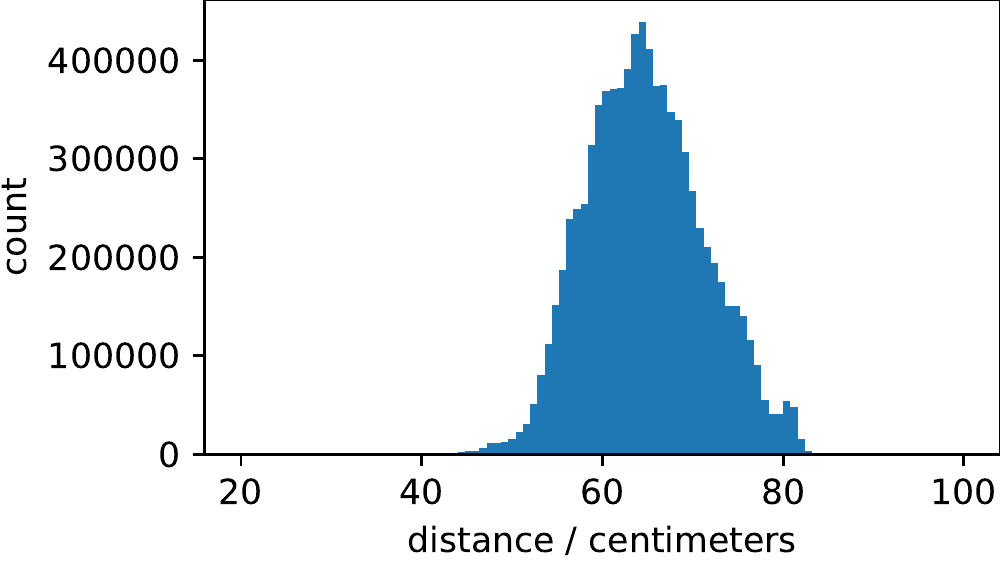} \\[1mm]
        \includegraphics[width=0.9\columnwidth]{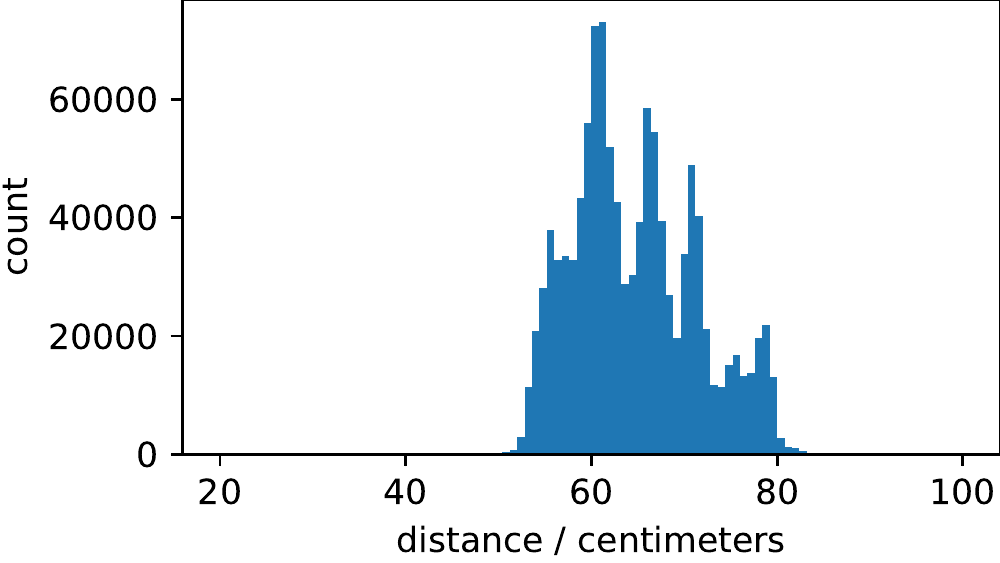} \\[1mm]
        \includegraphics[width=0.9\columnwidth]{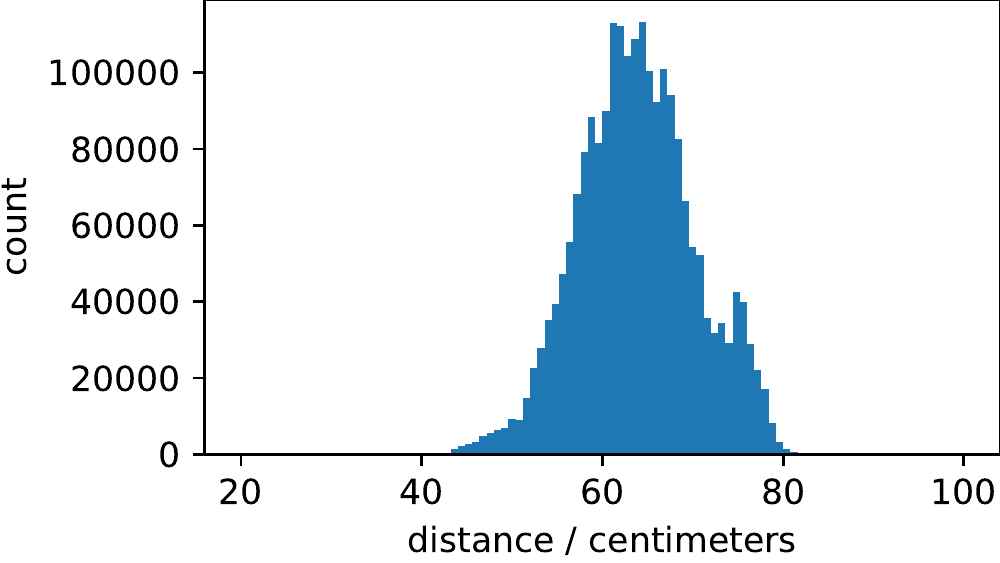} \\[1mm]
        \scriptsize\hspace{4mm}Distance to Head (cm)
    \end{subfigure}
    \vskip -1mm
    \caption{PoG (on-screen pixels), distance (in cm) and pupil size (in mm) distributions for the defined subsets of the proposed \datasetname dataset. The number of people involved are 54, 39, 5, 10 respectively for the full dataset, training subset, validation subset, and test subsets. The 2D histograms are coloured with a logarithmic scale, with values normalized by the size of the subset in concern.}
    \label{fig:misc_distributions}
    \vskip -1mm
\end{figure}

We find that the points-of-gaze (PoGs) in our dataset exhibit a screen-center-bias as previously reported in saliency literature \cite{Judd2009ICCV} (see Fig.~\ref{fig:misc_distributions}). However, this does not indicate that one can naively adjust all estimates of gaze direction to be screen-centered. According experiments are shown in Sec.~\ref{sec:refinenet_without_screen} of this document. A notable fact is that the PoG distribution is similar between the training set and the test set, with samples existing in the peripheral regions of the screen.

Measured pupil diameters (as reported by the Tobii Pro SDK and measured by the Tobii Spectrum Eye Tracker) range between 2mm and 4mm (see Fig.~\ref{fig:misc_distributions}). While this distribution shifts slightly for the test set participants, we find that the pupil sizes are relatively consistent across the defined subsets. Similarly, distances to the participants as estimated by our pre-processing pipeline (see Sec.~\ref{sec:preprocessing}) is consistent across the subsets, and in particular has a mode around the manufacturer recommended distance of 65cm. This demonstrates the care we took in positioning our participants, including a live monitoring of their posture throughout the capture session to avoid large eye tracker errors.

 \section{Offset Augmentation in \refinenetname}

We provide here a step-by-step explanation of our offset augmentation procedure. This method is introduced to address the large differences in performance in gaze estimation when evaluating a network trained on one set of people, on a new set of people. The person-specific differences are often described as being a consistent offset (also called ``angle kappa''), which do not appear in computed training losses, but only in the validation or test losses. We thus implement our augmentation to mimic the effect of this angle kappa.

First, given an estimate for gaze direction $\hat{\bv{g}}$, let us assume that this is represented in spherical coordinates representing pitch and yaw angles such that $\theta$ is pitch, and $\phi$ is yaw. Then the unit-vector notation of $\hat{\bv{g}}=\left(\theta,\,\phi\right)$ would be calculated with,
\begin{equation}
    \bvh{v}_h =
    \begin{pmatrix}
        -\cos\theta\sin\phi \\
        -\sin\theta \\
        -\cos\theta\cos\phi \\
    \end{pmatrix}.
\end{equation}

As the vector was previously defined such that $\left(\theta,\,\phi\right)=\left(0,\,0\right)$ points towards the camera, we must flip the vector via negation to bring it to the camera-relative coordinate system in which the head model (3DMM) is defined.

Assuming that we know the rotation of the head with respect to the camera (from which the input image was taken from), we then apply the inverse of this known rotation $\bv{R}_h$ to calculate the gaze direction relative to the head coordinate system:
\begin{equation}
    \bvh{v}_h=\bv{R}_h^T\bvh{v}_c.
\end{equation}

We now return this head-relative gaze direction value to spherical coordinates, with:
\begin{equation}
    \bvh{g}_h
    =
    \begin{pmatrix}
      \theta_h \\
      \phi_h
    \end{pmatrix}
    =
    \begin{pmatrix}
        \arcsin-\hat{y}_h \\
        \mathrm{arctan2}\left(-\hat{x}_h,\,-\hat{z}_h\right) \\
    \end{pmatrix},
\end{equation}
where $\bvh{v}_h=\left(\hat{x}_h,\,\hat{y}_h,\,\hat{z}_h\right)$.
The corresponding rotation matrix is then,
\begin{equation}
    \bv{R}_h=
    \begin{pmatrix}
        \cos\phi_h & 0 & \sin\phi_h \\
        0 & 1 & 0 \\
        -\sin\phi_h & 0 & \cos\phi_h \\
    \end{pmatrix}
    \begin{pmatrix}
        1 & 0 & 0 \\
        0 & \cos\theta_h & -\sin\theta_h \\
        0 & \sin\theta_h & \cos\theta_h \\
    \end{pmatrix}.
\end{equation}

This is the rotation that we can apply on top of a constant sequence-specific and synthetic ``offset''. Per given training sequence of length $T$ (to maintain consistency with the main paper, Eq.2), we acquire a sequence-specific offset $\kappa_i = \left(\theta_\kappa,\,\phi_\kappa\right) \sim\mathcal{N}\left(0,\,3^\circ\right)$ that is parameterized with pitch and yaw angle values as done in defining $\bvh{g}$. We determined the standard deviation of $3$ degrees empirically, and show this in degrees for convenience of understanding. In reality, the sampled values are in radians.

We convert the kappa values to unit vector notation and rotate it by the current gaze direction matrix,
\begin{equation}
    \bvh{v}_h^\mathrm{aug}=\bv{R}_h\bvh{v}_\kappa.
\end{equation}

This augmented gaze direction is transformed back to the normalized camera coordinates system such that the frontal gaze is defined with $\left(\theta,\,\phi\right) = \left(0,\,0\right)$.

\section{Implementation Details}
To facilitate faithful reproduction of our experiments, we provide additional implementation details of our architecture and its training and hyper-parameters.

\subsection{Validity of Ground-truth Labels}
The ground-truth data provided by the \datasetname dataset often comes from the Tobii Spectrum Pro eye tracker, and associated Tobii Pro SDK. As is often the case with eye trackers, there are cases where tracking fails, such as during eye blinks or when illumination conditions are too poor for features to be tracked. The ``validity'' of predicted ground-truth is provided by the SDK, and stored alongside all other labels. We apply the validity boolean values to our loss calculation, such that only valid ground-truth labels are used during training.

The collected screen frames and Tobii-origin data do not perfectly coincide in terms of reported timestamps. We perform a manual alignment to ensure consistency between images of the eye-region and the gaze data, and additionally perform bilinear interpolation in PoG given that valid labels exist on both sides (immediately before and immediately after) of the query timestamp. As the eye tracking data is collected at $150$Hz (as a reminder, the camera frames have been collected at $30$Hz or $60$Hz), and by the Nyquist-Shannon sampling theorem, we can assume that the eye tracking data has been reliably handled.

\subsection{\eyenetname}
In the main paper (cf. Sec.~4.1), we defined the loss terms for gaze direction as $\mathcal{L}_\mathrm{gaze}$ and for pupil size as $\mathcal{L}_\mathrm{pupil}$. We define the full loss as:
\begin{equation}
    \mathcal{L}_\mathrm{EyeNet} =
    \gamma_\mathrm{PoG} \mathcal{L}_\mathrm{gaze} +
    \gamma_\mathrm{pupil} \mathcal{L}_\mathrm{pupil}  ,
\end{equation}

and set $\gamma_\mathrm{gaze} = 1.0$ and $\gamma_\mathrm{pupil} = 1.0$ empirically.
The \eyenetname is trained using the Adam optimizer \cite{Kingma2014} for 8 epochs using a batch size of 16, and $l_2$ parameter decay of $0.005$. We apply exponential learning rate decay of factor $0.5$ every $1$ epoch, beginning from a learning rate of $0.016$.
The input eye image is resized to be $128\times 128$ pixels large.

\subsection{\refinenetname}
The \refinenetname adopts the a mean-squared error loss term for the final PoG (calculated via a soft-argmax layer, cf. Fig.~4b of main paper), and in addition applies a per-pixel cross-entropy loss for guiding the learning of the heatmap. When defining the cross-entropy based loss term as $\mathcal{L}_\mathrm{XE}$, we can then define the full loss as:
\begin{equation}
    \mathcal{L}_\mathrm{RefineNet} =
    \gamma_\mathrm{PoG} \mathcal{L}_\mathrm{PoG} +
    \gamma_\mathrm{XE} \mathcal{L}_\mathrm{XE}  .
\end{equation}
where we set $\gamma_\mathrm{PoG} = 0.001$ and $\gamma_\mathrm{XE}=1.0$ empirically.
The \refinenetname is trained using the Adam optimizer \cite{Kingma2014} for 4 epochs using a batch size of 8, and $l_2$ parameter decay of $0.0$. We apply exponential learning rate decay of factor $0.5$ every $0.5$ epochs, beginning from a learning rate of $0.008$.
The input screen content frame is resized to be $128\times 72$ pixels large.
Please note that during this stage of training, the \eyenetname weights are not updated.

 \section{Additional Results}

Here, we provide additional details with respect to the results shown in Sec.~5 of the main paper, as well as new experiments which further assess our \refinenetname architecture. In particular, we experiment with pre-training the gaze estimation network (\eyenetname) on an existing in-the-wild dataset, and applying it directly and without modification as part of the \refinenetname training. Next, we attempt to understand the inter-play of the proposed offset augmentation and screen content input. We then evaluate the robustness of the \refinenetname training to the different error characteristics of the 4 camera views. Lastly, we show the changes in \refinenetname performance with varying strength of offset augmentation applied during training.

\subsection{Evaluation Details}

In all experiments, we evaluate on the test split of the \datasetname dataset consisting of $10$ participants. To reduce the data load of both training and evaluation, we subsample all data such that we take $10$ samples per second. A sequence is defined to span $3$ seconds of time such that the shortly exposed image stimuli sequences can be trained on as well (exposure time of $3$ seconds to participants). Effectively, this means that we sub-sample the number of frames by a factor of $\frac{1}{6}$ and $\frac{1}{3}$ respectively for the machine vision camera and webcams.

For both training and evaluation, we cut all available video data into $3$-second-long sequences without gaps or overlaps. This results in $65,116$ sequences in the training sub-set, $7,676$ sequences in the validation sub-set, and $17,660$ sequences in the test sub-set. There are $2,392$ image-stimulus sequences, $10,472$ video-stimulus sequences, and $4,796$ wikipedia-stimulus sequences in the test sub-set.

\subsection{Training \eyenetname on GazeCapture}
\begin{table}[t]
    \centering
    \caption{Experiments where the \eyenetstaticname is trained on the GazeCapture dataset \cite{Krafka2016CVPR}. The initial error is high as is typical of eye-patch input gaze estimation networks evaluated in the cross-dataset setting. We see that despite the high initial error, a respectably low error is achieved when training a \refinenetgruname atop the predictions from the GazeCapture-trained \eyenetstaticname. We thus show that our refinement approach can be used in combination with existing gaze estimators to bridge dataset domain gaps
    }
    \label{tab:gazecapture}
    \renewcommand{\arraystretch}{1.3}
    \begin{tabular}{|l|l|l|l|}
        \hline
        Model & Gaze Dir. ($^\circ$) & PoG (cm) & PoG (px) \\
        \hline
        Baseline (both eyes) &         $7.93$        &         $8.86$        &         $288.85$        \\
        \refinenetgruname    & \resultb{3.93}{50.57} & \resultb{4.33}{51.12} & \resultb{150.29}{47.97} \\
        \hline
    \end{tabular}
\end{table}

In order to assess our contribution in the context of existing gaze estimation methods and datasets, we identified that training the gaze estimation part of our architecture (\eyenetstaticname) and using it without modification to learn the final refinement step, would be the most challenging benchmark. We evaluate this setting by training our \eyenetstaticname on the GazeCapture dataset \cite{Krafka2016CVPR} with equivalent pre-processing steps to our data, then train a \refinenetgruname while keeping the \eyenetstaticname fixed, to finally evaluate performance on the test set of our \datasetname dataset. We select our own test set as no other publicly available video-based gaze dataset exhibit natural eye movements. The baseline gaze direction error of $7.93^\circ$ shown in Tab.~\ref{tab:gazecapture} is typical of network architectures that take single-eye inputs (we perform single-eye gaze estimation to enable binocular gaze estimation in the future - an interesting output for studies on vergence), as shown in recent works \cite{Zhang2019TPAMI,Wood2016ECCV}. We find that a highly significant improvement can be made even with initial errors as large as 27\% of the screen height (1080 pixels). This shows that dataset differences can easily be overcome with our \refinenetname training, even in the absence of labeled data from test users, and while retaining the errors present in the trained \eyenetstaticname (its weights are not changed during \refinenetgruname training).

\subsection{Offset augmentation without screen content}
\label{sec:refinenet_without_screen}
\begin{table}[t]
    \centering
    \caption{Ablation study to further understand the effect in the absence of any screen content input. Each row adds a factor (such that the last row includes all changes). The refinement network without screen content simply refines a given heatmap, and thus could be considered a method of screen-center-bias enforcement, a form of gaze position prior.
    }
    \label{tab:without_screen}
    \renewcommand{\arraystretch}{1.3}
    \begin{tabular}{|l|l|l|l|}
        \hline
        Model & Gaze Dir. ($^\circ$) & PoG (cm) & PoG (px) \\
        \hline
        Baseline (both eyes)  &         $3.48$        &         $3.85$        &         $132.56$  \\
        + Refinement Network* &  \result{3.41}{ 2.18} &  \result{3.77}{ 2.17} &  \result{130.78}{ 1.34} \\
        + Offset Augmentation &  \result{3.00}{13.84} &  \result{3.31}{13.90} &  \result{115.10}{13.17} \\
        + Screen Content      & \resultb{2.49}{28.43} & \resultb{2.75}{28.49} & \resultb{ 95.59}{27.89} \\
        \hline
    \end{tabular}
    \vskip 0.5mm
    \begin{minipage}{0.8\columnwidth}\raggedleft\scriptsize
        \textbf{*} with GRU and skip connections between encoder and decoder.
    \end{minipage}
\end{table}

To better understand the effect of the screen content input, we performed an ablative study of our contributions in the absence of screen content. What this means is that no appearance-based context is given to the task of gaze estimate refinement, except for the dimensions of the heatmap with which PoG is represented. More specifically, the \refinenetname could be conjectured to be performing a center-bias application. We find in Tab.~\ref{tab:without_screen} that this assumption is only partly true, and that applying the \refinenetname alone without screen input nor offset augmentation results in comparable results to the baseline. This means that the center-bias present in the data is not useful in further improving gaze estimates. We do find however, that the offset augmentation still works relatively well in the absence of screen content. With screen content input we can reach the final best reported performance.

\subsection{Cross-Camera Evaluation}
\begin{table}[t]
    \caption{Final refined gaze direction errors (in degrees, lower is better) for cross-camera evaluations. While testing on the high-quality machine vision camera frames yield the best results, it can be seen that the refinement step is mostly agnostic to where the gaze data comes from and can generalize to gaze data from new views, despite differences in characteristics of the error }
    \label{tab:cross_camera_refinement}
    \centering
    \renewcommand{\arraystretch}{1.3}
    \begin{tabular}{|l|p{1.8cm}|p{1.8cm}|p{1.8cm}|p{1.7cm}|}
       \hline
       \diagbox[width=2.8cm]{Source}{Target} & Webcam (Left) & Webcam (Center) & Webcam (Right) & MVC \\
       \hline
        Webcam (Left)        & \resultb{3.03}{21.62} & \resultb{2.55}{23.47} & \result{3.12}{22.79}  & \resultb{2.24}{16.90} \\
        Webcam (Center)      & \resultb{3.04}{21.53}   & \resultb{2.55}{23.53}  & \result{3.11}{22.96}  & \result{2.26}{16.31}  \\
        Webcam (Right)       & \result{3.07}{20.52}   & \result{2.58}{22.66}  & \result{3.14}{22.17} & \result{2.29}{15.32}  \\
        MVC                  & \resultb{3.03}{21.69}   & \resultb{2.54}{23.70}  & \resultb{3.09}{23.36}  & \resultb{2.23}{17.50}  \\
       \hline
    \end{tabular}
    \\[1mm]
    \hspace{0.15\columnwidth}
    \begin{minipage}{0.85\columnwidth}
        \scriptsize\raggedright
        Note: MVC stands for ``Machine Vision Camera''. \\[0mm]
        \hspace{7mm} Improvements are with respect to initial PoG estimates from \eyenetgruname.
    \end{minipage}
\end{table}
To assess the sensitivity of our \refinenetname approach, we evaluate performance changes when training on predicted gazes from different camera views in Table~\ref{tab:cross_camera_refinement}, where gaze estimates are still provided from a pre-trained \eyenetgruname but the \refinenetgruname is trained from gaze data only from the source camera view, and tested on frames from the target camera view.
We find that in general, the best performances can be seen when evaluating on the machine vision camera frames, as image quality and detail are expectedly higher.
Nonetheless in general, improvements can be seen across the board, showing that the \refinenetname is not sensitive to changes in camera view (and the consequent change in the errors of initial PoG predictions).

\subsection{Effect of offset augmentation strength}
\label{sec:kappa_sweep}
\begin{figure}[t]
    \centering
    \includegraphics[width=0.7\columnwidth]{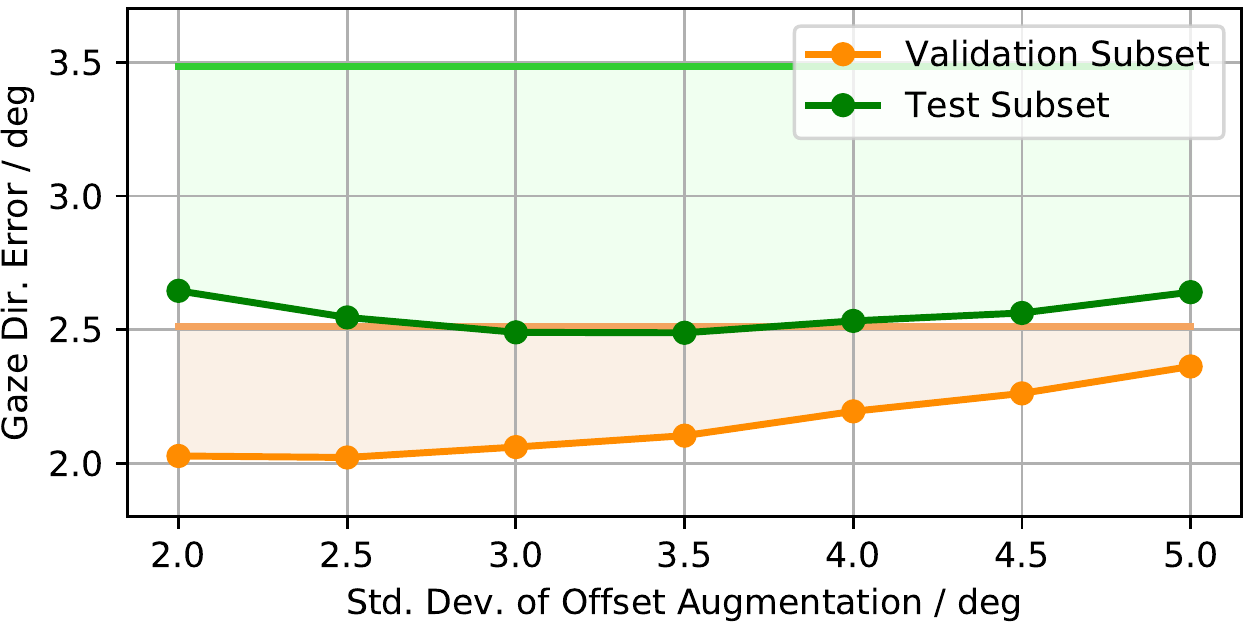}
    \vskip -2mm
    \caption{Varying $\sigma_\kappa$ results in differing performance improvements on the validation subset of the EVE dataset, compared to that on the test subset.
    Specifically, the test subset is significantly more challenging and thus a stronger amount of offset augmentation is required than in the case of the validation subset.
    }
    \label{fig:kappa_sweep}
\end{figure}
The amount of offset to apply to initial gaze direction predictions is an important hyperparameter. For example, a \refinenetgruname trained with weak offset augmentation may not handle high test-time offsets whereas a \refinenetgruname trained with strong offset augmentation may perform overly aggressive corrections. We show this trade-off in Fig.~\ref{fig:kappa_sweep} where we see that the relatively easier validation requires lower amounts of offset augmentation at training time compared to the test set.

A more comprehensive study of discrepancies between learned models' predictions of gaze direction should be performed in the future, in relation to the differences in demographics in various gaze datasets.
Furthermore, these offsets are most certainly not due to textbook anatomical differences only (between optical and visual axes in each eyeball).
For instance, the determination of 3D gaze origin is always done in an approximate manner and may vary greatly depending on (a) how the head pose was determined, and (b) how the head-pose-relative gaze origin was determined.
In pre-processing the \datasetname dataset, we apply a 3DMM fitting approach with interocular-distance-based scale-normalization to alleviate these issues.
 \section{Ethical Considerations}
In this work we effectively demonstrate that it is possible to improve predictions of PoG given the screen content, even without prompting the user (ground-truth label acquisition or gaze estimator calibration). We are certain that the field will progress quickly, and will soon be reporting methods and architectures which yield higher accuracy and robustness for screen-based eye tracking based on our initial insights and the \datasetname dataset.

We are aware of the ethical implications of further developments to our approach in the context of data privacy.
Specifically, a malicious agent could attempt to elicit information regarding a user's habits or preferences without their awareness.

To eliminate such efforts, we hope that operating system developers can build secure sandbox environments where front-facing camera usage is increasingly restricted. Furthermore, we recommend that the Computer Vision community work on:
\begin{inparaenum}[(a)]
\item allowing for light-weight model architectures through knowledge distillation or weight quantization to quickly enable edge-only prediction of eye gaze such as to restrict the transfer of original front-facing camera frames, and
\item development of eye movement descriptors which need not expose fine-grained person-specific traits yet assist in intelligent interactive systems such as user-state-aware interfaces (e.g. changing of layout or appearance based on perceived stress or cognitive load).
\end{inparaenum} 
\end{document}